\documentclass[sigconf]{acmart}
\usepackage[ruled]{algorithm2e}
\usepackage{enumitem}
\usepackage{multirow,booktabs}
\usepackage{arydshln}
\usepackage{pifont}
\makeatletter
\newcommand{\ssymbol}[1]{^{\@fnsymbol{#1}}}
\makeatother
\usepackage{threeparttable}

\usepackage[most]{tcolorbox}
\usepackage{makecell}

\usepackage{subfigure} 
\theoremstyle{definition}
\newtheorem{definition}{Definition}
\newtheorem{prop}{Proposition}

\usepackage{array} 

\newtcbox{\roundbox}[1][]{
    on line, 
    colback=orange!20, 
    colframe=orange!20, 
    arc=0.5mm, 
    boxrule=0.0pt, 
    left=-1pt, right=-1pt, top=-1.6pt, bottom=-2pt, 
    fontupper=\scriptsize, 
        valign=center, 

    #1 
}

\AtBeginDocument{%
  \providecommand\BibTeX{{%
    \normalfont B\kern-0.5em{\scshape i\kern-0.25em b}\kern-0.8em\TeX}}}

\setcopyright{acmcopyright}
\copyrightyear{2018}
\acmYear{2018}
\acmDOI{XXXXXXX.XXXXXXX}

\acmConference[Conference acronym 'XX]{Make sure to enter the correct
  conference title from your rights confirmation emai}{June 03--05,
  2018}{Woodstock, NY}
\acmPrice{15.00}
\acmISBN{978-1-4503-XXXX-X/18/06}

\begin{document}

\title{Devil's Hand: Data Poisoning Attacks to Locally Private Graph Learning Protocols}

\author{Longzhu He}
\affiliation{
  \institution{Beijing University of Posts and Telecommunications}
    \city{Beijing}
  \country{China}
}
\email{helongzhu@bupt.edu.cn}

\author{Chaozhuo Li}
\affiliation{
  \institution{Beijing University of Posts and Telecommunications}
    \city{Beijing}
  \country{China}
}
\email{lichaozhuo@bupt.edu.cn}

\author{Peng Tang}
\affiliation{
  \institution{Shandong University}
    \city{Qingdao}
  \country{China}
}
\email{tangpeng@sdu.edu.cn}

\author{Li Sun}
\affiliation{
  \institution{North China Electric Power University}
    \city{Beijing}
  \country{China}
}
\email{ccesunli@ncepu.edu.cn}

\author{Sen Su}
\authornote{Corresponding author.}
\affiliation{
  \institution{Beijing University of Posts and Telecommunications}
    \city{Beijing}
  \country{China}
}
\email{susen@bupt.edu.cn}

\author{Philip S. Yu}
\affiliation{
  \institution{University of Illinois Chicago}
    \city{Chicago}
  \country{USA}
}
\email{psyu@uic.edu}

\renewcommand{\shortauthors}{Trovato and Tobin, et al.}

\begin{abstract}
Graph neural networks (GNNs) have achieved significant success in graph representation learning and have been applied to various domains. However, many real-world graphs contain sensitive personal information, such as user profiles in social networks, raising serious privacy concerns when graph learning is performed using GNNs. To address this issue, locally private graph learning protocols have gained considerable attention. These protocols leverage the privacy advantages of local differential privacy (LDP) and the effectiveness of GNN's message-passing in calibrating noisy data, offering strict privacy guarantees for users' local data while maintaining high utility (\textit{e.g.}, node classification accuracy) for graph learning. Despite these advantages, such protocols may be vulnerable to \textit{data poisoning attacks}, a threat that has not been considered in previous research. Identifying and addressing these threats is crucial for ensuring the robustness and security of privacy-preserving graph learning frameworks. This work introduces the first data poisoning attack targeting locally private graph learning protocols. The attacker injects fake users into the protocol, manipulates these fake users to establish links with genuine users, and sends carefully crafted data to the server, ultimately compromising the utility of private graph learning. The effectiveness of the attack is demonstrated both theoretically and empirically. In addition, several defense strategies have also been explored, but their limited effectiveness highlights the need for more robust defenses.

\end{abstract}








\maketitle

\section{Introduction}\label{sec1}

Graph Neural Networks (GNNs) have achieved remarkable success in graph representation learning~\cite{hamilton2020graph} and have been widely applied in various domains, including graph mining~\cite{li2019graph,wu2022graph1}, recommendation systems~\cite{ying2018graph,wu2022graph}, and bioinformatics~\cite{fout2017protein,zhang2021graph}. However, most real-world graphs contain sensitive personal information, such as users' profiles and comments on social networks. Training GNNs on such graph data can raise serious privacy concerns, as malicious attackers can potentially infer sensitive information, such as user attributes, by interacting with the GNN model~\cite{wang2022group,zhang2022inference,meng2023devil}. Therefore, it is of particular significance to develop privacy-preserving GNN algorithms to safeguard user privacy.

 \begin{figure}
  \centering
  \includegraphics[width=0.95\linewidth]{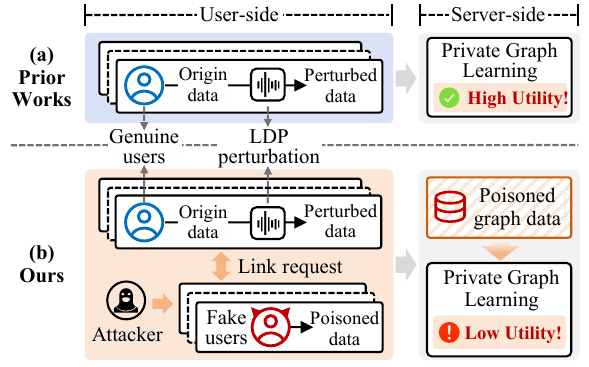}
  \vspace{-1.1em}
  \caption{Comparison of prior works with ours. (a) Prior work involves a cloud server and multiple genuine users, where users' sensitive data is perturbed with LDP to ensure privacy before being sent to the server for private graph learning. (b) In contrast, our work targets poisoning the private graph learning process. An attacker manipulates fake users, links them with genuine ones, and sends poisoned data to the server to compromise the utility of the learning process.}
  \vspace{-1em}
  \label{fig:1}
\end{figure}

Recently, locally private graph learning protocols have garnered significant attention from the security research community~\cite{sajadmanesh2021locally,lin2022towards,pei2023privacy,li2024privacy,qi2024linkguard,zhu2023blink,hidano2022degree,zhang2024locally}, as illustrated in Fig.~\ref{fig:1}(a). In these protocols, each user first perturbs their origin data locally using local differential privacy (LDP)~\cite{yang2024local}, typically through noise injection~\cite{dwork2006calibrating,wang2019collecting,sajadmanesh2021locally}, and then transmits the perturbed data to an untrusted third-party server that could potentially compromise the user's privacy. The server performs private graph learning on the collected noisy data. Given the superiority of LDP in protecting data privacy and the effectiveness of GNN's multi-hop message passing~\cite{feng2022powerful,abu2019mixhop} in calibrating noisy data, locally private graph learning protocols offer strict privacy guarantees for the user data while maintaining high utility for graph learning tasks. 

However, despite these advantages, these protocols can be vulnerable to \textit{data poisoning attacks} due to their distributed setup, which has not been considered in previous studies on locally private graph learning. In distributed privacy scenarios, although an attacker cannot take control of genuine users or directly compromise the graph database, it is easy to register fake accounts and create social identities with profiles. Consider a threat scenario as shown in Fig.~\ref{fig:1}(b), where an attacker aims to degrade the graph learning utility (\textit{e.g.}, node classification accuracy~\cite{DBLP:conf/iclr/KipfW17} or link prediction accuracy~\cite{zhang2018link}) of the original graph by injecting fake users. Specifically, the attacker first manipulates multiple fake users to be injected into the protocol, then sends friendship requests to genuine users, and finally transmits carefully crafted poisoned data to the server. Social users seeking social influence~\cite{carpenter2012narcissism,nitti2014friendship} are generally more inclined to accept friendship requests. As a result, some genuine users accept requests from attackers, allowing fake accounts to connect to genuine users and poison the original graph. When the GNN is trained on the poisoned graph data, malicious information spreads throughout the graph through message passing~\cite{feng2022powerful,abu2019mixhop}, significantly damaging the utility of privacy-preserving graph learning. In summary, this paper proposes a data poisoning attack to achieve the aforementioned objectives.

Unlike previous data poisoning attacks against GNNs~\cite{zou2021tdgia,zugner2018adversarial,sun2022adversarial,zugner2020adversarial}, achieving the above attack faces the following unique challenges. \ding{172} \textit{Strict data constraints}. Unlike traditional poisoning attacks that manipulate data without considering privacy concerns, this attack must be conducted within the LDP framework, which enforces specific rules for perturbing sensitive information. The challenge lies in balancing, on the one hand, the attacker’s need to carefully calibrate privacy-preserving perturbations to avoid detection and, on the other, the desire to maximize the compromise of private graph learning’s utility. This delicate balance makes the attack more difficult. \ding{173} \textit{Limited background knowledge}. In this attack scenario, a realistic black-box setup is considered, where the attacker has no access to information about the target node or its neighbors, nor any details about the architecture or parameters of the GNN model within the protocol. \ding{174} \textit{Complexity of the protocol}. Poisoning locally private graph learning protocols requires considering various LDP mechanisms (\textit{e.g.}, the piecewise mechanism~\cite{wang2019collecting,pei2023privacy}, the multi-bit mechanism~\cite{sajadmanesh2021locally,lin2022towards}, and the square wave mechanism~\cite{li2024privacy,li2020estimating}) and different GNN models (e.g., graph convolutional networks~\cite{DBLP:conf/iclr/KipfW17}, GraphSAGE~\cite{hamilton2017inductive}, and graph attention networks~\cite{DBLP:conf/iclr/VelickovicCCRLB18}). This complexity adds another layer of difficulty in ensuring the effectiveness and universality of the attack. In general, given these constraints, designing the data to be sent by the fake user to the server and determining how to connect with genuine nodes to maximize the attack’s effectiveness is a highly challenging task.

To address these challenges, we leverage a general framework for locally private graph learning protocols~\cite{sajadmanesh2021locally,lin2022towards,li2024privacy,pei2023privacy}
and identify potential attack surfaces by analyzing the phases most susceptible to manipulation. Focusing on the realism and universality of the attacks, we target the \textit{data perturbation} phase of the protocol. When carrying out the attack, for each genuine target node, we model the attack to maximally deviate from its original embedding (server-side aggregated vectors) by crafting the features of the fake nodes connected to it. To maintain the stealthiness of the attack, it is crucial that the crafted malicious node features adhere to the specific output rules of the LDP mechanism~\cite{sajadmanesh2021locally,pei2023privacy,li2024privacy}. With this limitation, we achieve the goal by selecting \textit{extreme feature values}. Additionally, we further enhance the attack's effectiveness by adjusting the links between the fake nodes to circumvent detection and increase disruption to the original graph.

The effectiveness of the attack is evaluated both theoretically and empirically. Theoretically, the error introduced by the poisoning attack on the embedding of target nodes is analyzed, and its impact on the entire graph is further assessed. Empirically, evaluations are conducted on six real-world datasets, applying three de facto LDP mechanisms~\cite{sajadmanesh2021locally,pei2023privacy,li2024privacy} and three GNN models~\cite{DBLP:conf/iclr/KipfW17,hamilton2017inductive,DBLP:conf/iclr/VelickovicCCRLB18}. Our empirical results confirm the effectiveness of our attacks. In addition, several potential defenses are explored, including graph homophily analysis~\cite{chenunderstanding, zhang2020gnnguard}, anomaly node detection~\cite{ma2021comprehensive}, and others. However, the limited effectiveness of these defenses highlights the urgent need for more robust countermeasures against our attack. Our contributions are summarized as follows.

\begin{itemize}[leftmargin=*, itemindent=0em]
    \item To the best of our knowledge, this is the first study of data poisoning attacks on locally private graph learning protocols. 
    \item We focus on the perturbation phase of the protocol and achieve a data poisoning attack, which aims to damage the learning utility.
    \item Through both theoretical analysis and empirical evaluation, it is demonstrated that our attacks can effectively degrade the utility of privacy-preserving graph learning, including tasks such as node classification and link prediction. 
    \item Several defense strategies are also explored to mitigate these attacks, with empirical results emphasizing the urgent need for more robust defensive countermeasures.
\end{itemize}
\textbf{Organization}. The rest of this paper is organized as follows. Sec.~\ref{sec:pre} introduces the problem definition and provides essential background knowledge. Sec.~\ref{33333333} presents the threat model and a detailed explanation of our proposed attack approach. Theoretical analysis is provided in Sec.~\ref{sec5}, while Sec.~\ref{sec:exp} empirically evaluates the effectiveness of the attack. Sec.~\ref{sec:def} explores various defense countermeasures. Sec.~\ref{sec:rek} reviews related literature, and Sec.~\ref{sec:con} concludes the paper.

\section{Preliminaries}\label{sec:pre}
Before introducing our poisoning algorithm, we first define the problem (Sec. \ref{2.1}) and outline the essential background on LDP (Sec. \ref{2.2}) and locally private graph learning protocols (Sec. \ref{2.3}).

\vspace{-0.2em}
\subsection{Problem Definition}\label{2.1}
The node data privacy problem is considered in the context of GNN-based graph learning. Formally, let $\mathcal{G}=(\mathcal{V},\mathcal{E})$ represent a graph, where $\mathcal{V}$ is the set of nodes and $\mathcal{E}$ is the set of edges. $\mathbf{X}\in\mathbb{R}^{|\mathcal{V}|\times d}$ is the set of node feature vectors. Each node/user\footnote{Note that in many real-world applications, graph nodes correspond to human users, and therefore, the terms “node” and “user” are used interchangeably in this paper.} $v\in\mathcal{V}$ has a $d$-dimensional feature vector $\mathbf{x}_v\in[\alpha,\beta]^d$, which contains sensitive information about the user. Uploading it directly to an untrusted server for GNN-based graph learning poses a significant privacy risk~\cite{zhang2022inference,meng2023devil,zhang2024survey}. Locally private graph learning protocols~\cite{sajadmanesh2021locally,lin2022towards,li2024privacy,pei2023privacy} aim to ensure node privacy while enabling effective graph learning through LDP~\cite{yang2024local}. This paper proposes poisoning a series of locally private graph learning protocols to expose their potential security vulnerabilities, thereby promoting more secure and robust privacy-preserving graph learning frameworks.

\vspace{-0.2em}
\subsection{Local Differential Privacy}\label{2.2}

LDP is widely adopted in decentralized data collection and distribution scenarios~\cite{yang2024local,cormode2018privacy,kairouz2016discrete,kairouz2014extremal,wang2019collecting,wang2019answering} due to its ability to protect individual privacy while enabling meaningful data analysis. It has seen widespread application in industry, including but not limited to tech giants like Google~\cite{erlingsson2014rappor}, Microsoft~\cite{ding2017collecting}, and Apple~\cite{apple2017learning}. By introducing random noise into the computation process, LDP provides robust privacy guarantees for the original data. Specifically, LDP employs a randomized perturbation mechanism $\mathcal{M}$ to perturb the user's original data before transmitting the perturbed version to the server. Formally, the definition of $\epsilon$-LDP is presented as follows.
\begin{definition}[$\epsilon\text{-LDP}$]\label{def:1}
A random perturbation mechanism $\mathcal{M}$ satisfies $\epsilon\text{-LDP}$, where $\epsilon > 0$, if and only if for any user's private data $x$ and $x^\prime$, and for all possible outputs $y \in Range(\mathcal{M})$, we have:
	\begin{equation}
		\Pr[\mathcal{M}(x) = y] \le e^\epsilon \cdot \Pr[\mathcal{M}(x^\prime) = y],
	\end{equation}
where the parameter $\epsilon$, known as the “\textit{privacy budget}”, serves to balance utility and privacy. A higher value of $\epsilon$ corresponds to weaker privacy guarantees but improved learning utility.
\end{definition}

\vspace{-0.7em}

\subsection{Locally Private Graph Learning Protocol}\label{2.3}
The locally private graph learning protocol $\Pi$ comprises three successive phases: \textit{perturbation}, \textit{calibration}, and \textit{GNN-based learning}. 

\subsubsection{Perturbation}\label{2.3.1}
In this phase, each user $v\in\mathcal{V}$ applies the LDP mechanism $\mathcal{M}$ to perturb their sensitive node features $\mathbf{x}_v\in\mathbb{R}^d$ to ensure privacy, formalized as $\mathbf{x}_v^\prime \leftarrow \mathcal{M}(\mathbf{x}_v,\epsilon) $. The primary LDP mechanisms used include the piecewise mechanism (PM)~\cite{wang2019collecting,pei2023privacy}, the multi-bit mechanism (MB)~\cite{sajadmanesh2021locally,lin2022towards}, and the square wave mechanism (SW)~\cite{li2024privacy,li2020estimating}. Given the high-dimensional nature of node features, these LDP mechanisms follow a structured approach to balance privacy protection and data utility. The general perturbation process consists of two steps: \ding{172} the $m\in\{1,2,\cdots,d\}$ values are randomly selected from the $d$-dimensional space. \ding{173} An $\epsilon/m$-LDP noise perturbation is applied to each of the sampled dimensions, while the remaining $d-m$ dimensions are set to
zero. Specifically,

\vspace{-0.2em}

\begin{itemize}[leftmargin=*, itemindent=0em]
    \item In the one-dimensional multi-bit mechanism~\cite{sajadmanesh2021locally,lin2022towards}, for any original input data $x\in[\alpha, \beta]$, the perturbed value $x^\prime$ falls within the range $\{-1, 1\}$ and is sampled from the following distribution:
    \begin{equation}
\mathrm{Pr}[x^\prime=c|x] =\begin{cases}
  \frac{1}{e^\epsilon +1}+\frac{x-\alpha}{\beta-\alpha}\cdot \frac{e^\epsilon -1}{e^\epsilon +1},  & \text{ if } c= 1\\
  \frac{e^\epsilon}{e^\epsilon +1}-\frac{x-\alpha}{\beta-\alpha}\cdot \frac{e^\epsilon -1}{e^\epsilon +1},& \text{ if } c=-1
\end{cases}.
\end{equation}
\item In the one-dimensional piecewise mechanism~\cite{wang2019collecting,pei2023privacy}, the input domain is $[\alpha, \beta]$, and the perturbed data range is $[-s,s]$, where $s=\frac{e^{\epsilon/2}+1}{e^{\epsilon/2}-1}$. Given an original value $x$, the perturbed value $x^\prime$ is sampled from the following probability density function:
\begin{equation}
	\mathrm{Pr}[x^\prime=c|x] = 
	\begin{cases}
		p,   &\text{if } c\in[l(x), r(x)]  \\
		p/e^\epsilon, &\text{if } c\in[-s, l(x))\cup (r(x), s] 
	\end{cases},
\end{equation}
where $p=\frac{e^\epsilon -e^{\epsilon/2 }}{2e^{\epsilon/2 }+2}$, $l(x)=\frac{s+1}{2} \cdot x-\frac{s-1}{2}$, and $r(x)=l(x)+s-1$.
\item In the one-dimensional square wave mechanism~\cite{li2024privacy,li2020estimating}, the perturbed data range is $[-b-1,b+1]$, where $b=\frac{\epsilon e^\epsilon-e^\epsilon+1}{e^\epsilon(e^\epsilon-\epsilon-1)}$. The noisy output $x^\prime$ is sampled from the following distribution:
\begin{equation}
	\mathrm{Pr}[x^\prime=c|x]\hspace{-0.2em} = \hspace{-0.2em}
	\begin{cases}
		p,   \hspace{-0.7em}&\text{if } c\in[x-b, x+b]  \\
		p/e^\epsilon, \hspace{-0.7em}&\text{if } c\in[-b-1, x-b)\cup (x+b, 1+b] 
	\end{cases},
\end{equation}
where $p=e^\epsilon/(2be^{\epsilon }+2)$. More details on this phase, see App.~\ref{app2}.
\end{itemize}

\subsubsection{Calibration}
In this phase, for node $v\in\mathcal{V}$, the server calibrates the noise by invoking the following $\textsc{Aggregate}(\cdot)$ function iteratively over the features of neighboring nodes up to $K$ steps away, without applying any non-linear transformation in between:
\begin{equation}
\widehat{\mathbf{h}}^{(k)}_{v}=\textsc{Aggregate}_k(\{\widehat{\mathbf{h}}_u^{(k-1)}, \forall u \in \mathcal{N}(v)\}),\label{hNv}
\end{equation}
where \( \mathbf{h}_v^{(k)} \) is the embedding of node \( v \) at layer \( k \), \( \mathcal{N}(v) \) is the set of neighbors of \( v \), and \( \textsc{Aggregate}_k(.) \) is a differentiable, permutation-invariant aggregation function (such as sum or mean) at layer \( k \). When initialized, we have \( \widehat{\mathbf{h}}_v^{(0)} = \mathbf{x}_v' \), \textit{i.e.}, the initial embedding of node \( v \) is its perturbed feature vector \( x_v^\prime \).

\vspace{-0.2em}

\subsubsection{GNN-based learning}
In this phase, the server performs graph learning tasks such as node classification~\cite{wu2019net}, by training GNN models \( f_{\theta} \), such as graph convolutional networks (GCN)~\cite{DBLP:conf/iclr/KipfW17}, GraphSAGE~\cite{hamilton2017inductive}, and graph attention networks (GAT)~\cite{DBLP:conf/iclr/VelickovicCCRLB18}, using node embeddings \( \widehat{\mathbf{h}}_v^K \) obtained from the previous calibration phase.

In summary, the locally private graph learning protocol is defined as $\Pi=(\mathcal{M},\mathcal{A},f_\theta)$, where $\mathcal{M}$, $\mathcal{A}$, and
$f_\theta$ correspond to the three phases described above. Due to the black-box setup considered in this paper, the attacker cannot directly manipulate the $\mathcal{A}$ and
$f_\theta$ phases on the server side. Therefore, the poisoning attack is carried out by targeting the $\mathcal{M}$ phase, located on the user side.

\section{Methodology}\label{33333333}
\begin{figure}
  \centering
\includegraphics[width=\linewidth]{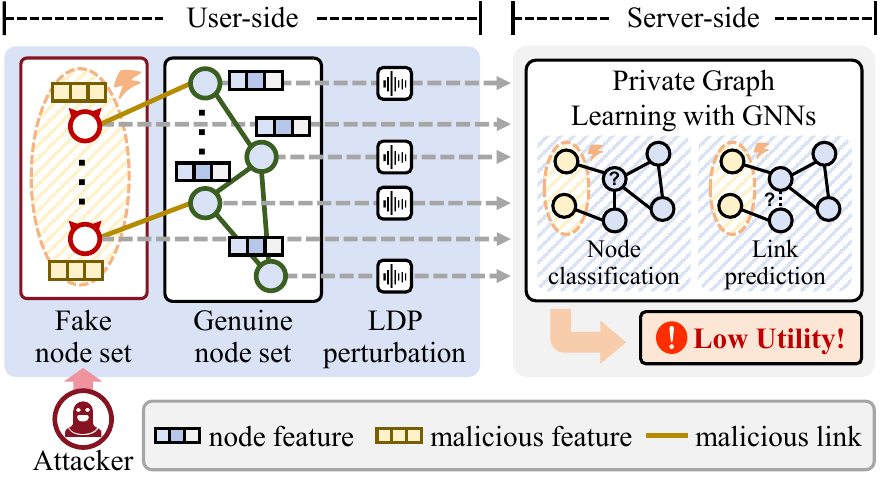}
  \caption{Overview of the proposed data poisoning attacks. The attacker injects multiple fake nodes, establishes links with genuine target nodes, and crafts malicious features for the fake nodes, aiming to maximally degrade the utility of private graph learning, such as node classification accuracy.}
  \vspace{-0.5em}
  \label{fig2}
\end{figure}

In this section, we first present the threat model in Sec.~\ref{sec:threat}, followed by a detailed description of the data poisoning attacks proposed against protocol $\Pi$ in Sec.~\ref{sec:4}. Fig.~\ref{fig2} illustrates the attack process.
\subsection{Threat Model}\label{sec:threat}
\subsubsection{Attacker’s capability} 
We assume that the malicious attacker can inject multiple fake nodes \(v_{\text{atk}} \in \mathcal{V}_{\text{atk}}\) into the original graph \( \mathcal{G} = (\mathcal{V}, \mathcal{E}) \), and establish malicious links between the target node \(v_t\in\mathcal{V}\) and the fake nodes, as well as between the fake nodes themselves. The sets of fake and target nodes are defined as \( \mathcal{V}_{\text{atk}} \) and \( \mathcal{V}_t \subset\mathcal{V}\), respectively, with \( 1<|\mathcal{V}_{\text{atk}}| \le|\mathcal{V}_t|\ll |\mathcal{V}| \). The set of malicious links injected is defined as $\mathcal{E}_{\text{atk}}$. Considering the realism and subtlety of the attack, \textit{only one} malicious link is allowed to connect to each target node $v_t$. Furthermore, for each fake node, the attacker has the freedom to specify its graph data.

\vspace{-0.2em}

\subsubsection{Attacker’s background knowledge.} We assume that the attacker knows the predefined parameters of the LDP mechanisms, as these are publicly available to all users. In addition, we consider a practical black-box scenario in which the attacker has \textit{no access to} any information about the target node, including its features, labels, or neighborhood information, as well as any details about the architecture or parameters of the GNN model in protocol $\Pi$.
\vspace{-0.2em}
\subsubsection{Attacker’s goal.} 
The attacker's goal is to deceive the GNN model $f_\theta$ in protocol $\Pi$ by strategically crafting malicious fake nodes and malicious edges. This attack aims to compromise the utility of all target nodes while also further degrading the overall learning utility of the entire graph, such as reducing node classification accuracy~\cite{DBLP:conf/iclr/KipfW17}, by exploiting the message-passing mechanism of GNNs. Formally, the attacker’s objective is to
\vspace{-0.1em}
\begin{equation}
\min \mathcal{L}_{\text{atk}}  \left (f_{\theta }(\mathcal{G}^\prime)  \right ), \quad\text{subject to} \left \| \mathcal{G}^\prime -  \mathcal{G} \right \|_p\le \bigtriangleup,   
\label{eq6}
\end{equation}
where $\mathcal{L}_{\mathrm{atk}}$ can be defined as $-\mathcal{L}_{\mathrm{train}}$. $\mathcal{L}_{\mathrm{train}}$ denotes the classification loss (e.g., cross-entropy), which measures how well the GNN model performs. $\mathcal{G}^\prime=(\mathcal{V}\cup \mathcal{V}_{\text{atk}},\mathcal{E}\cup \mathcal{E}_{\text{atk}})$ is the poisoned graph with fake node set $\mathcal{V}_{\text{atk}}$ and malicious link set $\mathcal{E}_{\text{atk}}$. $\bigtriangleup$ denotes the allowed perturbation or stealthiness of the attack. $\left \| \cdot\right \|_p$ defines the distance measure between the original and modified graphs.

\subsection{Attack Details}\label{sec:4}

\subsubsection{Crafting malicious local graph data}\label{4.1}
In implementing the poisoning attack, we link a single fake node \( v_{\text{atk}} \) to each target node \( v_t \). The attacker's goal is to compromise the post-aggregation embedding of node \( v_t \) by crafting the features \( \mathbf{x}^\prime_{v_{\text{atk}}} \) of node \( v_{\text{atk}} \), which subsequently degrades the utility of privacy-preserving graph learning. However, the design of \( \mathbf{x}^\prime_{v_{\text{atk}}} \) must adhere to the output constraints \( \mathbb{R} \) of a specific LDP mechanism~\cite{sajadmanesh2021locally,wang2019collecting,li2024privacy} for stealth purposes. Thus, the challenge lies in refining the design of \( \mathbf{x}^\prime_{v_{\text{atk}}} \) to maximally undermine the embedding of the node \( v_t \) while satisfying the constraints \( \mathbb{R} \). The attack objectives are as follows.

\begin{definition}[Attack Objective]
For any target node $v_t\in\mathcal{V}_t$, the attacker aims to maximize the deviation from its embedding $\widehat{\mathbf{h}}_{\mathcal{N}(v_t)}$ by manipulating the feature $\mathbf{x}^\prime_{v_\text{atk}}=[x^\prime_i]_{i=1}^d$ of fake nodes  $v_\text{atk}$, \textit{i.e.},
\begin{equation}
\max_{\mathbf{x}^\prime_{v_\text{atk}}} \left\| \widehat{\mathbf{h}}_{\mathcal{N}(v_t)\cup\{v_{\text{atk}}\} }\hspace{-0.2em}-\hspace{-0.2em}\widehat{\mathbf{h}}_{\mathcal{N}(v_t)} \right\|,\,\text{s.t.}\,\, \mathbb{R}: x^\prime_i\sim  \begin{cases}
 [-B,B], & \hspace{-0.4em}i\in\mathcal{S}  \\
 0, & \hspace{-0.4em}i\notin \mathcal{S}
\end{cases},
  \end{equation}
where $\widehat{\mathbf{h}}_{\mathcal{N}(v_t)\cup\{v_{\text{atk}}\} }$ is the embedding of node $v_t$ after the attack. $B$ denotes the perturbation boundary, which takes the values $\frac{(e^{\bar{\epsilon}/2}+1)d}{e^{\bar{\epsilon}/2}-1}$, $\frac{d(\beta-\alpha )}{2m} \frac{e^{\bar{\epsilon}}+1}{e^{\bar{\epsilon}}-1} +\frac{\alpha+\beta}{2}$ and $\frac{\bar{\epsilon}e^{\bar{\epsilon}}-e^{\bar{\epsilon}}+1}{e^{\bar{\epsilon}}(e^{\bar{\epsilon}}-\bar{\epsilon}-1)}$ (where $\bar{\epsilon}=\epsilon/m$) for PM~\cite{wang2019collecting}, MB~\cite{sajadmanesh2021locally} and SW~\cite{li2024privacy}, respectively. $\mathcal{S}$ represents a set of $m$ values drawn uniformly at random without replacement from $\{1,2,\cdots,d\}$. To achieve the above objective, the feature vector of the fake node $v_\text{atk}$ should be chosen to be as dissimilar as possible to the target node $v_t\in\mathcal{V}_t$, thus maximizing the Euclidean distance between the original and modified embeddings. This can be accomplished by selecting \textit{extreme feature values} according to the output of the private graph learning LDP protocol. Therefore, $\mathbf{x}^\prime_{v_\text{atk}}$ is generated as follows: first, randomly select the $m$ dimensions from the $d$-dimensional space and assign the data of the selected dimension as $B\cdot(-1)^{\mathbf{1}(\mathcal{U}>0.5)}$, where $\mathcal{U}\sim \mathrm{Uniform}(0,1)$; then, the remaining $d-m$ dimensions are set to zero. Furthermore, in addition to the node features, we randomly assign a label to each fake node.
\end{definition}

\vspace{-1em}

\subsubsection{Linking fake nodes to target nodes}\label{4.2}
Consider the set of target nodes \( \mathcal{V}_t \) and the set of fake nodes \( \mathcal{V}_{\text{atk}} \), where $|\mathcal{V}_{\text{atk}}| \le|\mathcal{V}_t|$. During the linking process between the target nodes and fake nodes, we propose a cyclic matching strategy \( \mathbb{M} \) to ensure that each target node is assigned a fake node. Under this strategy, the fake nodes are recycled, guaranteeing that all target nodes can be successfully matched. For any node \( v_t \in\mathcal{V}_t\),
\begin{equation}
    \mathbb{M}: \mathcal{V}_t\to\mathcal{V} _{\text{atk}}, \,\,\mathrm{where}\,\,\,  \mathbb{M}(v_t)=(v_{\mathrm{atk}})_{(j\,\mathrm{mod} \,|\mathcal{V}_{\mathrm{atk}}|+1)},
\end{equation}
where \( \mathbb{M}(v_t) \) denotes the fake node to which the target node \( v_t \) is matched, while \( j \) is the index of the target node \( v_t\), with values ranging from \( 1 \leq j \leq |\mathcal{V}_t| \). For each node \( v_t \), we ensure that the indexes of the fake nodes are recycled within the range \([1, |\mathcal{V}_{\text{atk}}|]\) using the modulo operation. This ensures that fake nodes are duplicatedly assigned when there are insufficient fake nodes. This matching mechanism guarantees that all target nodes are matched by effectively utilizing the fake nodes, maximizing their usage when their number is limited, and thereby allowing the attacker to achieve the greatest coverage effect with constrained resources.

\subsubsection{Constructing inner links of fake nodes}
\label{4.3}
From the perspective of attack stealthiness, we aim to maintain the average degree of the poisoned graph $\mathcal{G}^\prime$ equal to that of the original graph $\mathcal{G}$ to evade potential homomorphism detection defenses. To achieve this, we connect the $|\mathcal{V}_{\text{atk}}|$ fake nodes with a probability $q$ between them. By controlling the connection probability between the fake nodes, we ensure that the newly added nodes do not significantly alter the global degree distribution of the graph, thereby preserving the original structure of the graph as much as possible in terms of statistical features. Choosing an appropriate connection probability $q$ is crucial, and we define $q$ as follows.
\begin{prop}\label{prop1}
Let $\langle d_{\mathrm{orig.}}\rangle $ represent the average degree of the original graph $\mathcal{G}$. After adding $|\mathcal{V}_{\text{atk}}|$ fake nodes, the probability $q$ of linking these fake nodes to each other is
\begin{equation}\label{9999}
q=\frac{\langle d_{\mathrm{orig.}}\rangle-2\cdot|\mathcal{V}_t|/|\mathcal{V}_{\text{atk}}|}{|\mathcal{V}_{\text{atk}}|-1},  
\end{equation}
where $|\mathcal{V}_{\text{atk}}|>2|\mathcal{V}_t|/\langle d_{\mathrm{orig.}}\rangle$, $|\mathcal{V}_t |\ge (\langle d_{\mathrm{orig.}}\rangle+1)^2/8$, and $\langle d_{\mathrm{orig.}}\rangle\ge2$. These conditions are easy to fulfill in realistic scenarios. Please refer to App.~\ref{app2.1} for the proof and further explanation of Eq.~(\ref{9999}).
\end{prop}

\subsubsection{Optimizing the features of all fake nodes}
\label{4.4}
Note that further optimizing the malicious feature vectors of all fake nodes from a global perspective is beneficial to maximize the damage to the target nodes. Specifically, as described in Sec.~\ref{4.1}, the target node $v_t\in\mathcal{V}_t$ is linked to a fake node. Furthermore, as detailed in Sec.~\ref{4.3}, this fake node is connected to multiple other fake nodes with the estimation of $q\cdot(|\mathcal{V}_{\text{atk}}|-1)$. In this topology, crafting malicious node features in a way that maximizes the impact of the attack is crucial. To this end, we consider three ways of crafting the non-zero bits in the malicious node features: \ding{172} \textit{Random}. The non-zero bits in the features of all fake nodes are randomly selected. \ding{173} \textit{Diverse}. The non-zero bits in the features of all fake nodes differ from each other, which is formalized as $\mathcal{S}_v\cap  \mathcal{S}_u=0$, for $v,u\in \mathcal{V}_{\text{atk}}$. \ding{174} \textit{Identical}. The non-zero bits in the features of all fake nodes are identical, formalized as $\mathcal{S}_v=\mathcal{S}_u$, for $v,u\in \mathcal{V}_{\text{atk}}$.  The analysis in Sec.~\ref{sec5.1} demonstrates that the “Identical” feature setting maximizes the attack's effectiveness, inflicting the most damage on the target node. We adopt this setting in subsequent experiments.

\section{Theoretical Analysis}\label{sec5}
In this section, we successively analyze the error  propagation affecting the target nodes and the entire graph in Sec.~\ref{sec5.1} and Sec.~\ref{sec5.2}, and finally discuss the security-privacy trade-off in Sec.~\ref{sec5.3}.

\subsection{Error Analysis of Target Nodes}\label{sec5.1}
In \( \Pi \), for each target node \( v_t \), the two properties—\textit{unbiased expectation} (Def.~\ref{def33}) and \textit{variance minimization}—ensure optimal perturbation of the node features, forming the basis for high-utility private graph learning. However, the injection of fake nodes disrupts this balance, may introducing biased expectations and increasing variance, ultimately degrading the utility of the learned representations.
\begin{definition}[Unbiased Expectation]\label{def33}
In $\Pi$, the perturbed node features remain unbiased, \textit{i.e.}, $\mathbb{E}[\mathbf{x}^\prime  ] =\mathbf{x}$. Furthermore, the subsequent aggregator defined by Eq.~(\ref{hNv}) also provides an unbiased estimate.
\end{definition}
As in Prop.~\ref{prop2}, we quantify the bias \( \Delta_\mathbb{E}\) and \(  \Delta_{\text{Var}}  \) introduced by the expectation and variance of our attack on the target node. Through this analysis, we demonstrate that a \textit{identical} approach to crafting malicious features facilitates maximum damage to the target node.
\begin{prop}\label{prop2}
Considering two-layer aggregation, our attack does not introduce an expectation bias. However, it causes a variance bias $\Delta_{\text{Var}} $. We quantify the expected value of the variance of the aggregated embedding before and after the attack in App.~\ref{app2.2}. 
\end{prop}

\subsection{Error Analysis of the Entire Graph}\label{sec5.2}
After injecting fake nodes and propagating malicious information through the $K$-layer aggregation process, our attack severely distorts the node representation in the entire graph $\mathcal{G}=(\mathcal{V},\mathcal{E})$. To quantify this distortion, we define the global error energy function~\cite{ortega2018graph} \( \Psi(\mathcal{G}) \) before the attack as follows:
\begin{equation}
\Psi(\mathcal{G})=  \textstyle \sum_{v\in\mathcal{V}}  \| \mathbf{h}_v^{(K)}-\mathbf{x}_v  \|^2+\lambda \sum_{(u,v)\in \mathcal{E}}\| \mathbf{h}_u^{(K)}-\mathbf{h}_v^{(K)}    \|^2, 
\end{equation}
where \(\lambda \) represents the aggregation balancing parameter. For the graph after the attack, denoted $\mathcal{G}^\prime$, the global error energy function is defined as \( \Psi(\mathcal{G}^\prime).\)  Furthermore, the global error increment is quantified as $\Delta\Psi= \Psi(\mathcal{G}^\prime )-\Psi(\mathcal{G})$ and the expected deviation is
\begin{equation}\label{1111}
\mathbb{E}[ \Delta\Psi]=\left ( 1+\lambda q/(1-\lambda q) \right ) \cdot |\mathcal{V}_{\mathrm{atk} } |\cdot B^2\cdot K,
\end{equation}
where \( |\mathcal{V}_{\mathrm{atk} } | \) is the number of injected fake nodes. The result indicates that the perturbation impact is linearly related to \( |\mathcal{V}_{\mathrm{atk} } | \). Furthermore, higher values of \( B \) and \( K \) exacerbate the adversarial impact on the graph. This reveals the fundamental vulnerability of GNNs in our poisoning attack. More details on Eq.~(\ref{1111}) in App.~\ref{b.3}.
\subsection{Security-Privacy Trade-Off}\label{sec5.3}
Additionally, we observe a security-privacy trade-off in our attack on protocol $\Pi$, highlighting the inherent tension between enhancing privacy protection and maintaining robustness against poisoning attacks. As the privacy budget decreases, the level of noise injection increases, reducing the model's sensitivity to individual data points. However, this also amplifies the impact of adversarial perturbations. Mathematically, this relationship can be expressed as $\mathbb{E}[ \Delta\Psi]\propto e^{-\epsilon }$, where privacy budget $\epsilon$ controls the level of privacy, and a lower $\epsilon$ exacerbates the adversarial impact. More details on this phenomenon can be found in the experiments in Sec.~\ref{sec6.2}.

\begin{figure*}[htp]
    \centering
    \hspace{-0.3cm}\raisebox{-1mm}{\rotatebox{90}{targeted}} 
    \hspace{-0.4cm} 
    \begin{minipage}{\textwidth}
        \centering
        \hspace{0.4cm}\subfigure[\normalsize Cora]{\includegraphics[width=0.15\textwidth]{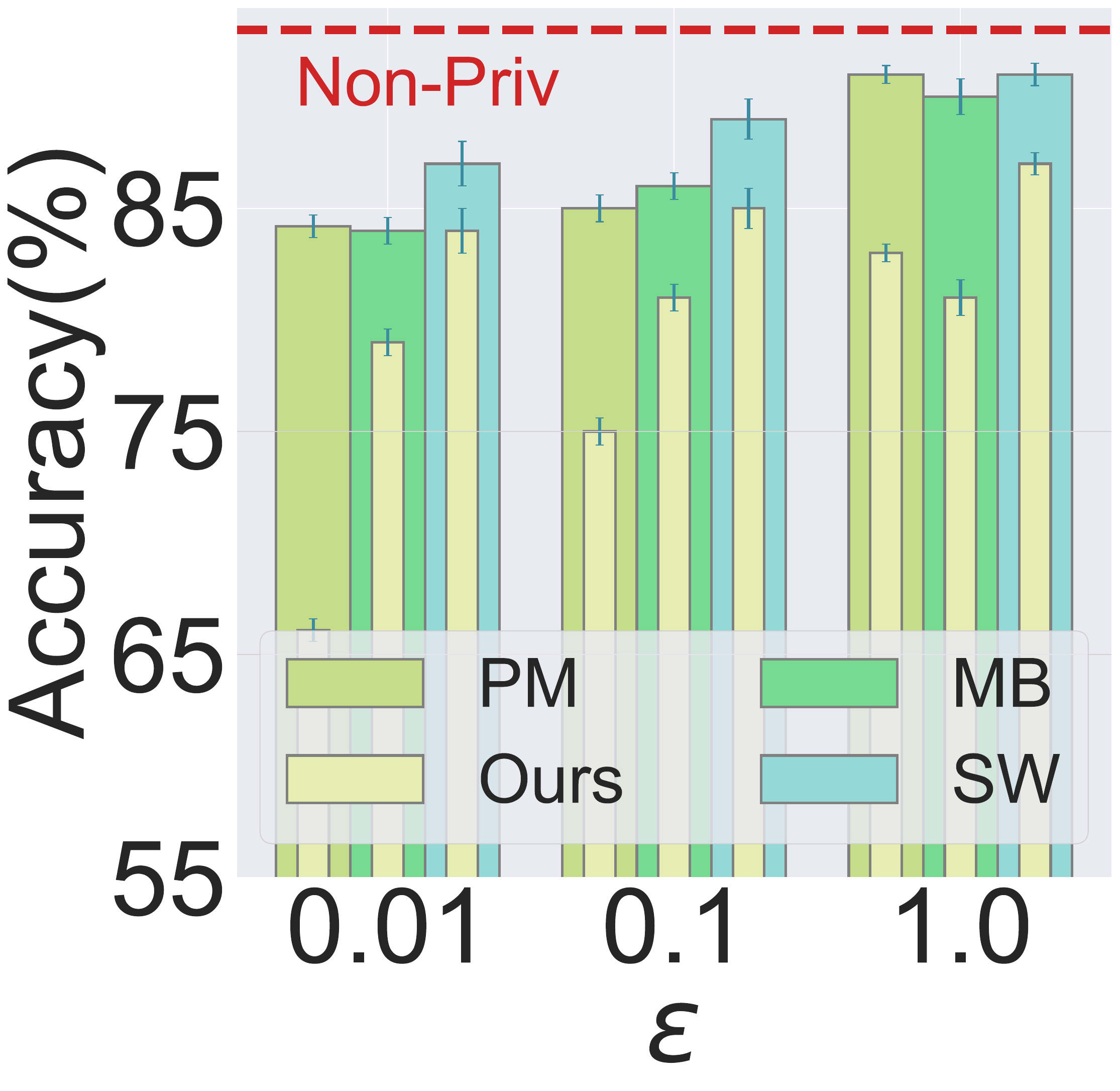}}
        \hspace{0.2cm}\subfigure[\normalsize Citeseer]{\includegraphics[width=0.15\textwidth]{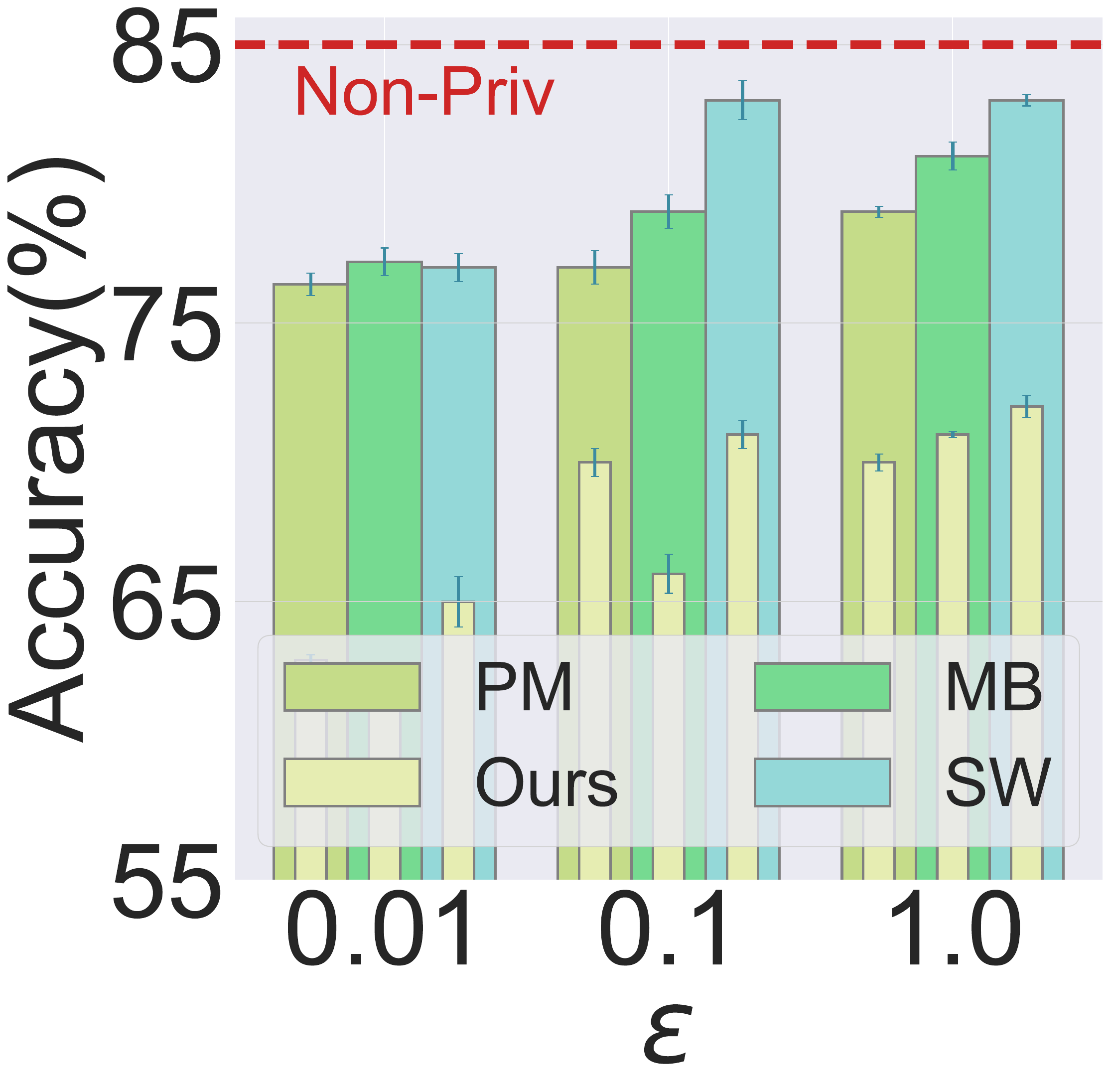}}
        \hspace{0.2cm}\subfigure[\normalsize Pubmed]{\includegraphics[width=0.15\textwidth]{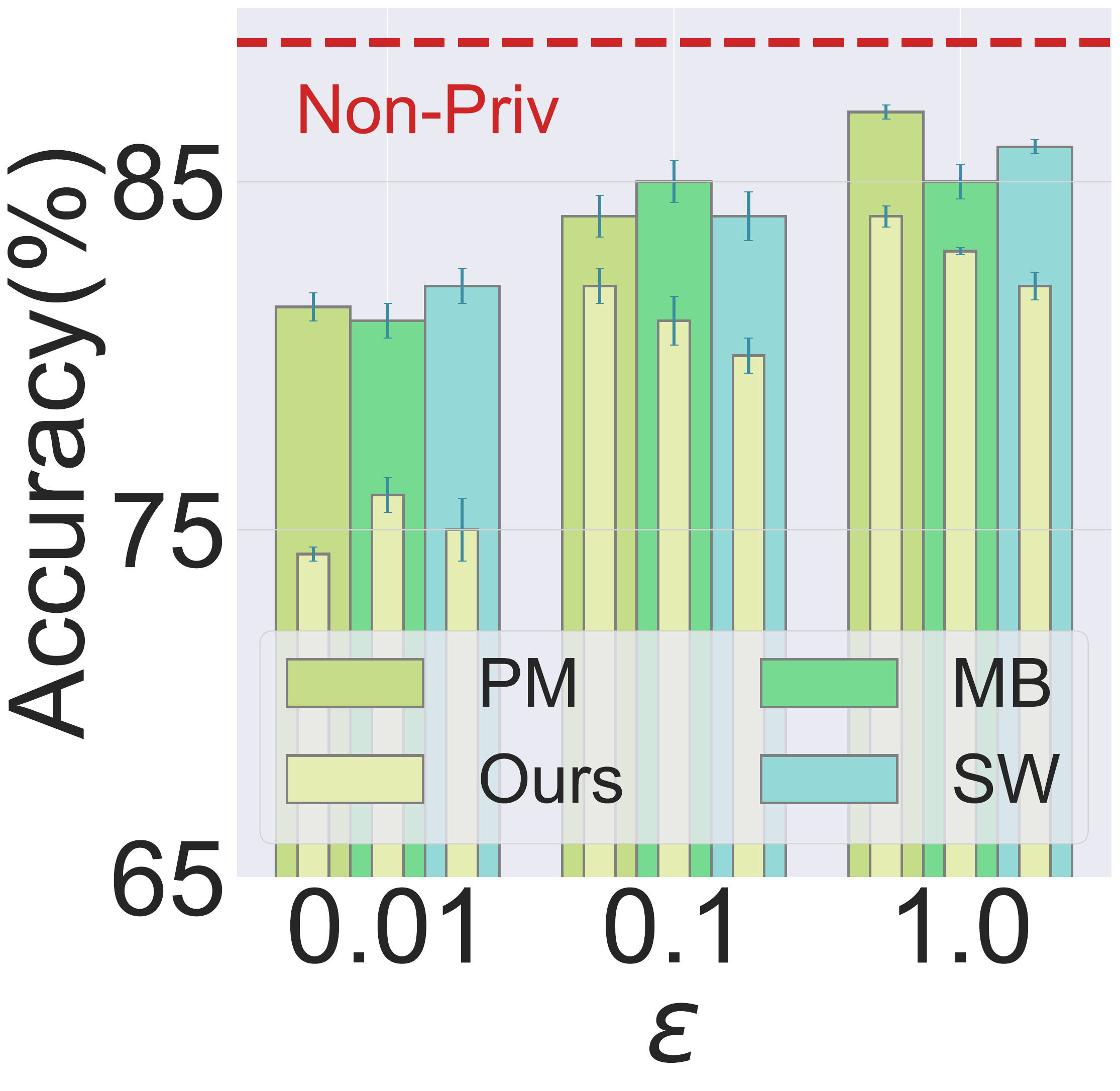}}
        \hspace{0.2cm}\subfigure[\normalsize LastFM]{\includegraphics[width=0.15\textwidth]{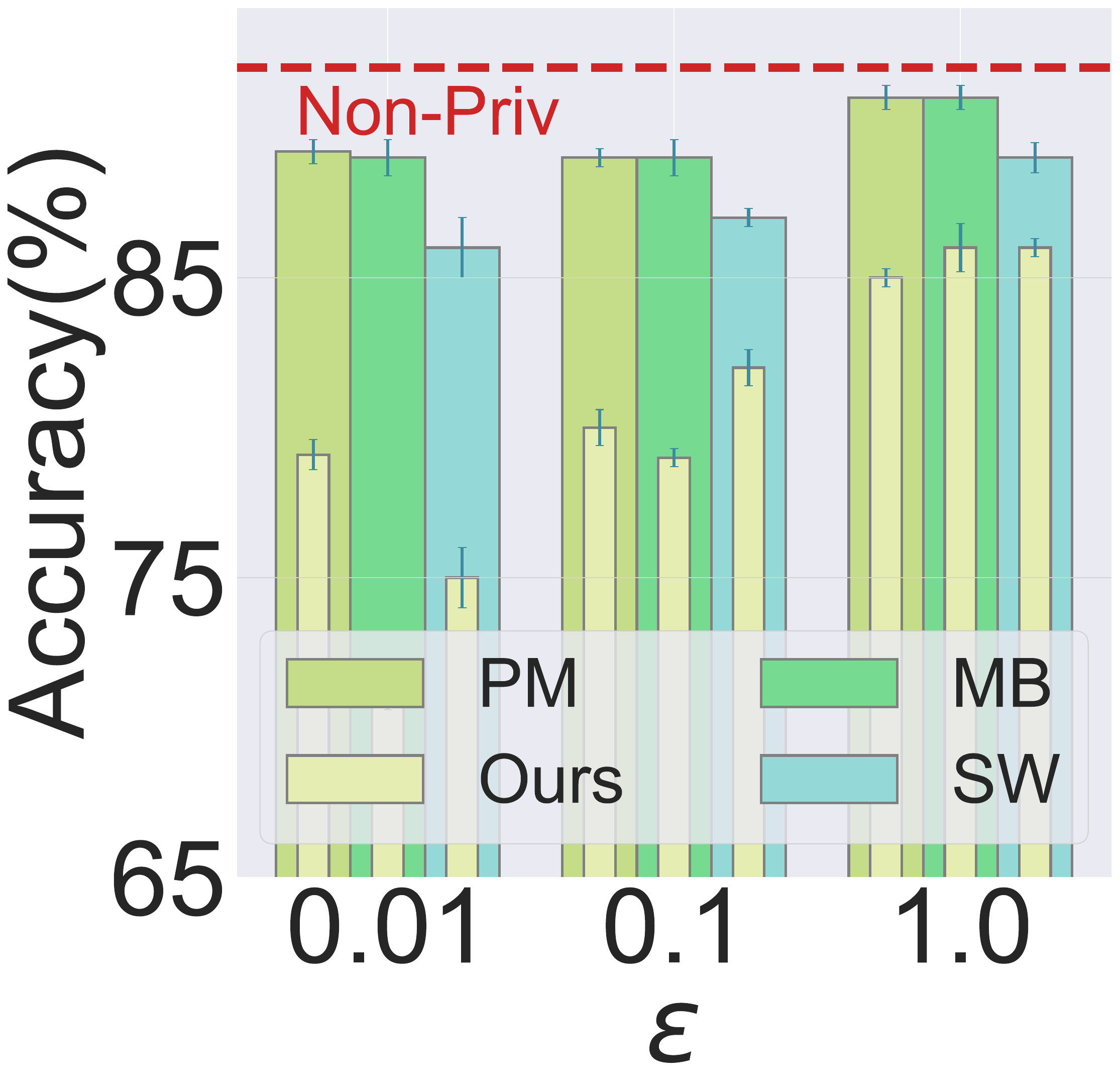}}
        \hspace{0.2cm}\subfigure[\normalsize Twitch]{\includegraphics[width=0.15\textwidth]{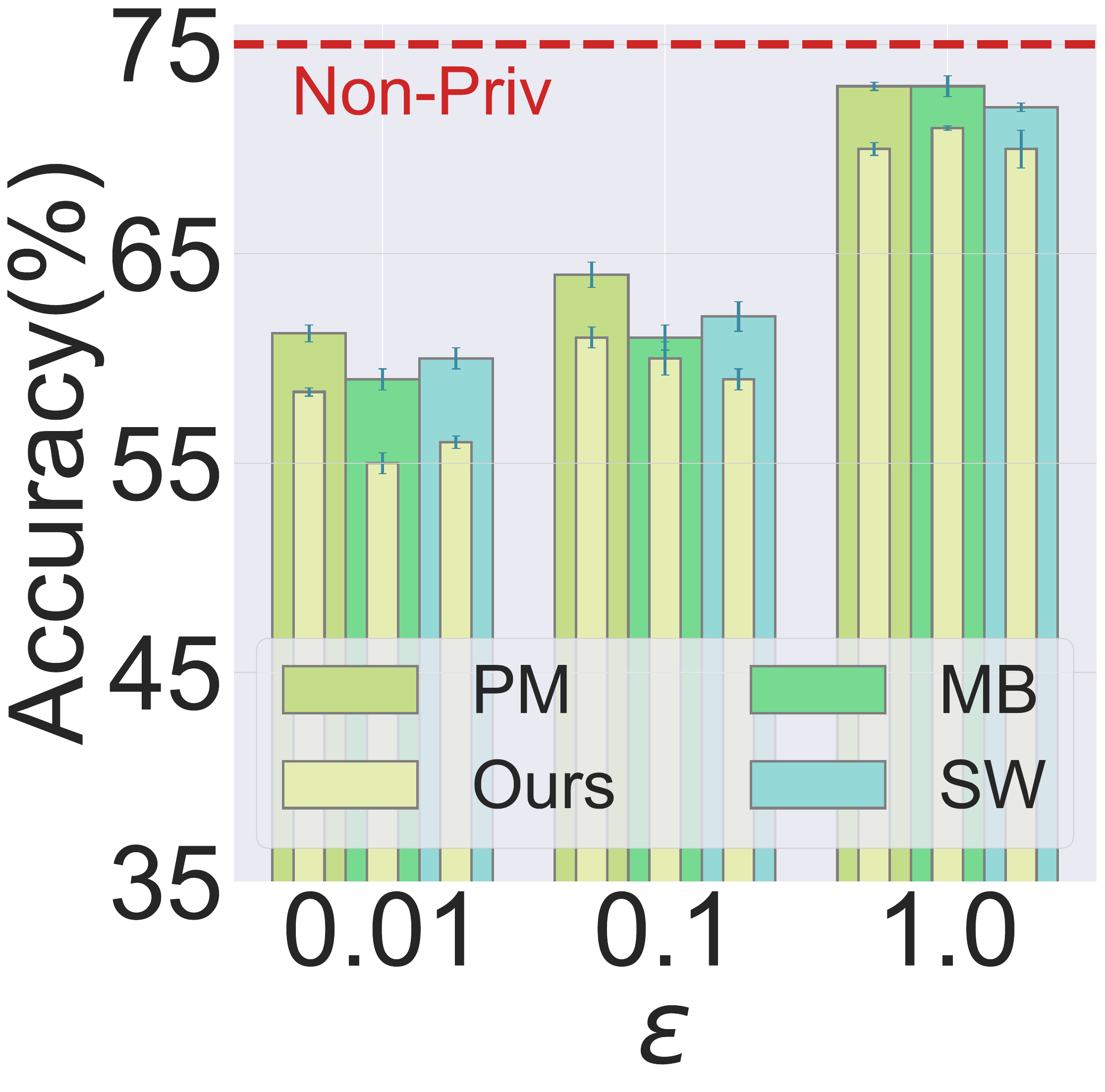}}
        \hspace{0.2cm}\subfigure[\normalsize Facebook]{\includegraphics[width=0.15\textwidth]{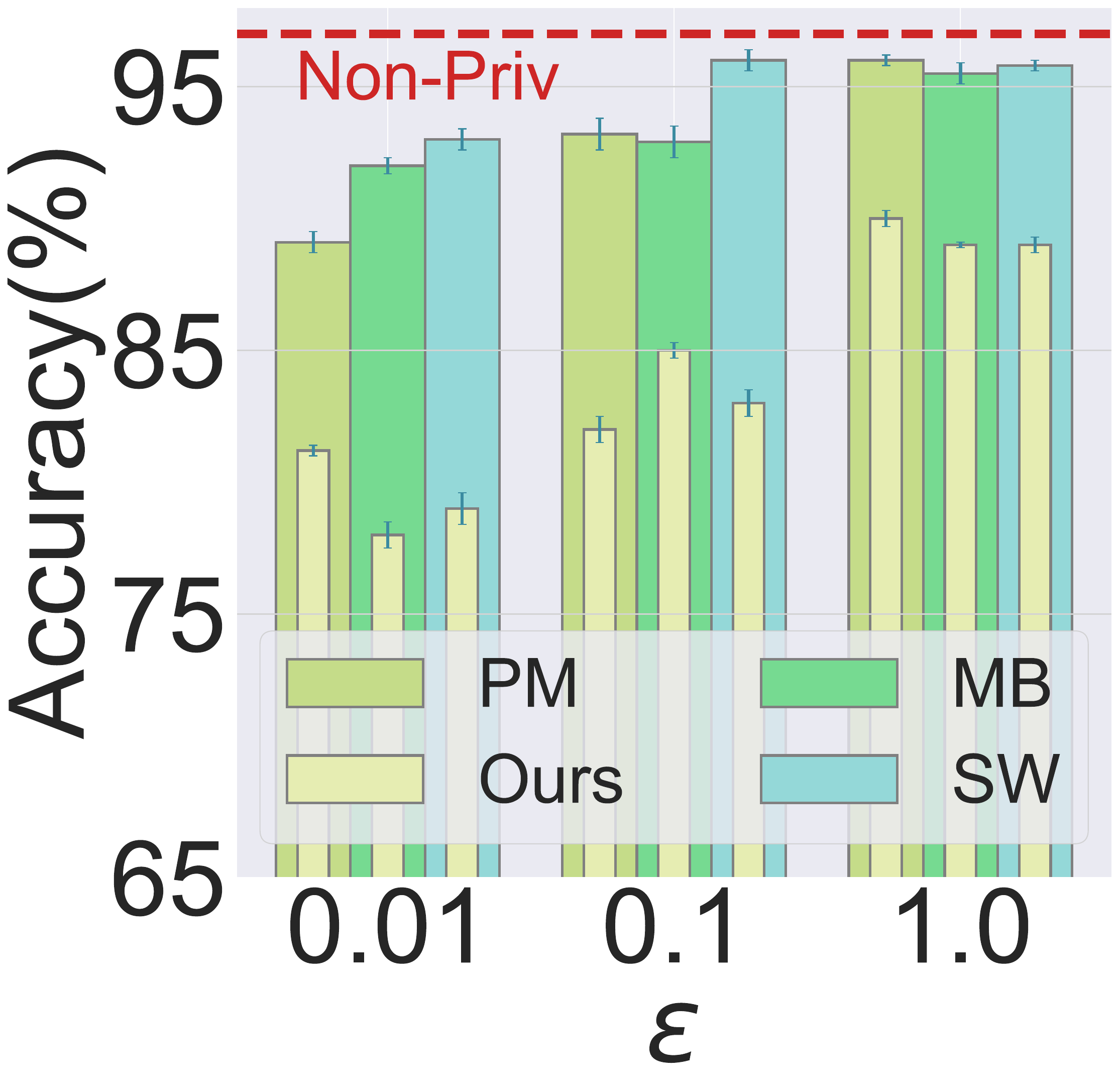}}
    \end{minipage}


     \hspace{-0.3cm}\raisebox{-3mm}{\rotatebox{90}{untargeted}} 
    \hspace{-0.4cm} 
   \begin{minipage}{\textwidth}
        \centering
        \hspace{0.4cm}\subfigure[\normalsize Cora]{\includegraphics[width=0.15\textwidth]{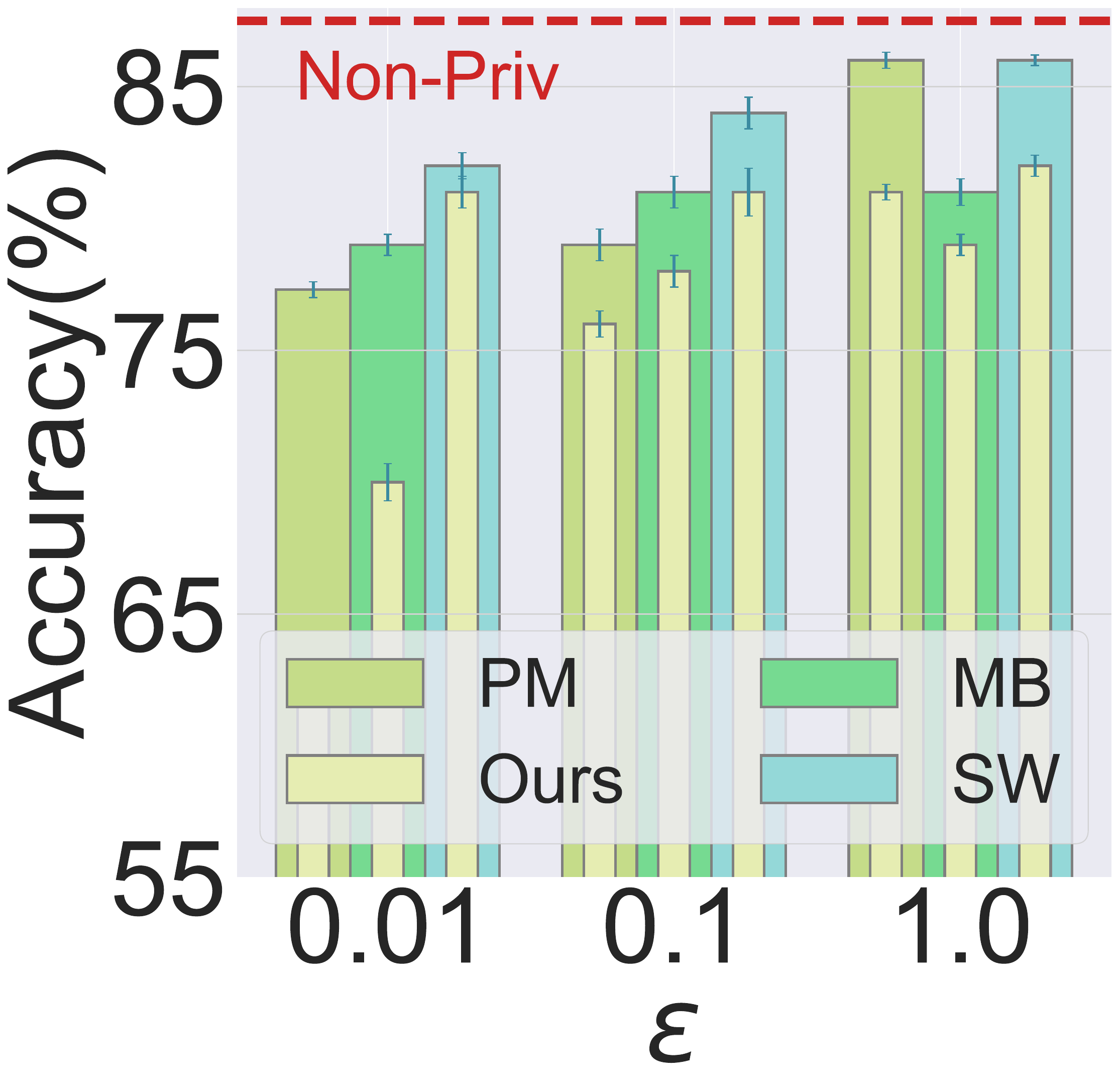}}
        \hspace{0.2cm}\subfigure[\normalsize Citeseer]{\includegraphics[width=0.15\textwidth]{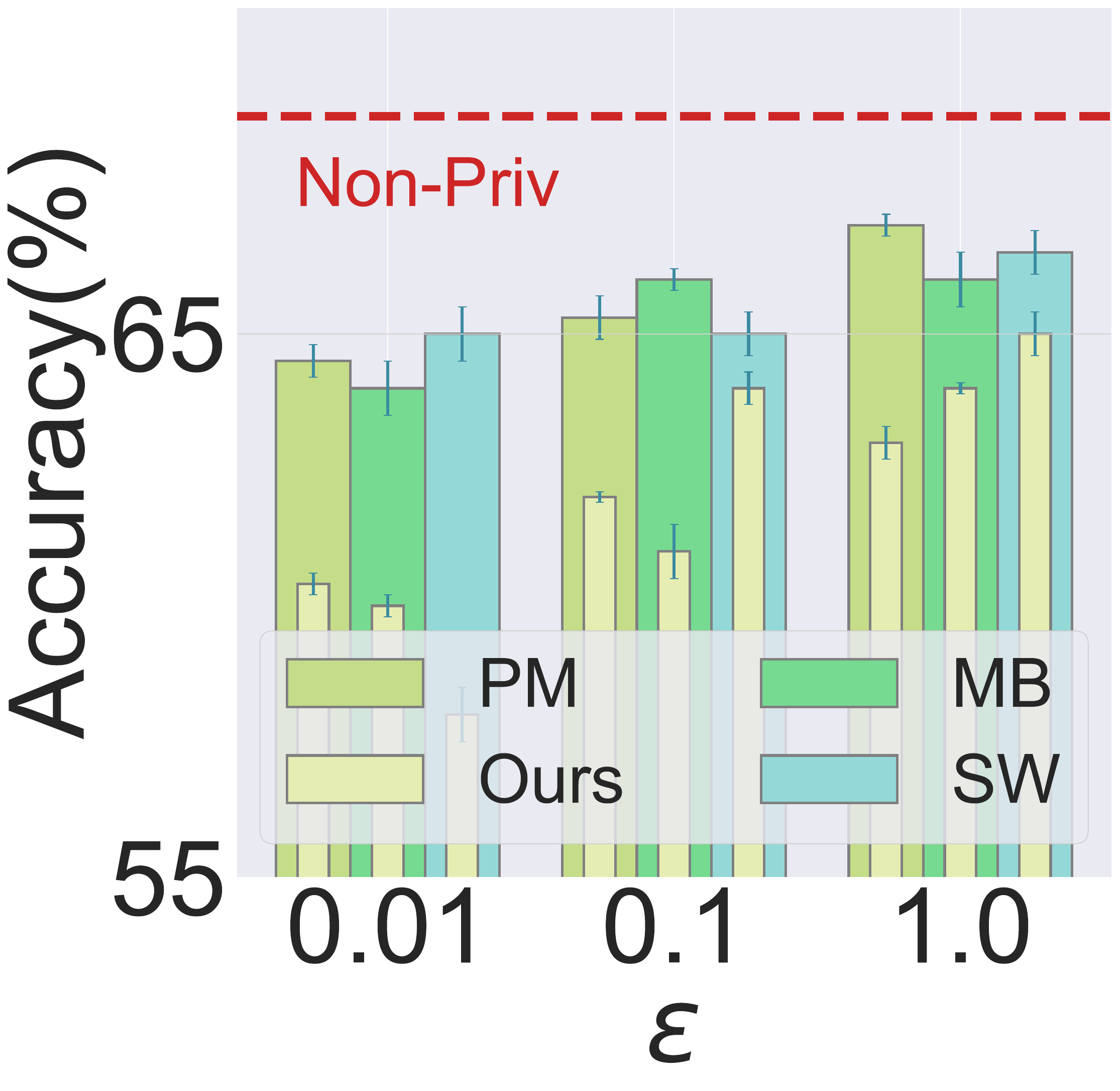}}
        \hspace{0.2cm}\subfigure[\normalsize Pubmed]{\includegraphics[width=0.15\textwidth]{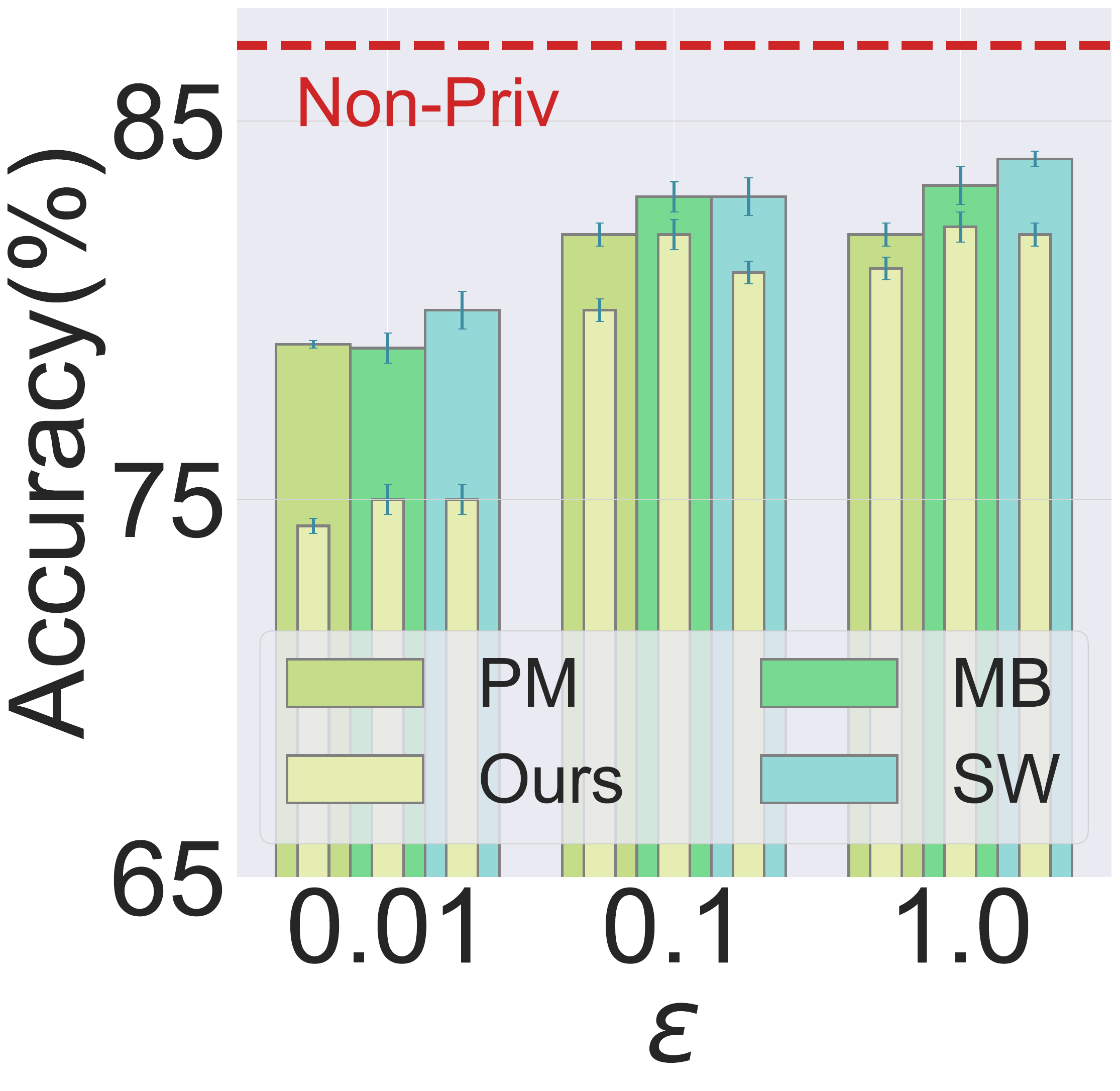}}
        \hspace{0.2cm}\subfigure[\normalsize LastFM]{\includegraphics[width=0.15\textwidth]{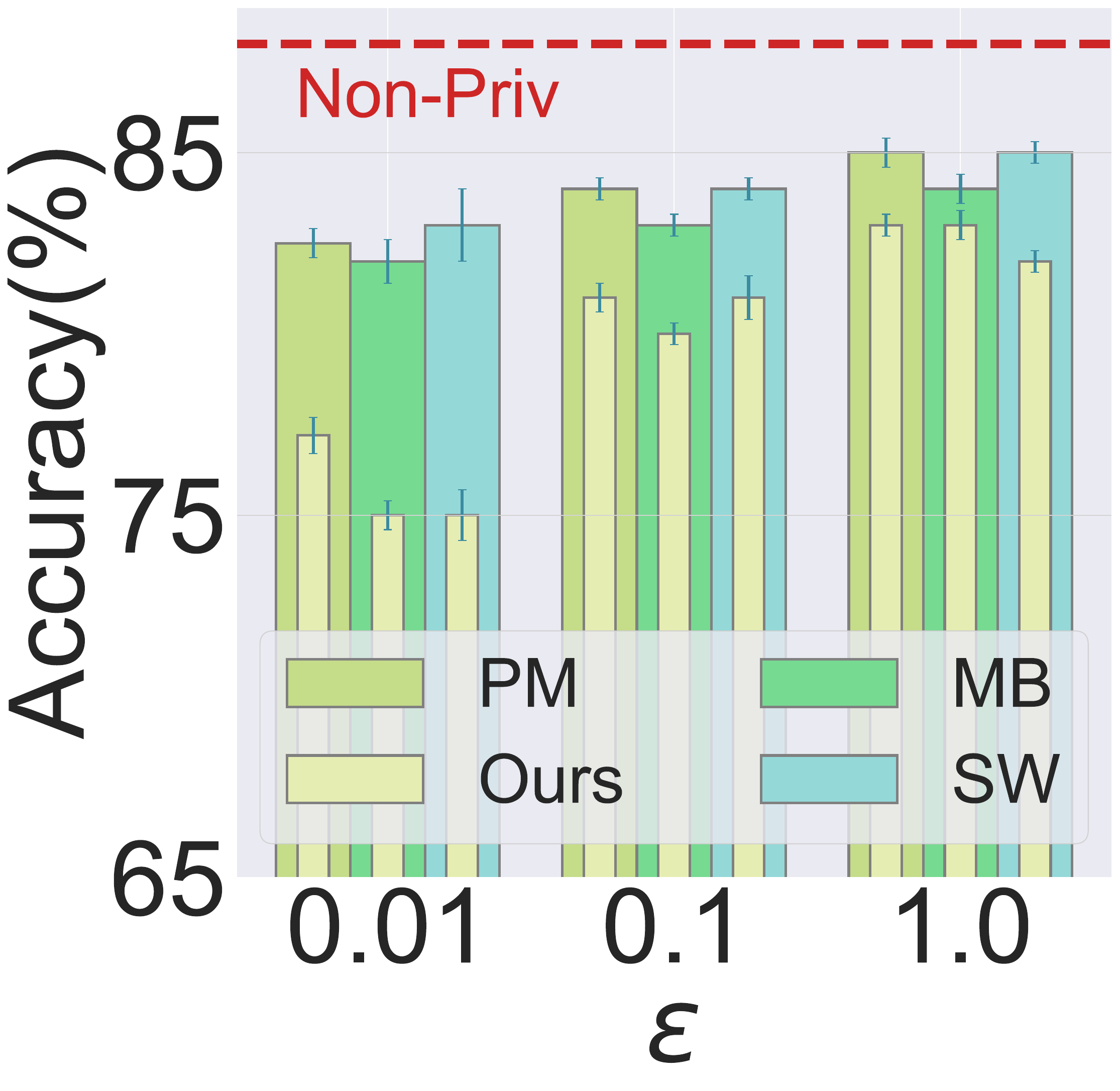}}
        \hspace{0.2cm}\subfigure[\normalsize Twitch]{\includegraphics[width=0.15\textwidth]{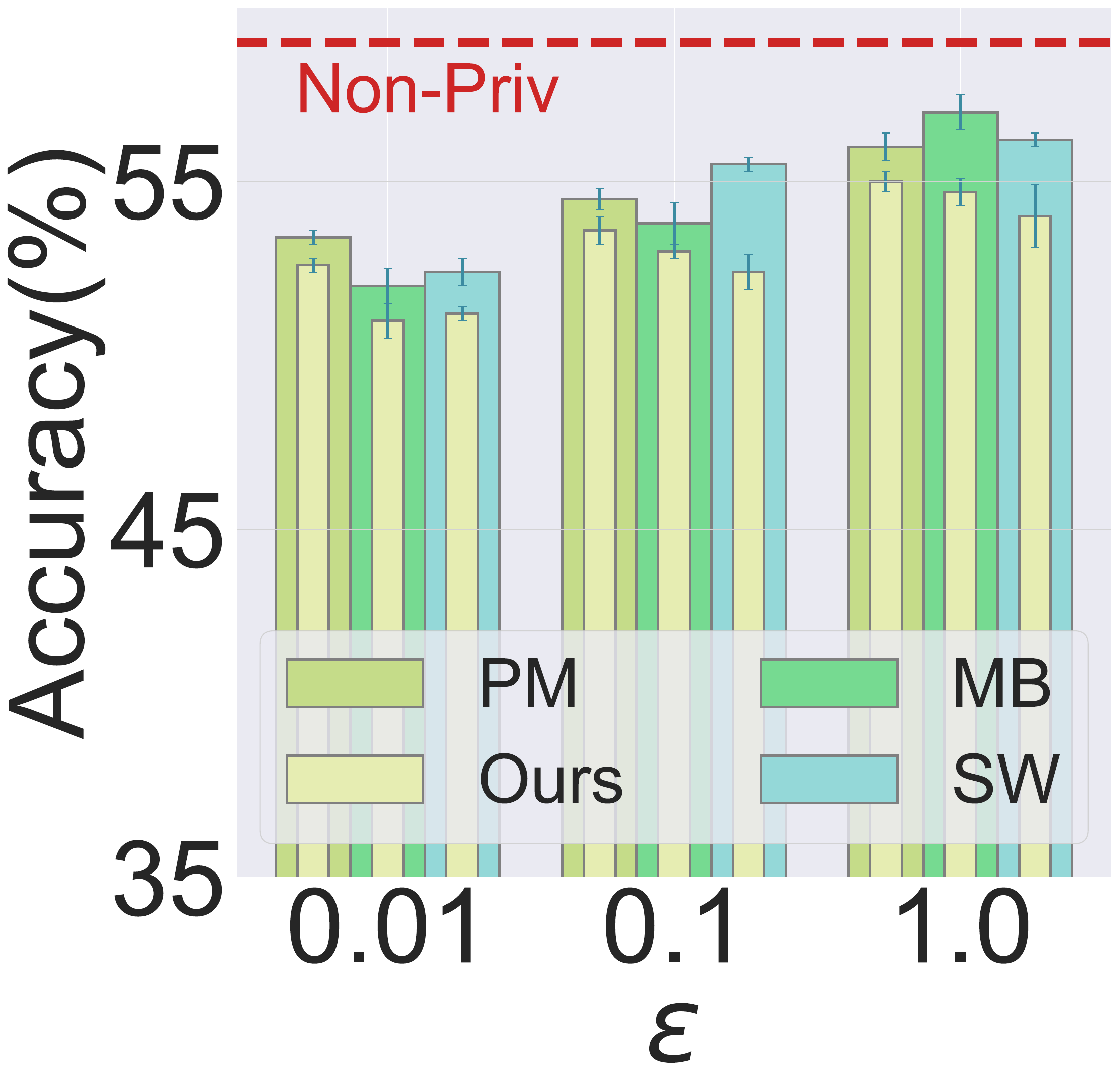}}
        \hspace{0.2cm}\subfigure[\normalsize Facebook]{\includegraphics[width=0.15\textwidth]{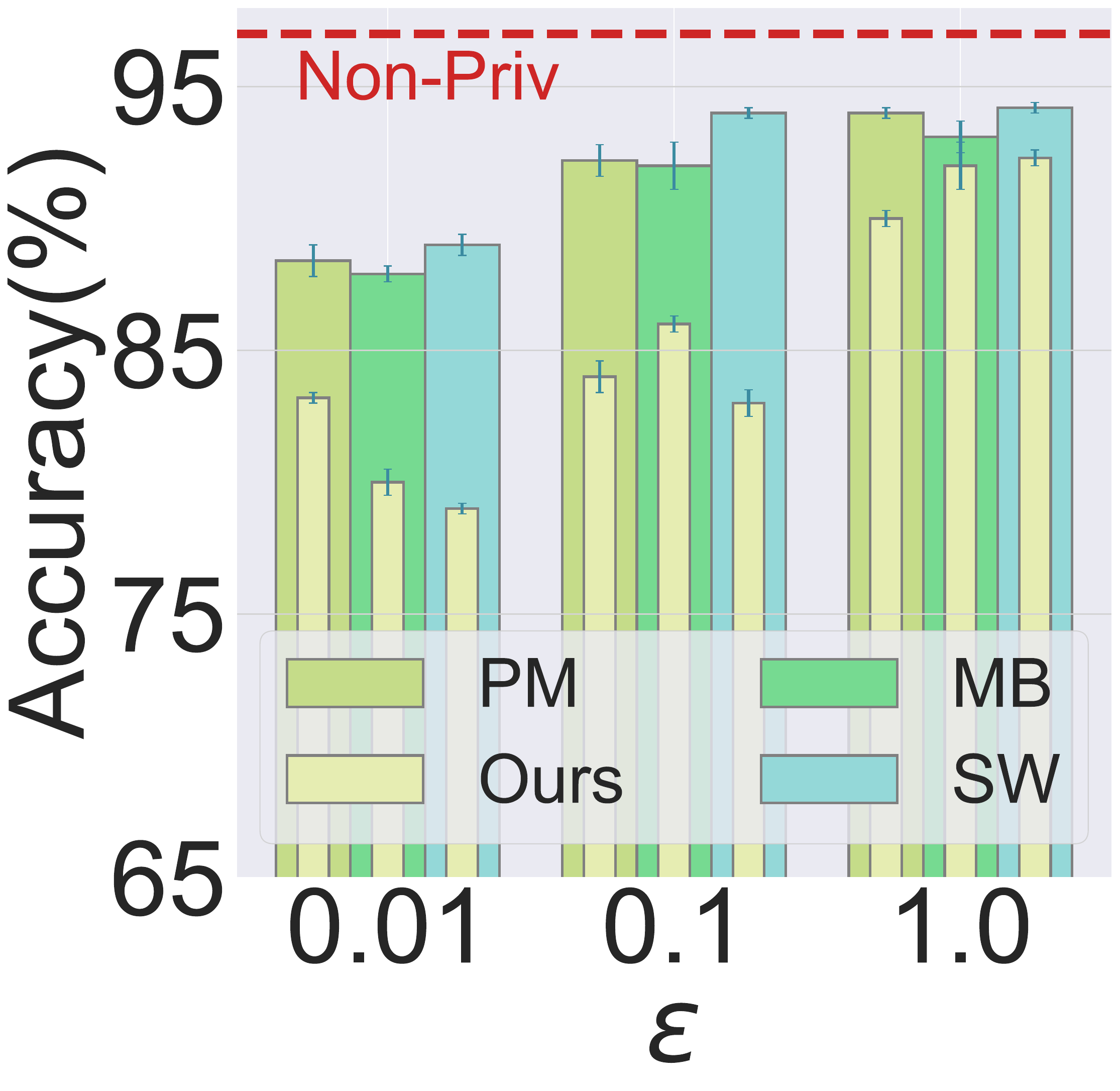}}
    \end{minipage}
     \vspace{-1.3em}
    \caption{Attack performance under different privacy budgets $\epsilon$ and LDP mechanisms for \textit{node classification} task. The x-axis represents the privacy budget $\epsilon$, while the y-axis shows the accuracy for both the targeted and untargeted attacks. “Non-Priv” denotes the non-perturbed baseline. Our attacks significantly reduce the accuracy of various LDP mechanisms across all cases.}
    \label{fig3}
    \vspace{-0.5em}
\end{figure*}

\begin{table}[t]
  \centering
  \caption{Statistics of datasets.}
  \vspace{-1em}
  \resizebox{\linewidth}{!}{
    \begin{tabular}{llrrrrr}
    \toprule
    \textbf{Type} & \textbf{Dataset} & \textbf{Nodes} & \textbf{Edges} & \textbf{Features} & \textbf{Classes} & \textbf{Avg. Deg.}\\
    \midrule
    \multirow{3}{*}{\makecell[l]{Citation\\Network}} 
    & Cora & 2,708 & 5,278 & 1,433 & 7 & 3.90 \\
     & Citeseer & 3,327 & 4,552 & 3,703  & 6 & 2.74 \\
    & Pubmed & 19,717 & 44,324 & 500 & 3 & 4.50\\
     \midrule
    \multirow{3}{*}{\makecell[l]{Social\\Network}} 
    & LastFM & 7,624 & 27,806 & 7,842 & 18&7.29 \\
    & Twitch & 4,648 & 61,706 & 128 & 2&16.19 \\
    & Facebook & 22,470 & 170,912 & 4,714 & 4 & 15.21\\
    \bottomrule
    \end{tabular}}
    \begin{tablenotes}
     \item[1] \textit{Avg. Deg.} in the table denotes the average node degree.
   \end{tablenotes}
    \vspace{-1em}
  \label{tab:1}
\end{table}

\vspace{-0.5em}
\section{EVALUATION}\label{sec:exp}
In this section, we comprehensively evaluate the performance of our attack based on node classification and link prediction tasks.\footnote{The code will be made publicly available upon acceptance of the paper.}

\vspace{-0.5em}
\subsection{Experimental Settings}
\subsubsection{Datasets} 
We conduct experiments on six representative real-world datasets: three citation networks (Cora~\cite{yang2016revisiting}, Citeseer~\cite{yang2016revisiting}, and Pubmed~\cite{yang2016revisiting}) and three social networks (LastFM~\cite{rozemberczki2020characteristic}, Twitch~\cite{rozemberczki2021multi}, and Facebook~\cite{rozemberczki2021multi}). We summarize the dataset statistics in Table~\ref{tab:1}.

\subsubsection{GNN models} 
We consider three state-of-the-art GNN models: graph convolutional networks (GCN)~\cite{DBLP:conf/iclr/KipfW17}, GraphSAGE (SAGE) \cite{hamilton2017inductive}, and graph attention networks (GAT)~\cite{DBLP:conf/iclr/VelickovicCCRLB18}. All GNN models consist of two graph convolutional layers, with 64 neurons in each hidden layer and ReLU as the activation function~\cite{klambauer2017self}, followed by dropout. The GAT model has four attention heads. We implement the GNN models in PyTorch using PyTorch-Geometric (PyG)\footnote{https://www.pyg.org}. All experiments are carried on a machine running Ubuntu 20.04 LTS, equipped with two Intel\textsuperscript{\textregistered} Xeon\textsuperscript{\textregistered} Gold 6348 CPUs, 100GB RAM and an NVIDIA\textsuperscript{\textregistered} A800 80GB GPU. By default, the model used is GCN.

\subsubsection{LDP mechanism}
We consider three de facto LDP perturbation mechanisms for protecting node features: the multi-bit mechanism (MB)~\cite{sajadmanesh2021locally,lin2022towards}, the piecewise mechanism (PM)~\cite{pei2023privacy,wang2019collecting},  and the square wave mechanism (SW)~\cite{li2024privacy,li2020estimating}. Both mechanisms perturb the node features based on a privacy budget $\epsilon$, ensuring strict privacy guarantees. By default, the LDP mechanism used is PM. 

\subsubsection{Parameter settings}
For all datasets, we randomly divide the nodes into training, validation, and testing sets with a 50/25/25\% split. The privacy budget $\epsilon$ ranges from \{0.01, 0.1, 1.0\}, the ratio of target nodes to the original graph, $\eta_1=|\mathcal{V}_t|/|\mathcal{V}|$, ranges from \{0.01, 0.03, 0.05, 0.07, 0.09\}, and the ratio of fake nodes to target nodes, $\eta_2=|\mathcal{V}_{\text{atk}}|/|\mathcal{V}_t|$, ranges from \{0.6, 0.7, 0.8, 0.9, 1.0\}. In default, $\epsilon$ is set to 0.01; $\eta_1$ and $\eta_2$ are set to 0.09 and 0.8, respectively. The parameter \(K\) for the calibration phase takes values from the range \(\{0, 2, 4, 8, 16\}\).  For more details on hyperparameters, refer to App.~\ref{c.1}.

\subsubsection{Evaluation metrics}
We conduct experiments based on node classification~\cite{sajadmanesh2021locally,lin2022towards,pei2023privacy,li2024privacy} and link prediction~\cite{zhang2018link,zhao2024conformalized} tasks. We use node classification accuracy (or link prediction accuracy) as the evaluation metric to assess the effectiveness of attacks on all target nodes (\textit{targeted attacks}) and on the entire graph (\textit{untargeted attacks}). The default task is node classification. We measure the accuracy of the test set over 10 consecutive runs and report the mean and 95\% confidence intervals, calculated via bootstrapping with 1000 samples. More details, refer to App.~\ref{c.2}.

\begin{table}[t]
	\centering
	\caption{Comparison of accuracy before and after the attack across different GNN models. ↓ represents accuracy drop.}
    \vspace{-1em}
	\label{tab:ldpp}
	\resizebox{\linewidth}{!}{\begin{tabular}{c|l|llllll}
		\toprule
		GNN  & Meth. & Cora  & Citeseer   & Pubmed  & LastFM  & Twitch & Facebook         \\
		\midrule
		 \multirow{2}*{GCN}   
		& \textsc{pm}    & 84.2 & 76.4 & 81.4 & 89.2& 61.2&89.1       \\
		& \textsc{pm}$\ssymbol{2}$     & 66.1 \roundbox{↓}   & 62.9 \roundbox{↓} & 74.3 \roundbox{↓}& 79.1 \roundbox{↓} & 58.4 \roundbox{↓} & 81.2 \roundbox{↓}    \\		
		\midrule
        \multirow{2}*{SAGE}   
		& \textsc{pm}       & 82.7   & 71.9& 78.5& 81.8& 61.8 & 83.3       \\
		& \textsc{pm}$\ssymbol{2}$     & 62.3 \roundbox{↓}   & 55.2 \roundbox{↓} & 55.0 \roundbox{↓}& 54.8 \roundbox{↓} & 56.5 \roundbox{↓} & 73.9 \roundbox{↓}    \\
		\midrule
		 \multirow{2}*{GAT}   
		& \textsc{pm}       & 83.1   &77.9& 77.7& 84.3& 64.3& 83.6      \\
		& \textsc{pm}$\ssymbol{2}$     & 67.8 \roundbox{↓}   & 54.9 \roundbox{↓}& 72.2 \roundbox{↓} & 58.9 \roundbox{↓}& 62.3 \roundbox{↓} & 73.4 \roundbox{↓}    \\
		\bottomrule
	\end{tabular}}
    \begin{tablenotes}
     \item[1] $\ssymbol{2}$ represents the accuracy after targeted attack.
   \end{tablenotes}
    \vspace{-0.9em}
 \label{tab2}
\end{table}

\subsection{Evaluating the Effectiveness of the Attack}\label{sec6.2}
\subsubsection{Node Classification}
In this experiment, we assess the attack performance under different privacy budgets and LDP mechanisms using the node classification task, with the results shown in Fig.~\ref{fig3}. As demonstrated in the figure, our attack method significantly reduces the accuracy of the three LDP mechanisms in all cases, validating its effectiveness. Notably, the accuracy reduction is more pronounced under the targeted attack (affecting the target nodes, top row of Fig.~\ref{fig3}) than under the untargeted attack (affecting the entire graph, bottom row). This is because our attack specifically targets the set of nodes, which further compromises the utility of the entire graph through malicious message passing, making the impact more significant for the target nodes. Furthermore, our attack method proves to be more effective under higher privacy settings. For instance, in Fig.~\ref{fig3}(c), when $\epsilon=1.0$, there is an average accuracy degradation of 3\%, while for $\epsilon=0.01$, the degradation rises to 8.4\%. This is due to the fact that higher privacy settings result in more noise being injected into the node features, reducing the proportion of benign information, thereby amplifying the impact of malicious node features on the private graph learning process.
\begin{figure}[t]
	\centering
    \vspace{-1em}
	\begin{tabular}{cc}
		\multicolumn{2}{c}{\textbf{}} \\
  \includegraphics[width=0.45\linewidth]{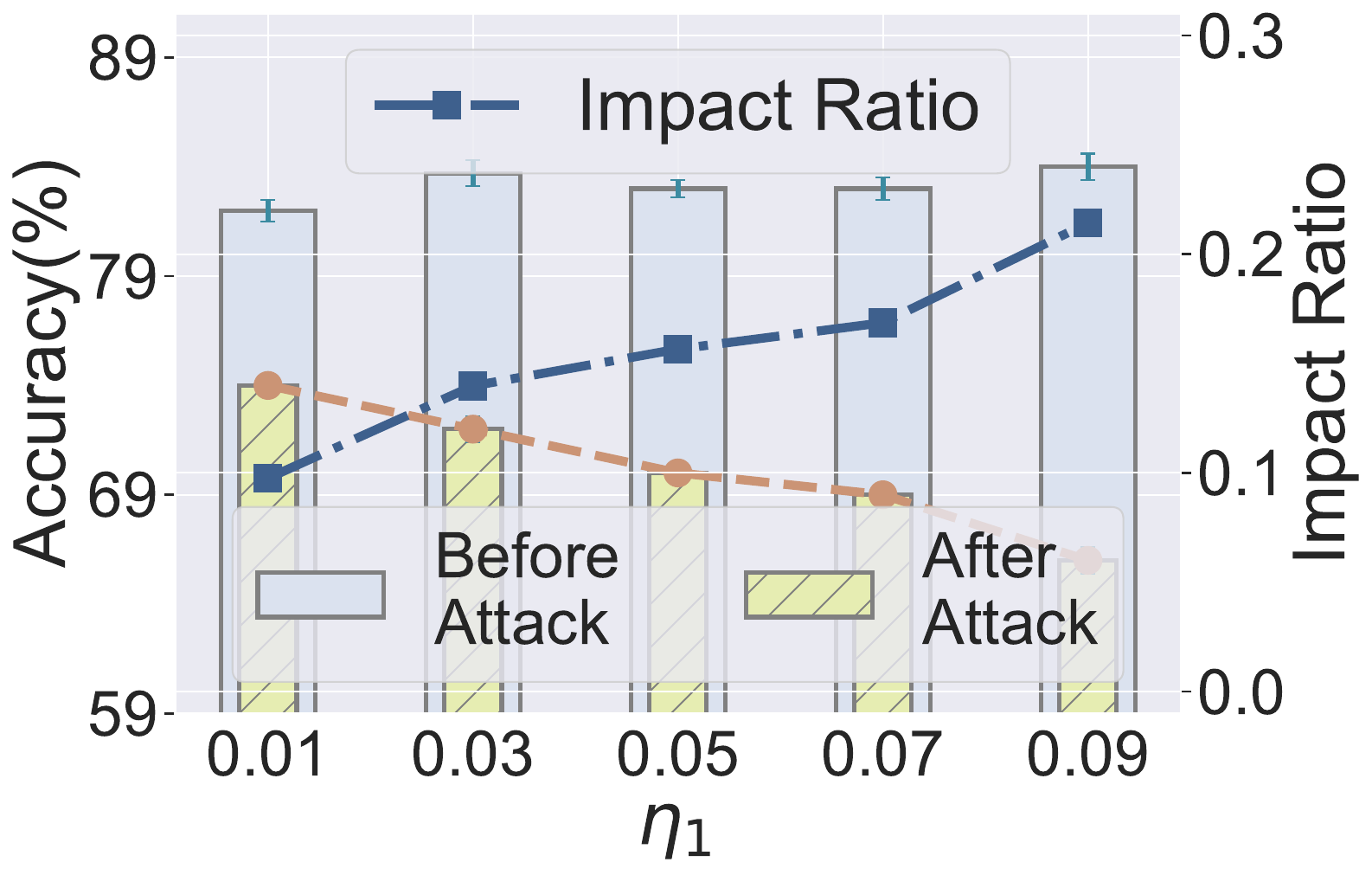} & \includegraphics[width=0.46\linewidth]{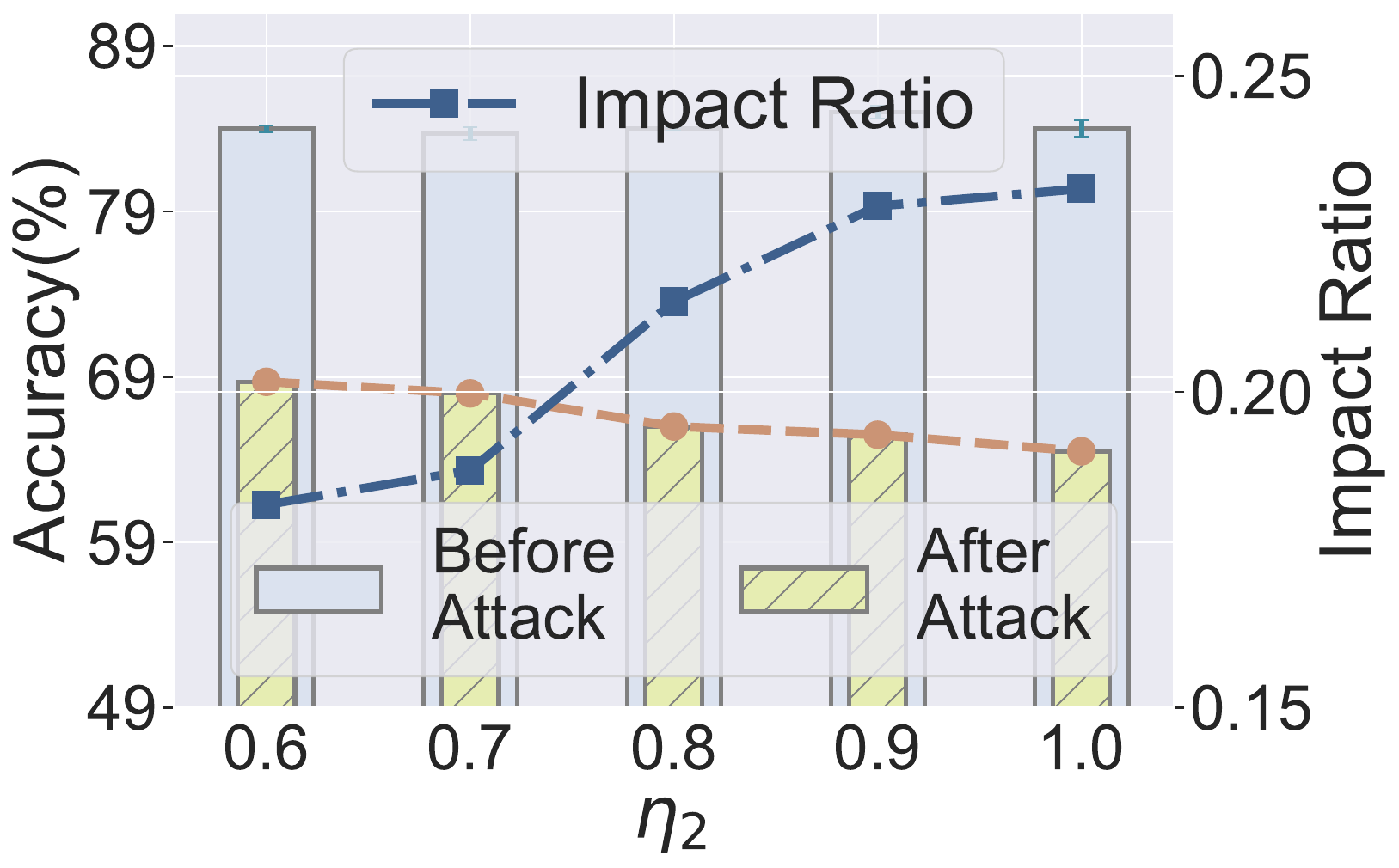}
	\end{tabular}
    \vspace{-1em}
	\caption{Variation in target attack accuracy at different $\eta_1$ and $\eta_2$ values. The impact ratio represents the proportion of accuracy reduction compared to the normal accuracy.}
    \vspace{-1em}
	\label{fig444}
\end{figure}
\subsubsection{Link Prediction}
\begin{figure*}[htp]
    \centering
    \hspace{-0.3cm}\raisebox{-1mm}{\rotatebox{90}{targeted}} 
    \hspace{-0.4cm} 
    \begin{minipage}{\textwidth}
        \centering
        \hspace{0.4cm}\subfigure[\normalsize Cora]{\includegraphics[width=0.15\textwidth]{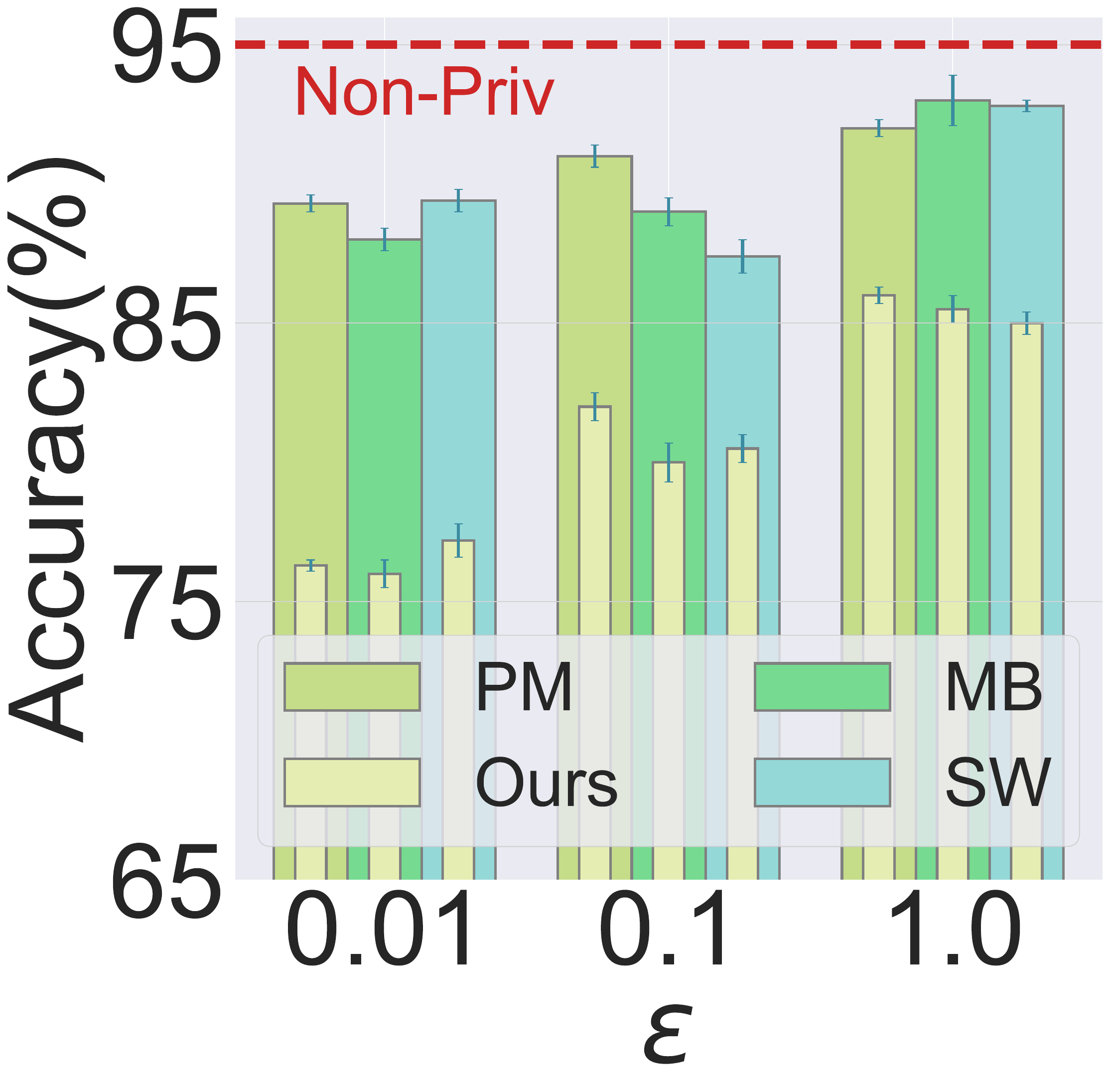}}
        \hspace{0.2cm}\subfigure[\normalsize Citeseer]{\includegraphics[width=0.15\textwidth]{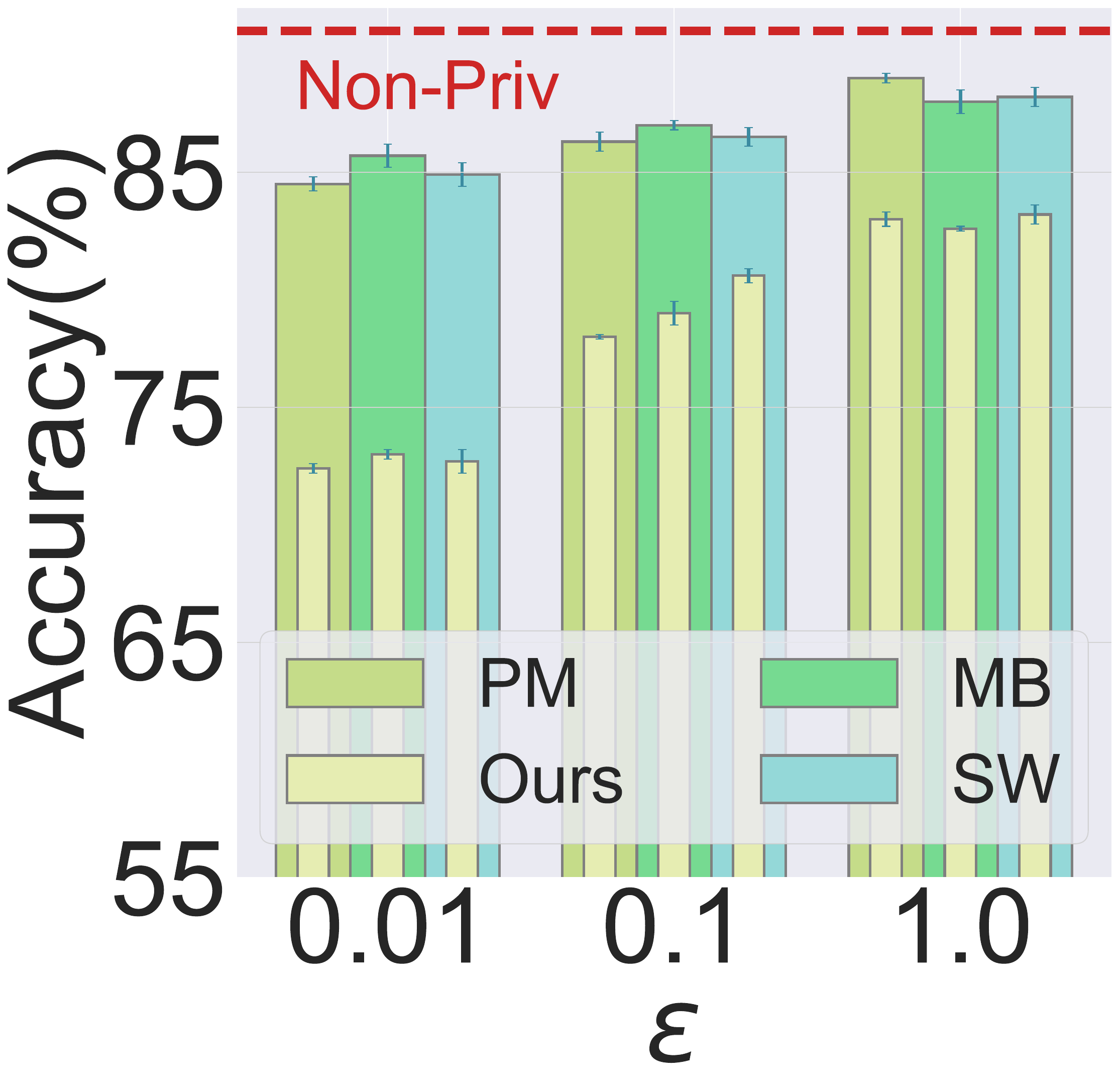}}
        \hspace{0.2cm}\subfigure[\normalsize Pubmed]{\includegraphics[width=0.15\textwidth]{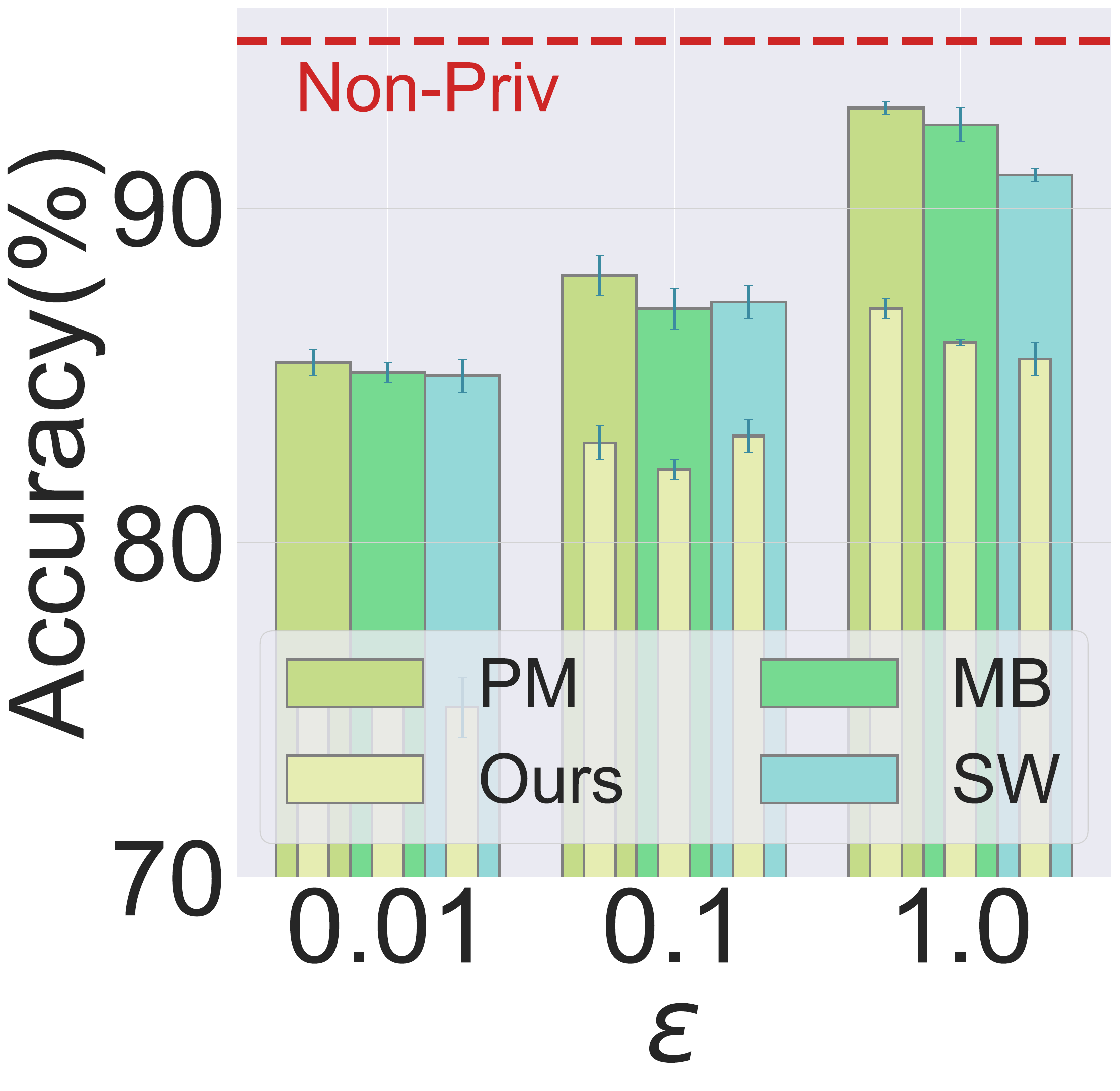}}
        \hspace{0.2cm}\subfigure[\normalsize LastFM]{\includegraphics[width=0.15\textwidth]{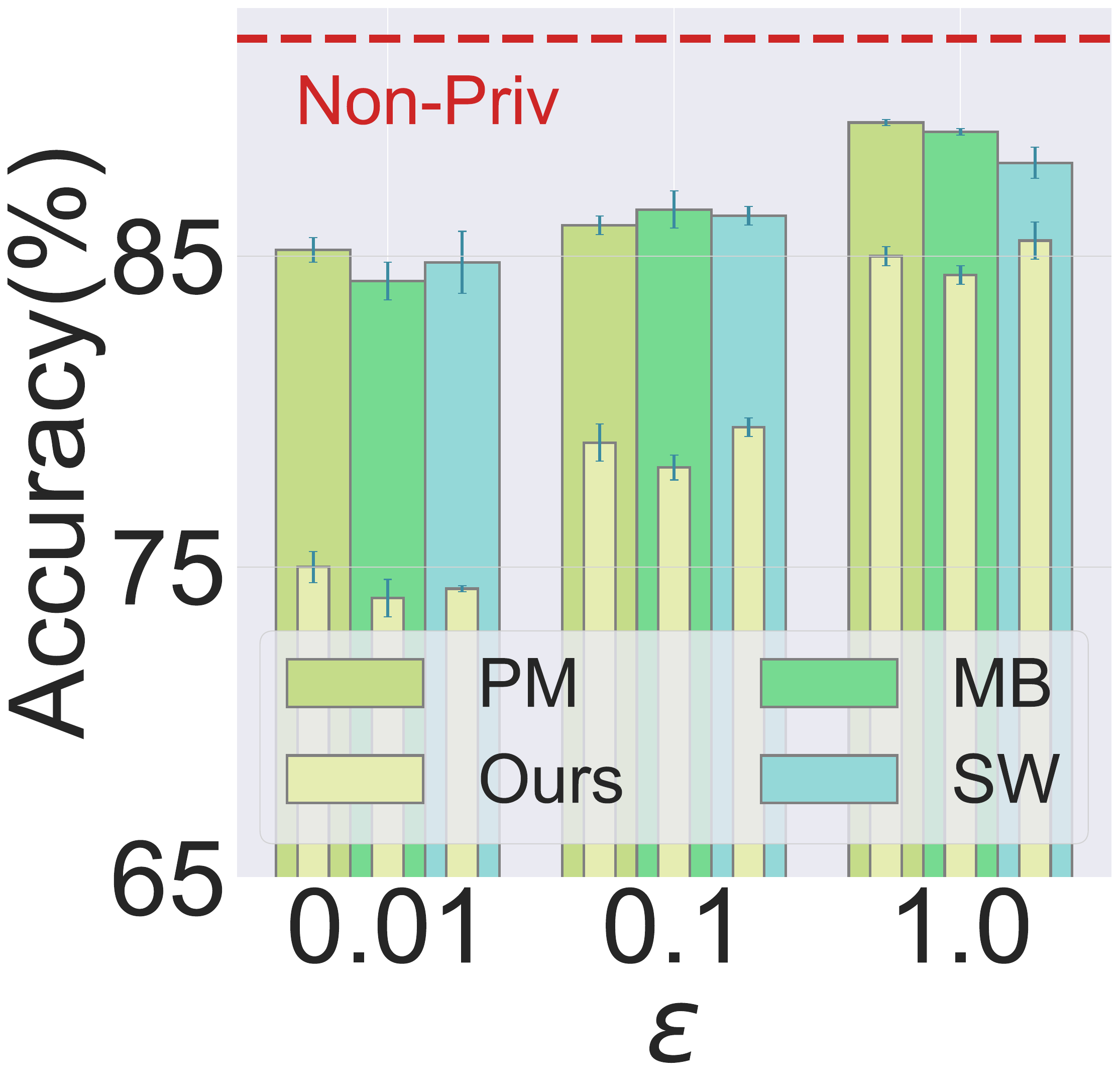}}
        \hspace{0.2cm}\subfigure[\normalsize Twitch]{\includegraphics[width=0.15\textwidth]{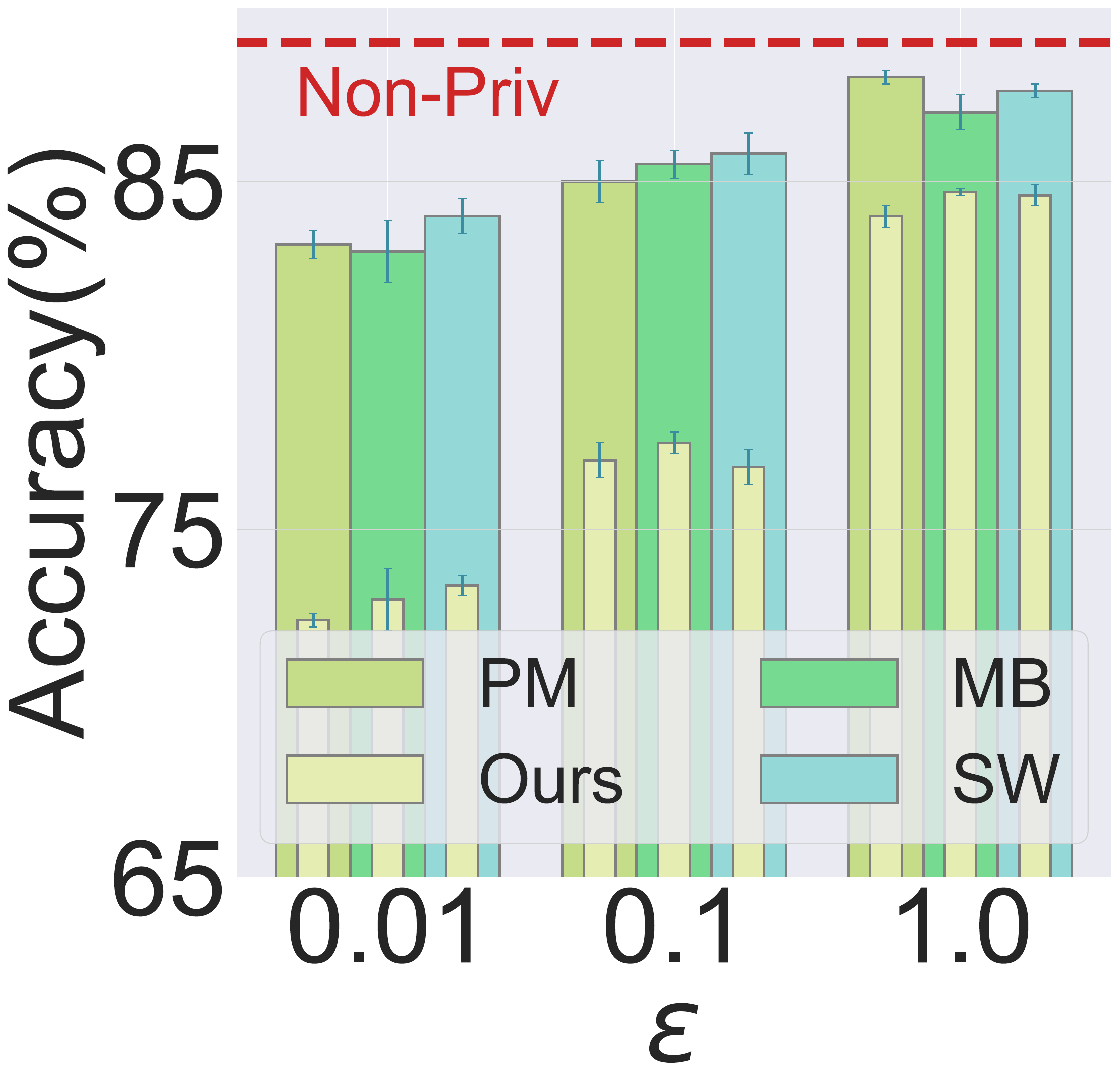}}
        \hspace{0.2cm}\subfigure[\normalsize Facebook]{\includegraphics[width=0.15\textwidth]{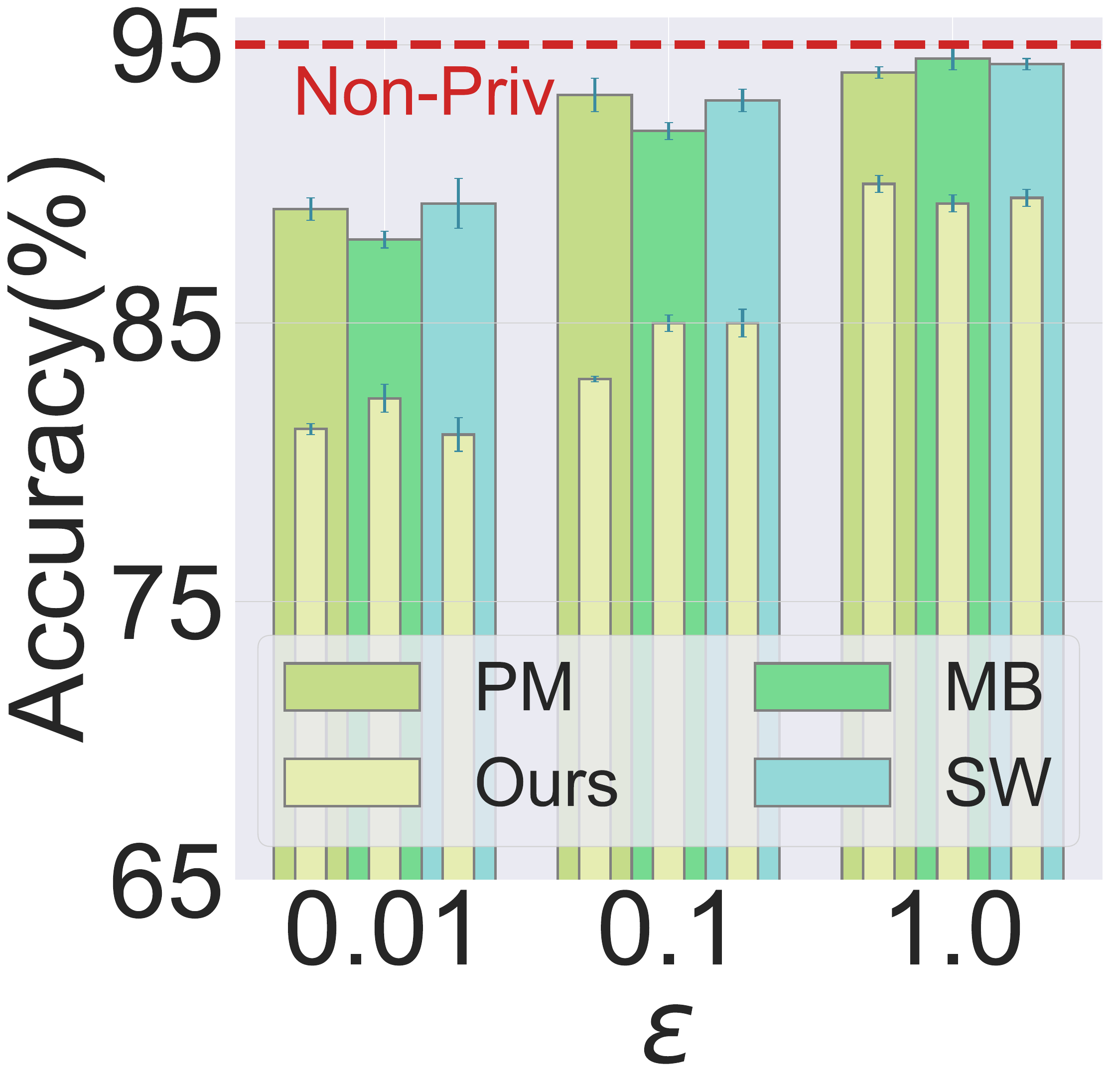}}
    \end{minipage}


     \hspace{-0.3cm}\raisebox{-1mm}{\rotatebox{90}{untargeted}} 
    \hspace{-0.4cm} 
   \begin{minipage}{\textwidth}
        \centering
        \hspace{0.4cm}\subfigure[\normalsize Cora]{\includegraphics[width=0.15\textwidth]{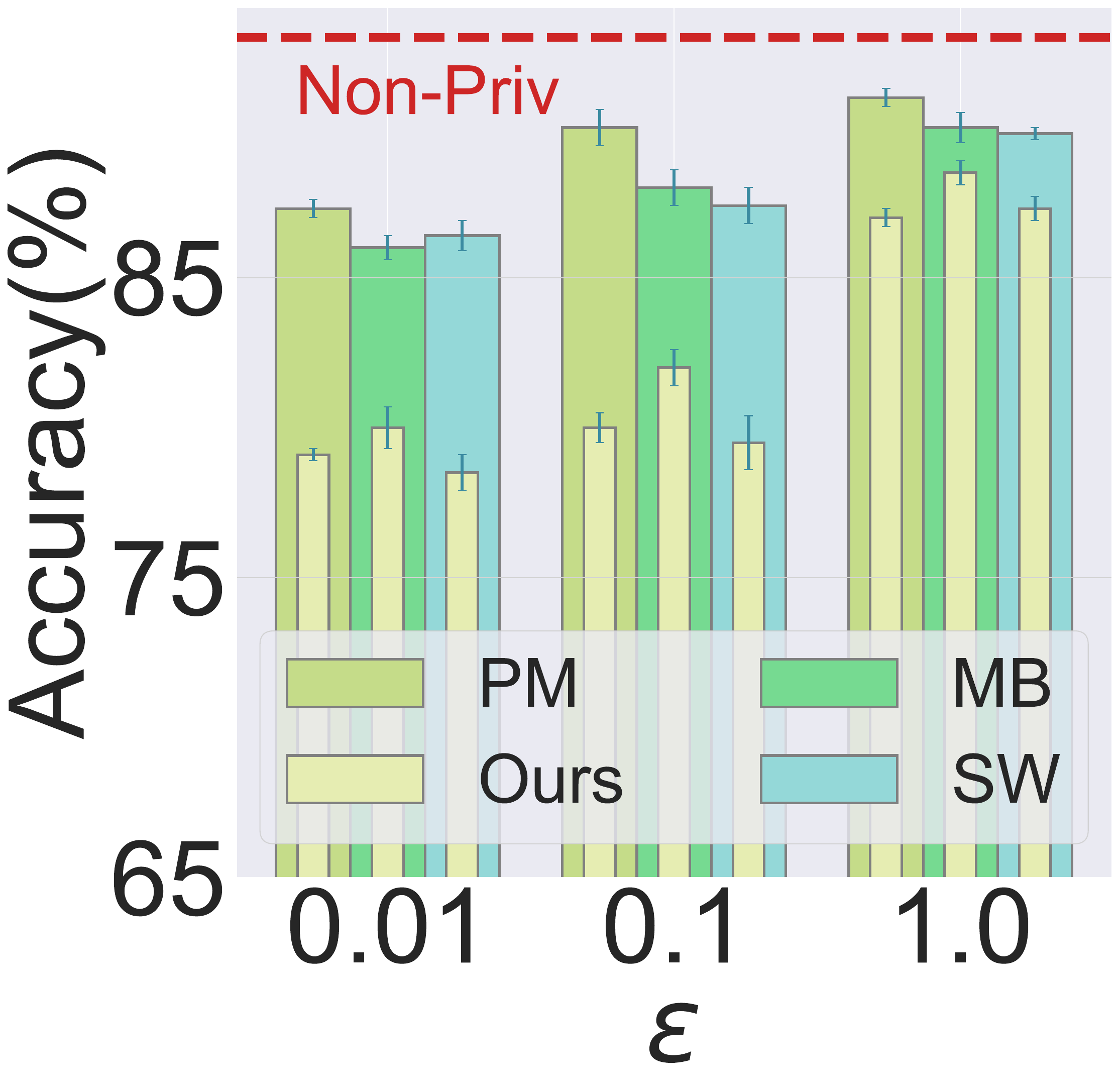}}
        \hspace{0.2cm}\subfigure[\normalsize Citeseer]{\includegraphics[width=0.15\textwidth]{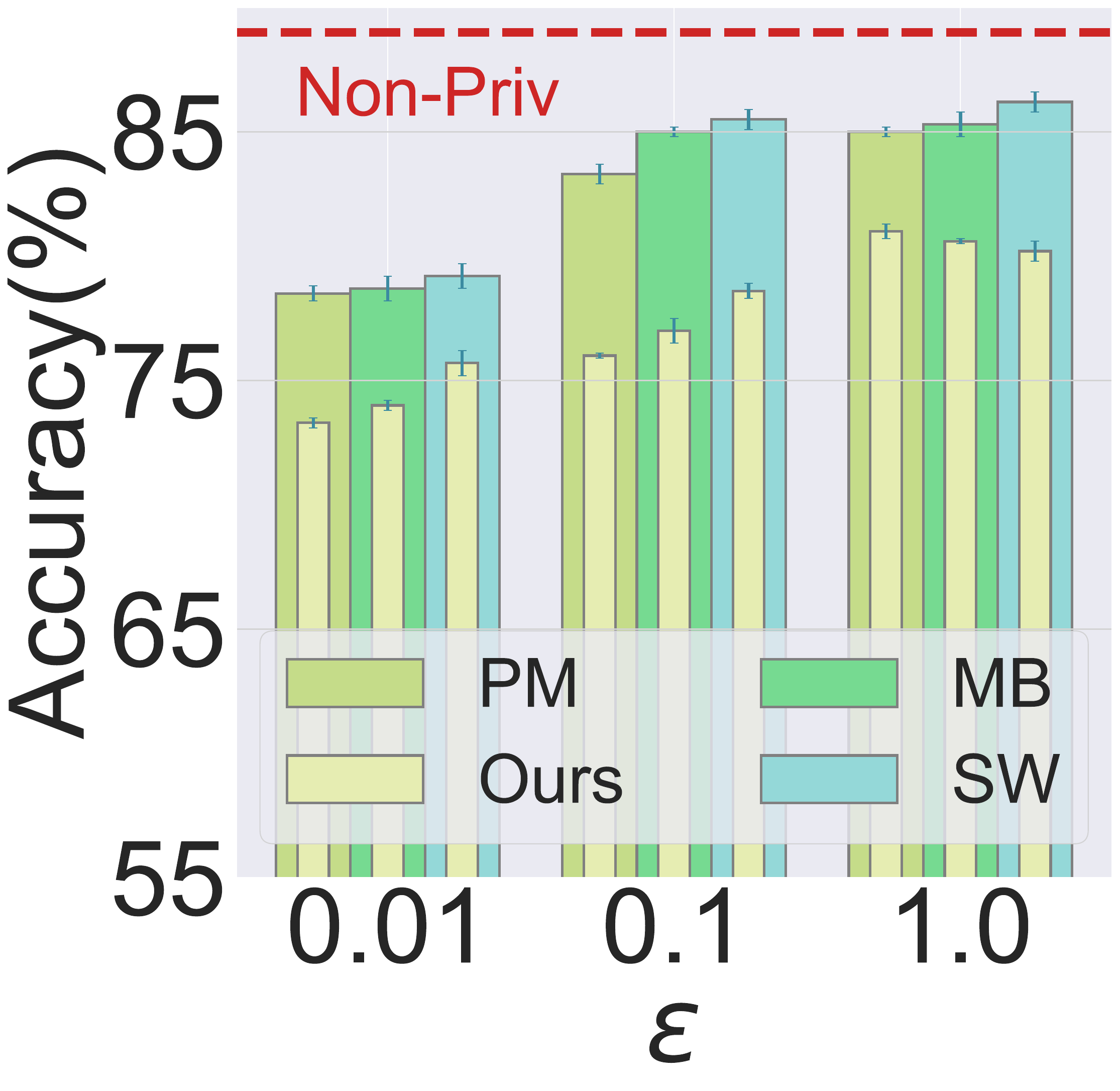}}
        \hspace{0.2cm}\subfigure[\normalsize Pubmed]{\includegraphics[width=0.15\textwidth]{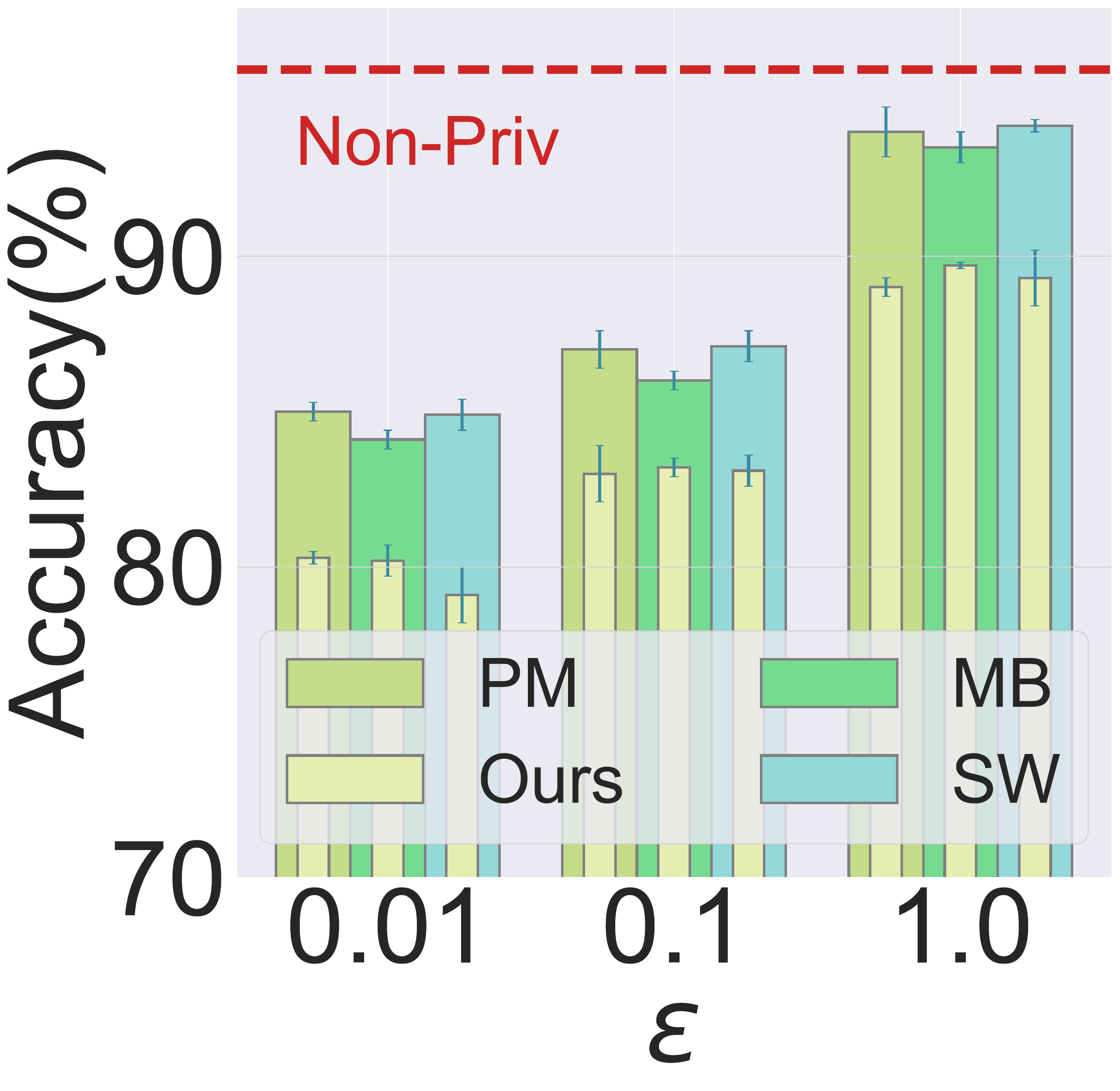}}
        \hspace{0.2cm}\subfigure[\normalsize LastFM]{\includegraphics[width=0.15\textwidth]{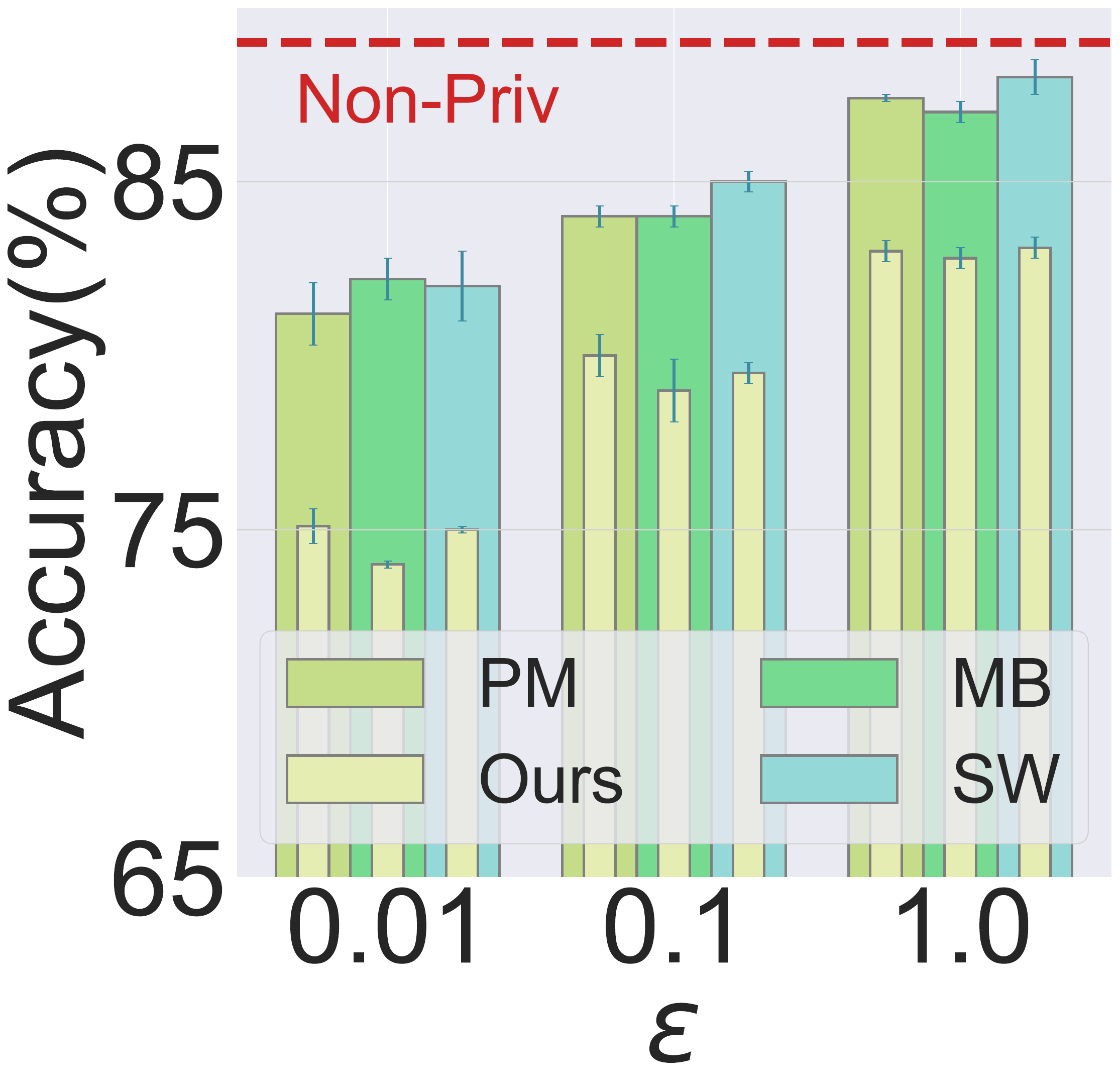}}
        \hspace{0.2cm}\subfigure[\normalsize Twitch]{\includegraphics[width=0.15\textwidth]{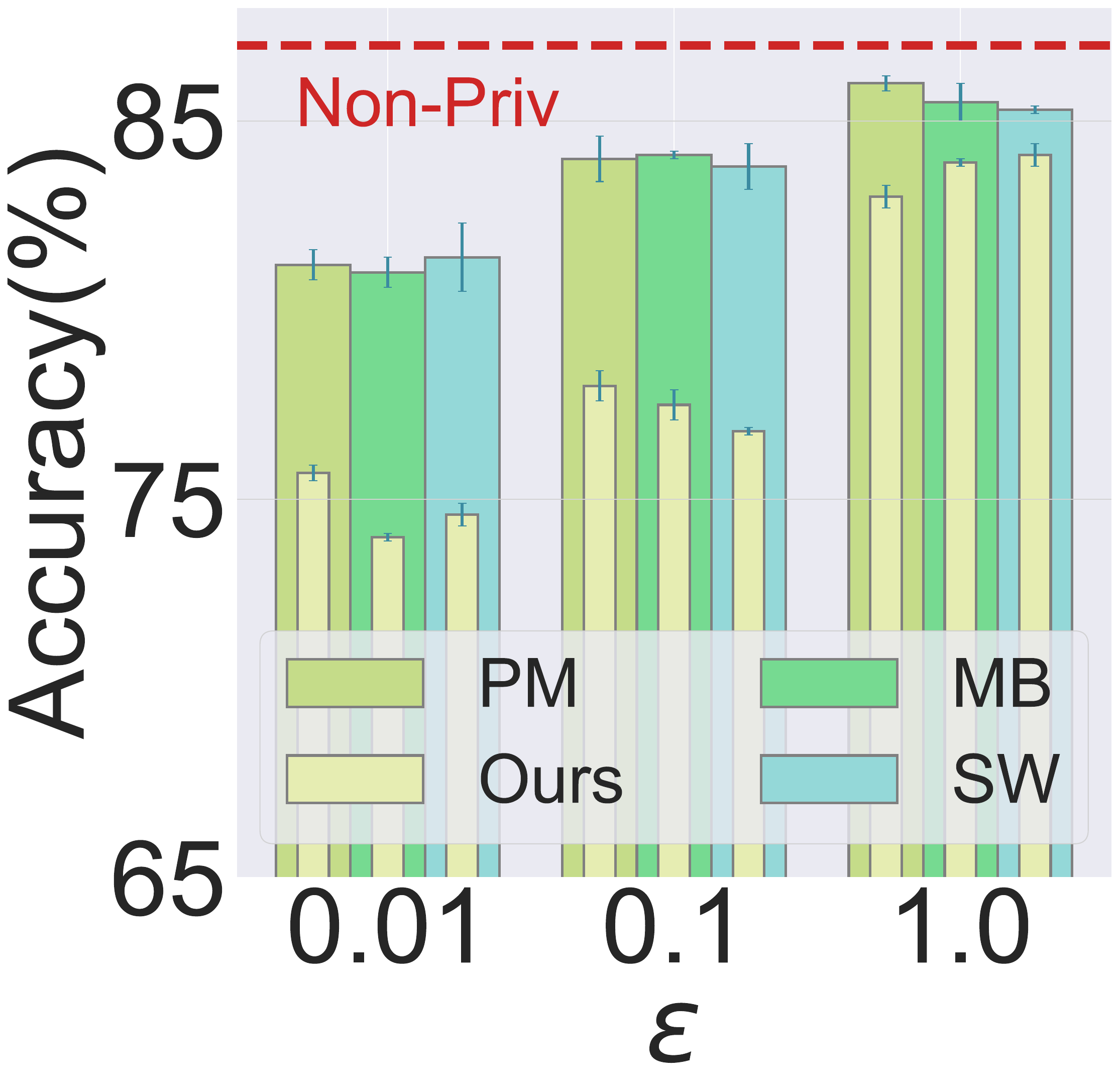}}
        \hspace{0.2cm}\subfigure[\normalsize Facebook]{\includegraphics[width=0.15\textwidth]{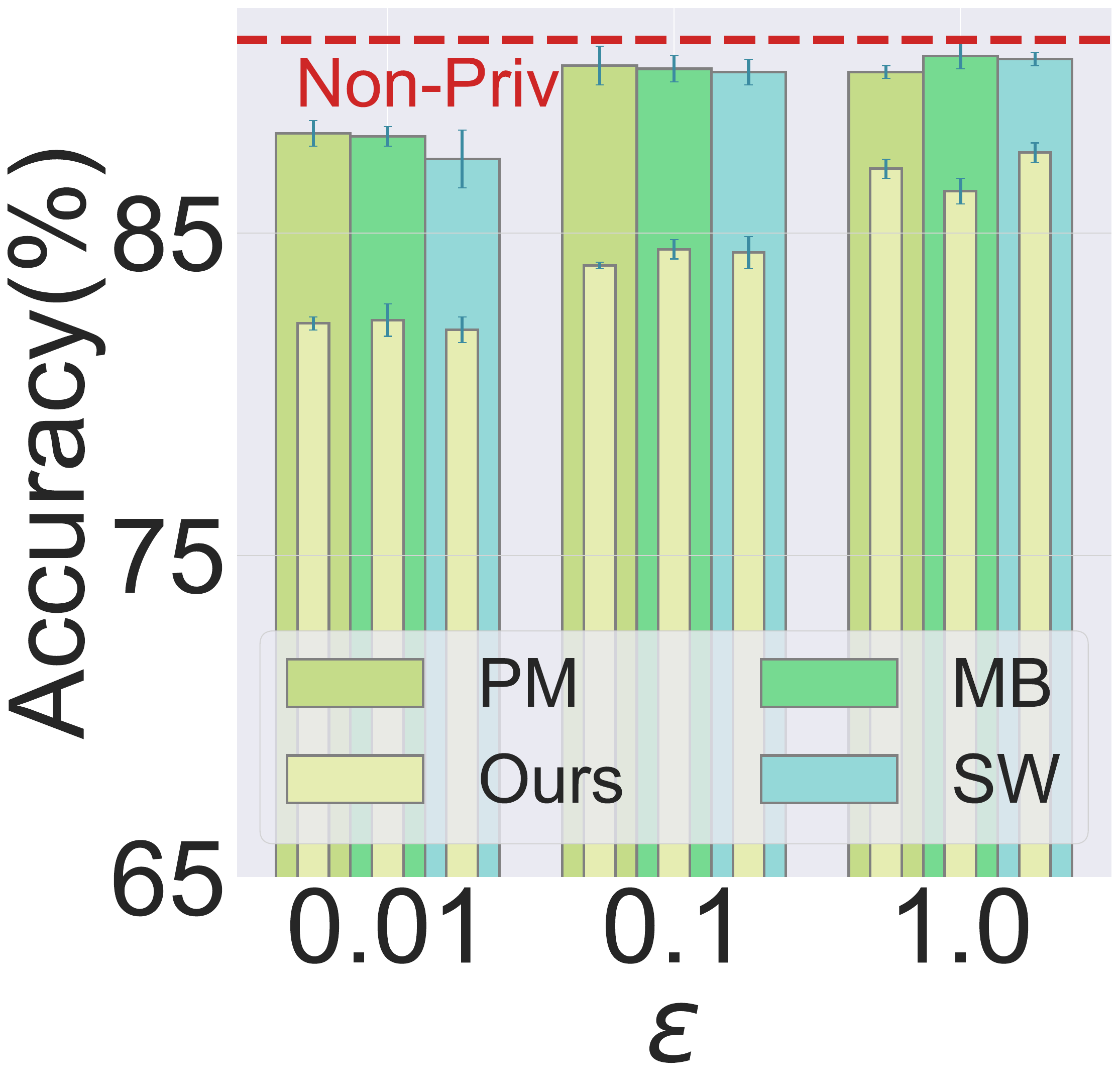}}
    \end{minipage}
     \vspace{-1em}
    \caption{Attack performance under different privacy budgets $\epsilon$ and LDP mechanisms for \textit{link prediction} task. The x-axis represents the privacy budget $\epsilon$, while the y-axis shows the accuracy for both the targeted and untargeted attacks. “Non-Priv” denotes the non-perturbed baseline. Our attacks significantly reduce the accuracy of various LDP mechanisms across all cases.}
    \label{fig4}
    \vspace{-0.6em}
\end{figure*}

In addition to node classification, we also evaluate the attack performance under different privacy budgets $\epsilon\in\{0.01,0.1,1.0\}$ and LDP mechanisms (PM, MB, and SW) using the link prediction task, with the results presented in Fig.~\ref{fig4}. As shown in the figure, our attack methods significantly reduce the accuracy of the different LDP mechanisms in this task setting as well, confirming the effectiveness of our attacks. Link prediction works by inferring the likelihood of a link between two nodes based on the similarity of their embeddings. Our attack compromises the embedding of the target node, which then propagates through the graph via malicious message passing, affecting the embeddings of neighboring nodes. This interference disrupts the node-to-node link prediction and results in a substantial decrease in utility. Furthermore, similar experimental trends to those observed in the node classification task are evident here, so these results are not repeated.

\begin{figure}[t]
	\centering
    \vspace{-1em}
	\begin{tabular}{cc}
		\multicolumn{2}{c}{\textbf{}} \\
  \includegraphics[width=0.45\linewidth]{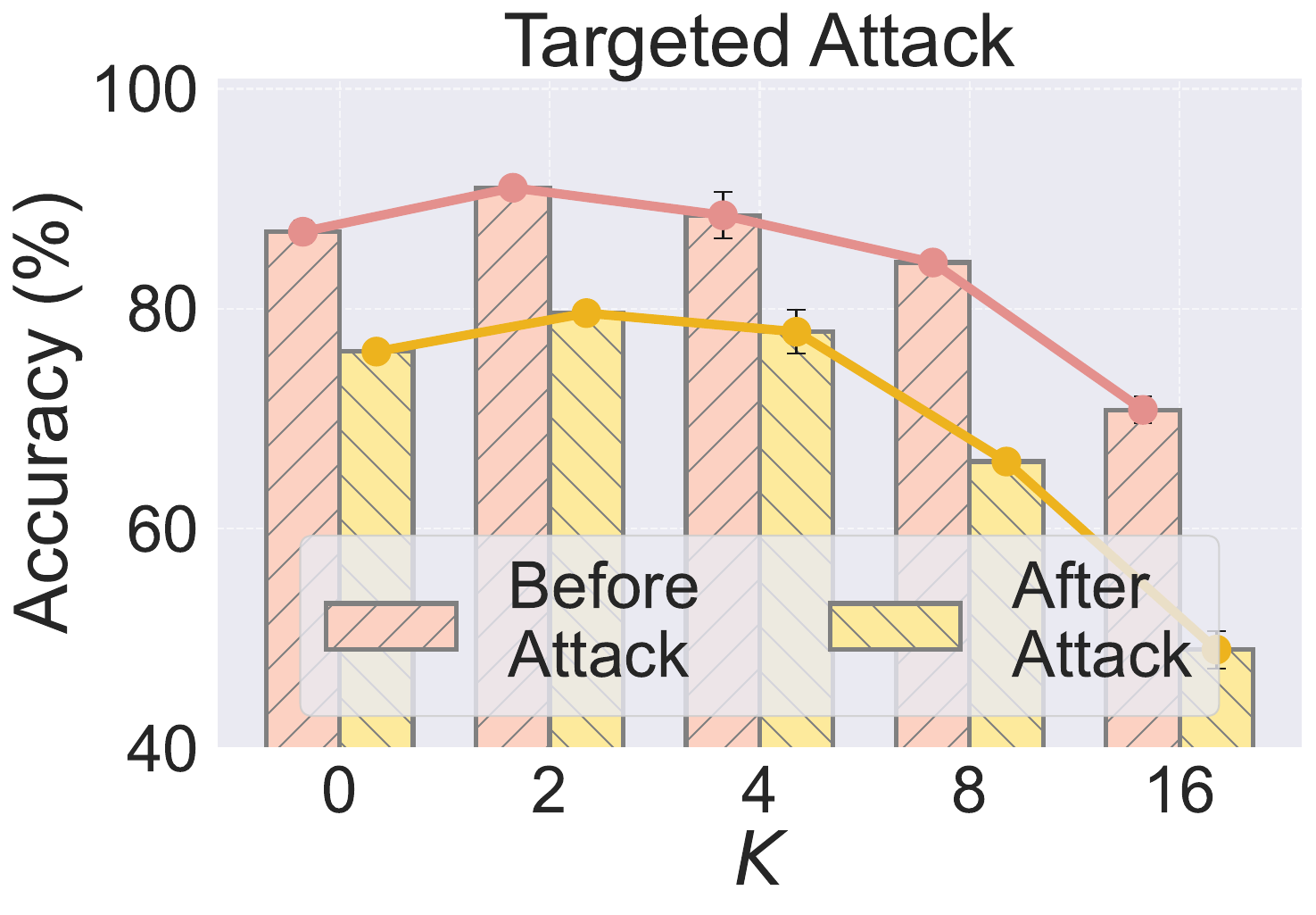} & \includegraphics[width=0.45\linewidth]{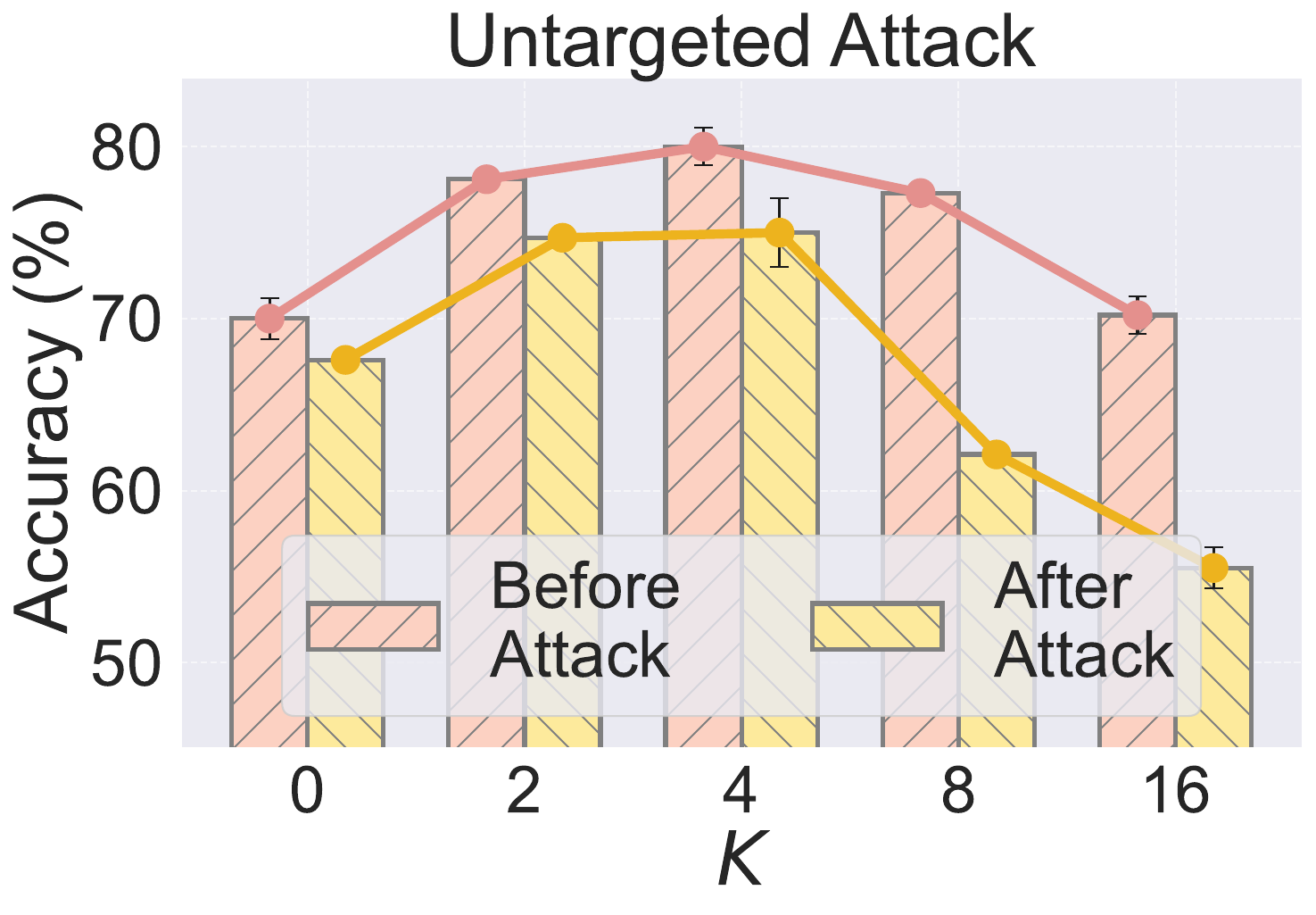}\\
	\end{tabular}
    \vspace{-1.5em}
	\caption{Variation in accuracy before and after targeted (left plot) or untargeted (right plot) attack for different $K$ values.}
    \vspace{-1em}
	\label{fig6}
\end{figure}

\subsection{Evaluating the Effectiveness of the Attack Across Different GNN Models}\label{sec6.3}
In this experiment, we evaluate the performance of the attack across six representative datasets using three GNN models (GCN, SAGE, GAT). As shown in the targeted attack results in Table~\ref{tab2}, our attack significantly reduces the accuracy of the original PM mechanism across different GNN models and datasets, with an average accuracy reduction of over 10\% for all three models. These results demonstrate the generalization and effectiveness of our attack method across various GNN models. Additional experimental results and analysis on different GNN models can be found in App.~\ref{e.2}.

\subsection{Parameter Analysis}
\subsubsection{Analyzing the effect of the parameter $\eta_1$}\label{6.4.1}
As shown in the left panel of Fig.~\ref{fig444}, the accuracy before attack remains relatively stable for different values of $\eta_1\in\{0.01,0.03,0.05,0.07,0.09\}$. However, as $\eta_1$ increases—meaning the number of target nodes the attacker can damage grows—the accuracy after the attack steadily decreases. Correspondingly, the impact ratio (the ratio of reduced accuracy to normal accuracy) increases, highlighting the expanding malicious impact of our attack. This occurs because as $\eta_1$ increases, the attacker can influence more genuine nodes, leading to greater disruption in the utility of the privacy-preserving learning process.

\subsubsection{Analyzing the effect of the parameter $\eta_2$}\label{6.4.2}
As shown in the right panel of Fig.~\ref{fig444}, we validate the effect of different $\eta_2$ on the attack, and the observed trend is similar to that of $\eta_1$. As $\eta_2$ increases, the attacker can inject more fake nodes, thereby introducing more malicious information. This results in greater damage to the private learning utility. For more details on $\eta_1$ and $\eta_2$, see App.~\ref{e.4}.

\subsubsection{Analyzing the effect of the parameter $K$}\label{6.4.3}
As shown in Fig.~\ref{fig6}, we validate the change in accuracy before and after the attack for different values of $K\in\{0,2,4,8,16\}$. As $K$ increases, accuracy follows a trend of first increasing and then decreasing. The initial increase is due to noise calibration through neighbor aggregation, while the subsequent decrease is caused by over-smoothing~\cite{li2018deeper,keriven2022not}, which leads to the convergence of node embeddings. Notably, our method significantly reduces the accuracy of the original method across all values of $K$, demonstrating the generalization and effectiveness of our approach.
Further analysis can be found in App~\ref{e.5}.

\begin{figure*}[htp]
    \centering
    \begin{minipage}{\textwidth}
        \centering
\subfigure[\normalsize Cora]{\includegraphics[width=0.16\textwidth]{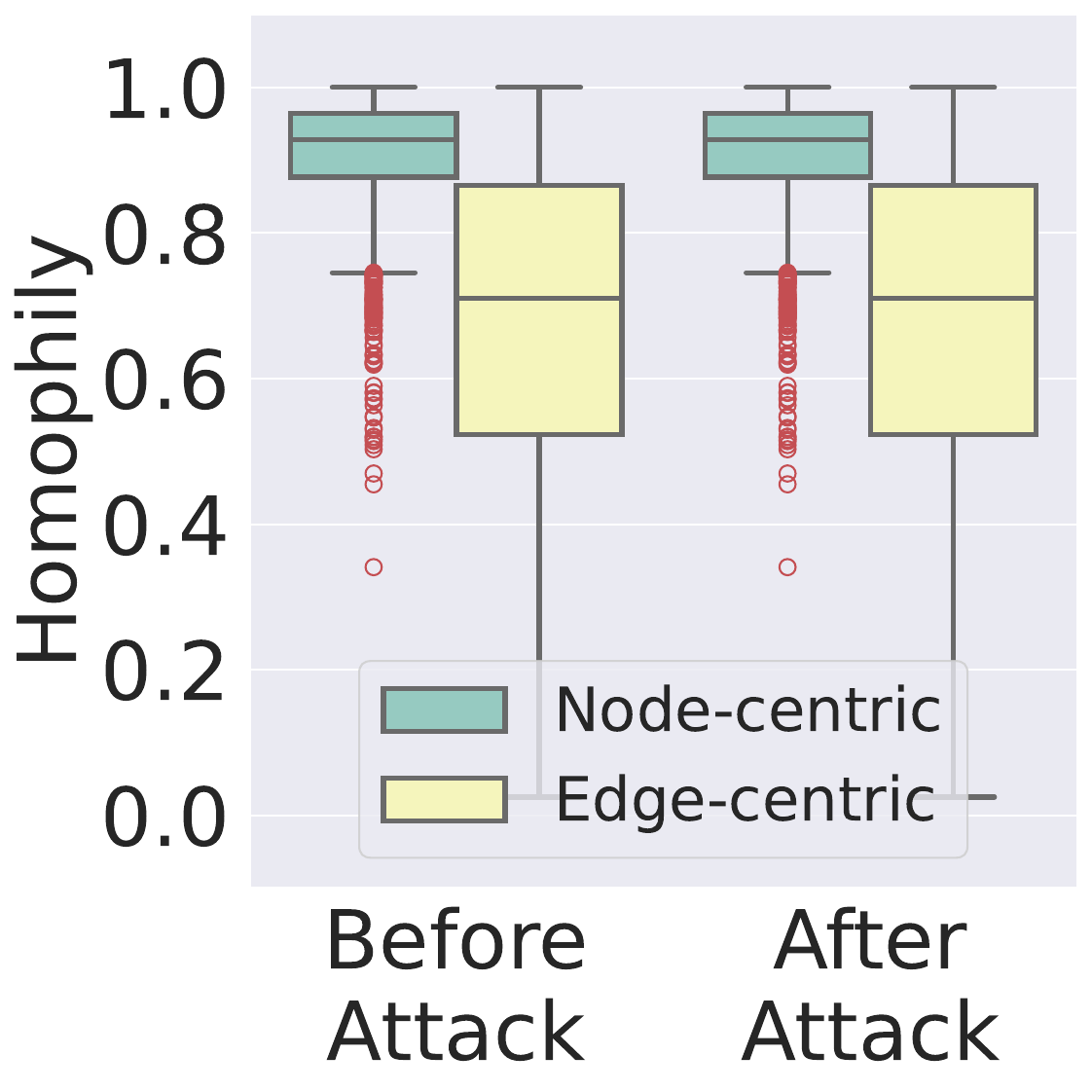}}
   \hspace{0.1em}\subfigure[\normalsize Citeseer]{\includegraphics[width=0.16\textwidth]{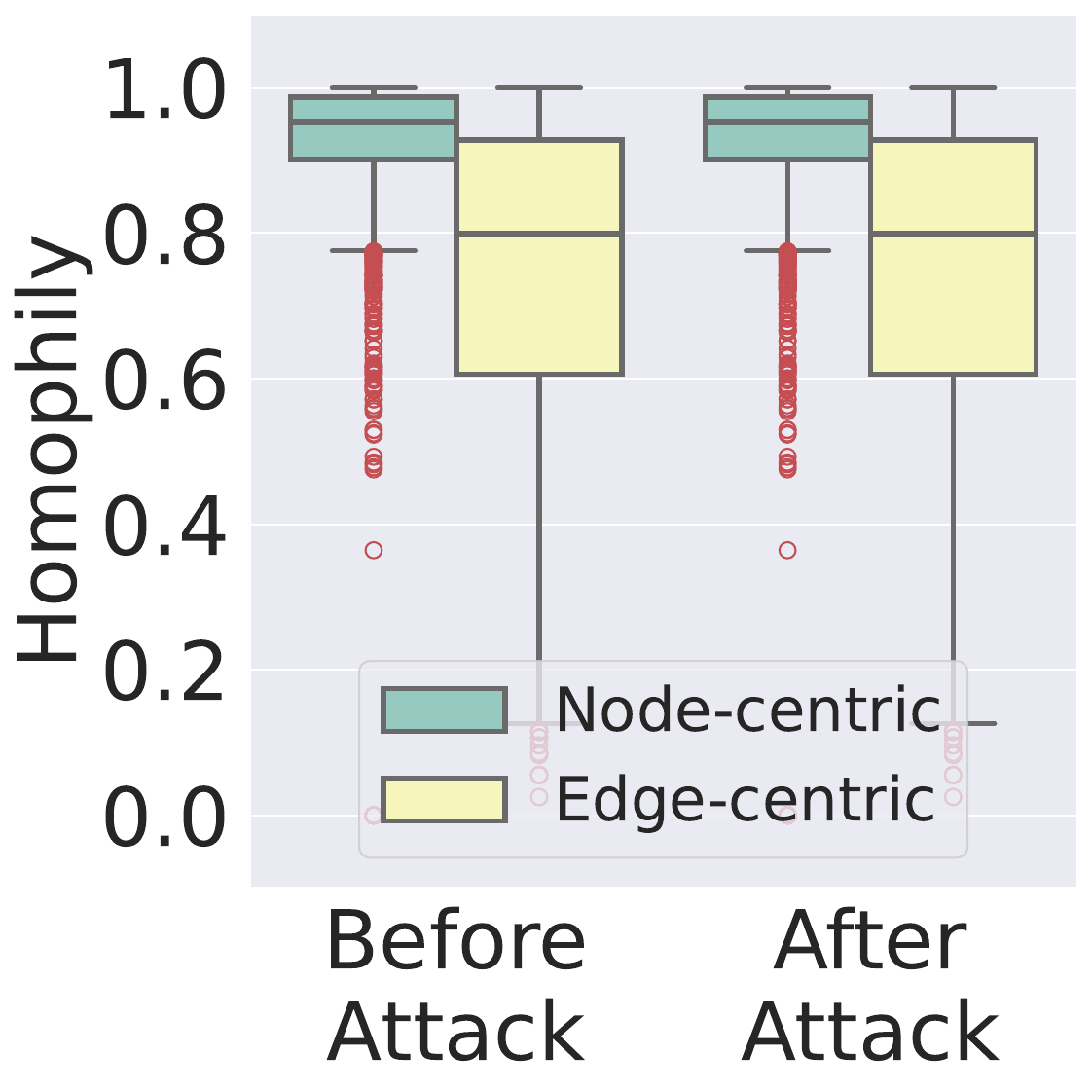}}
 \hspace{0.1em}\subfigure[\normalsize Pubmed]{\includegraphics[width=0.16\textwidth]{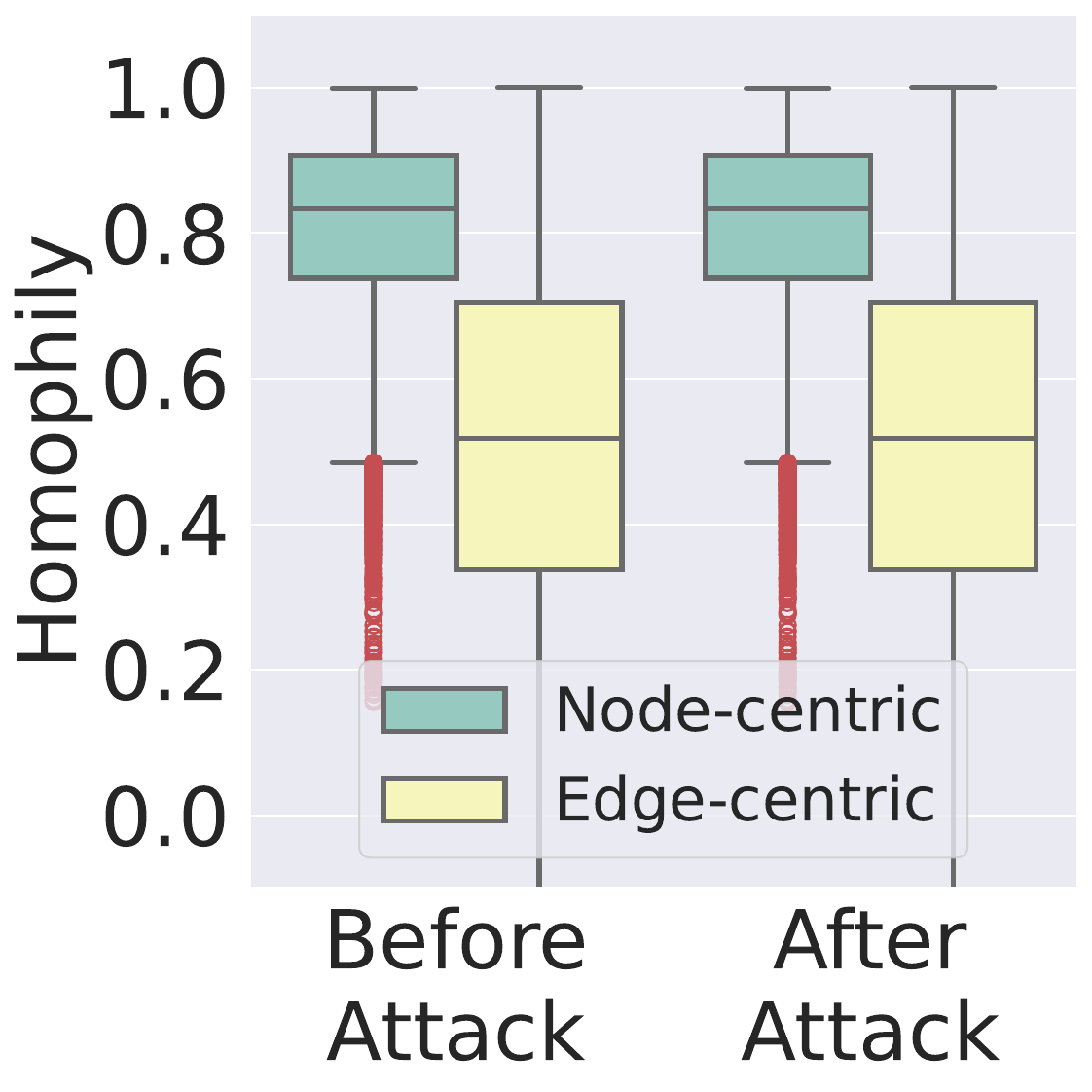}}
\hspace{0.1em}\subfigure[\normalsize LastFM]{\includegraphics[width=0.16\textwidth]{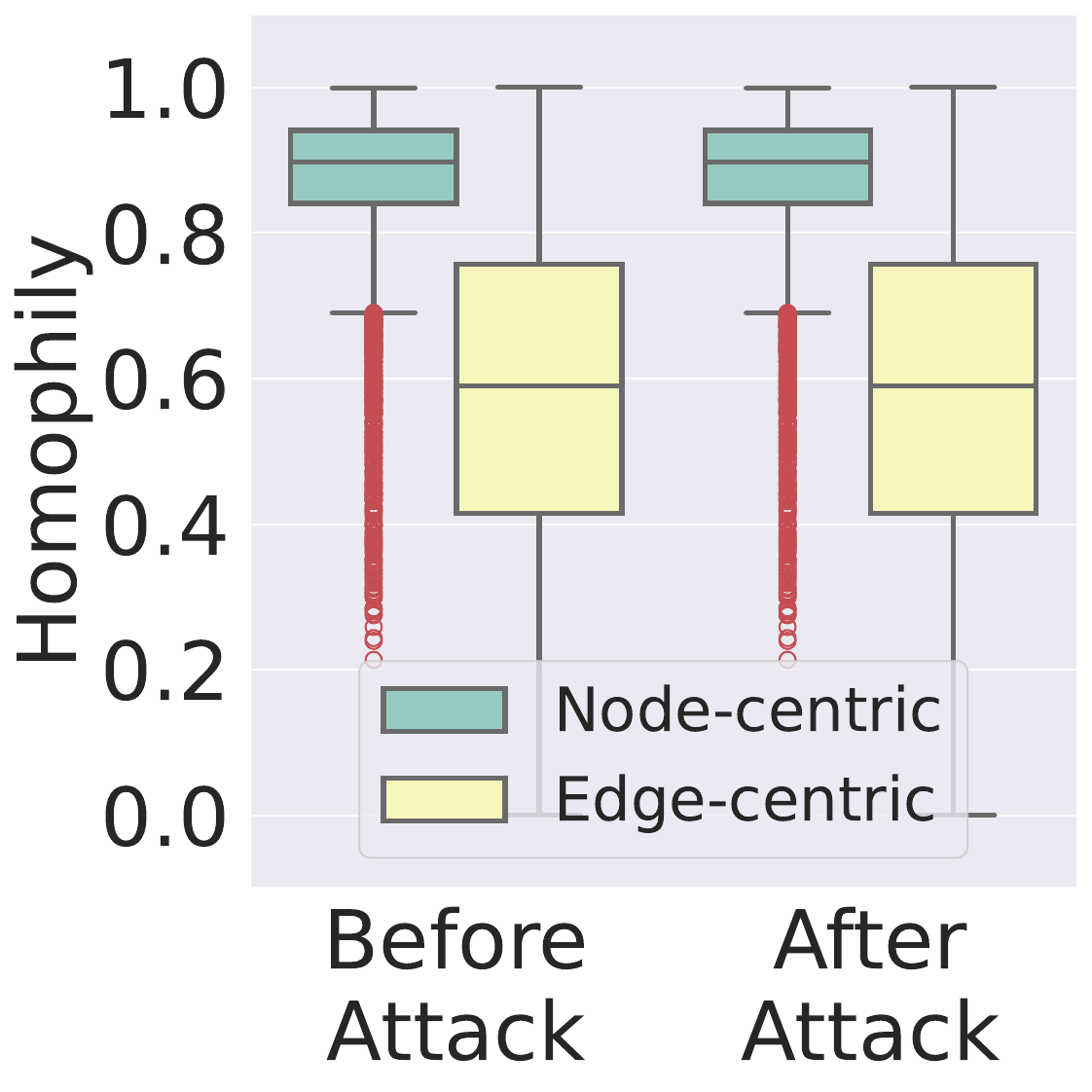}}
\hspace{0.1em}\subfigure[\normalsize Twitch]{\includegraphics[width=0.16\textwidth]{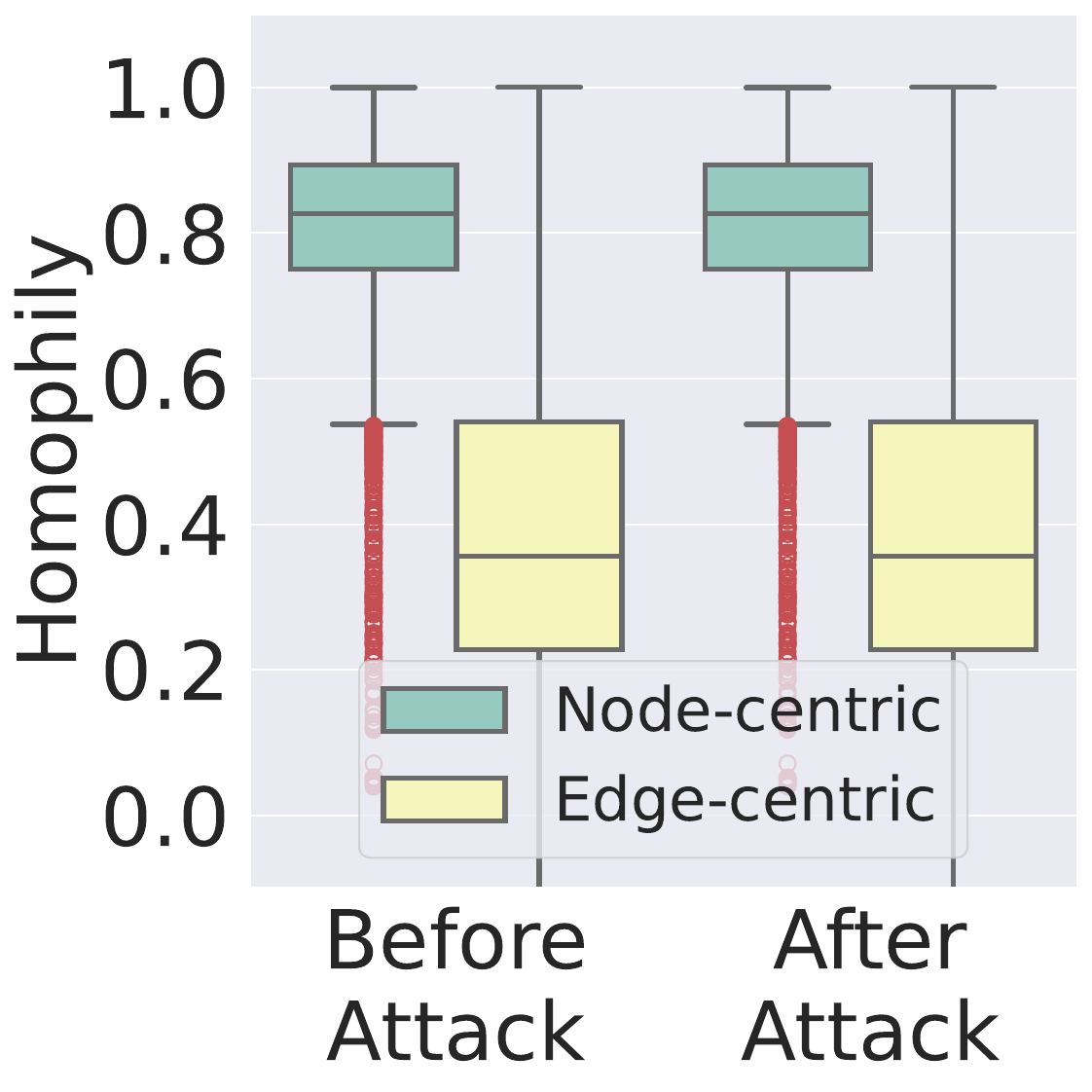}}
\hspace{0.1em}\subfigure[\normalsize Facebook]{\includegraphics[width=0.16\textwidth]{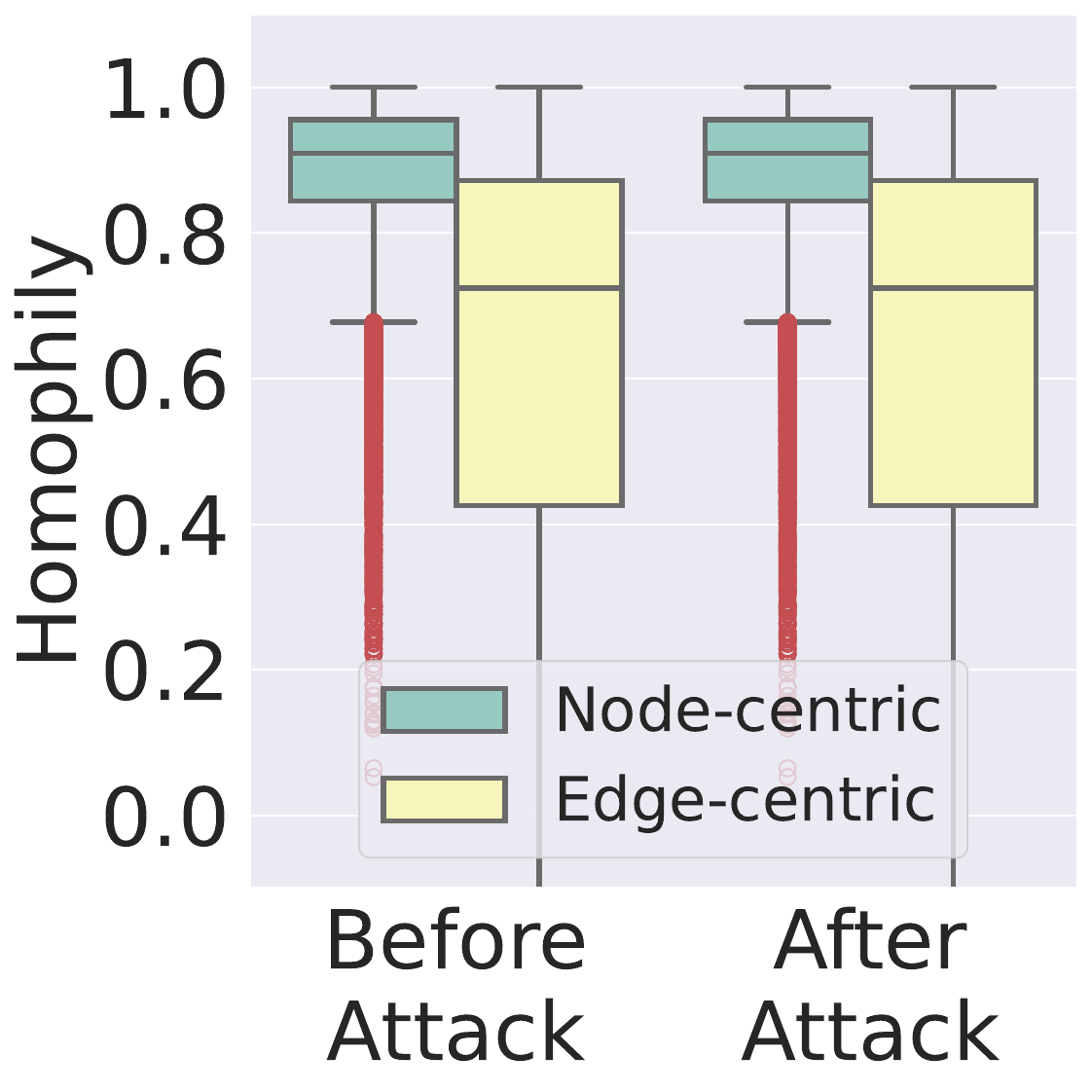}}
    \end{minipage}
    \vspace{-1em}
    \caption{Comparison of homophily distributions before and after the attack. The results show that before and after the attack do not exhibit significant deviations, which indicates that our attack method is highly resistant to homophily-based defense.}
    \vspace{-0.5em}
    \label{fig7}
\end{figure*}

\begin{figure}[t]
	\centering
    \vspace{-1em}
	\begin{tabular}{cc}
		\multicolumn{2}{c}{\textbf{}} \\
\hspace{-0.5em}\includegraphics[width=0.48\linewidth]{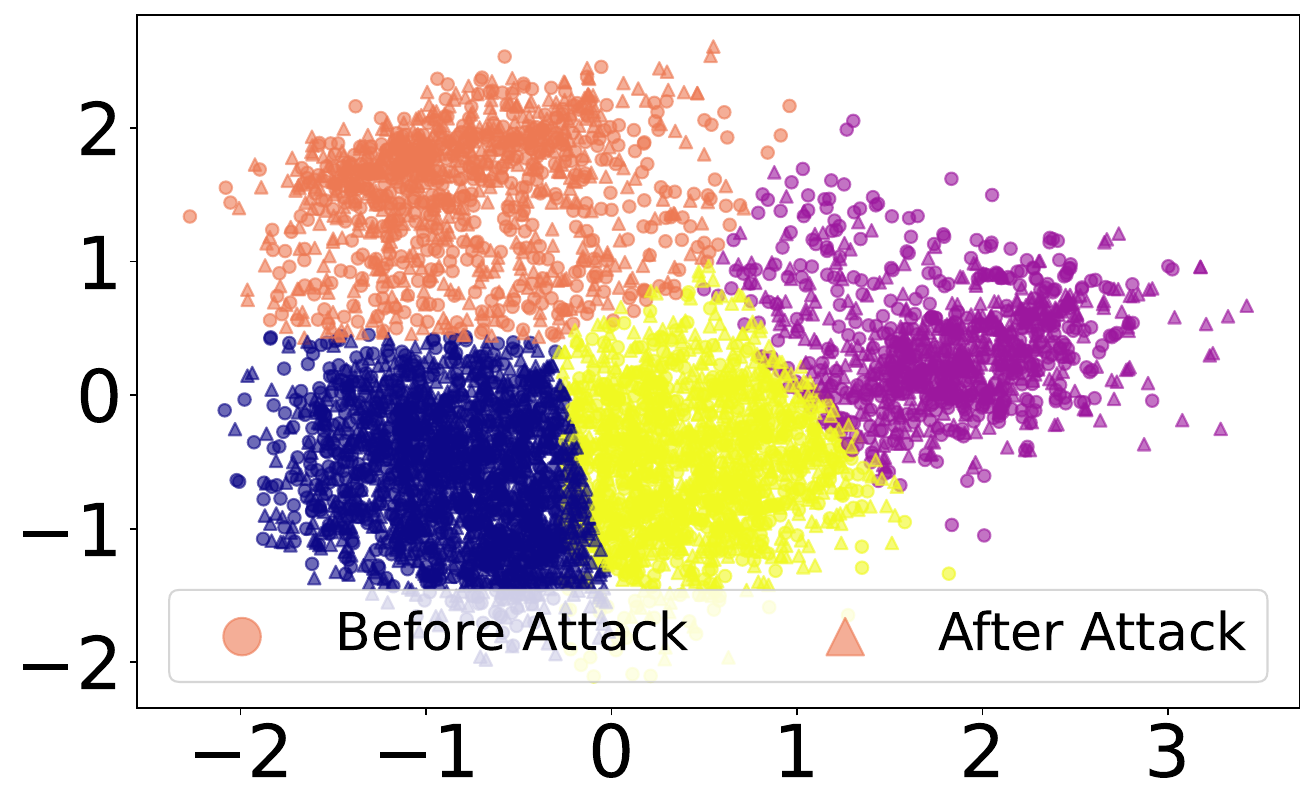} & \hspace{-0.2em}\includegraphics[width=0.48\linewidth]{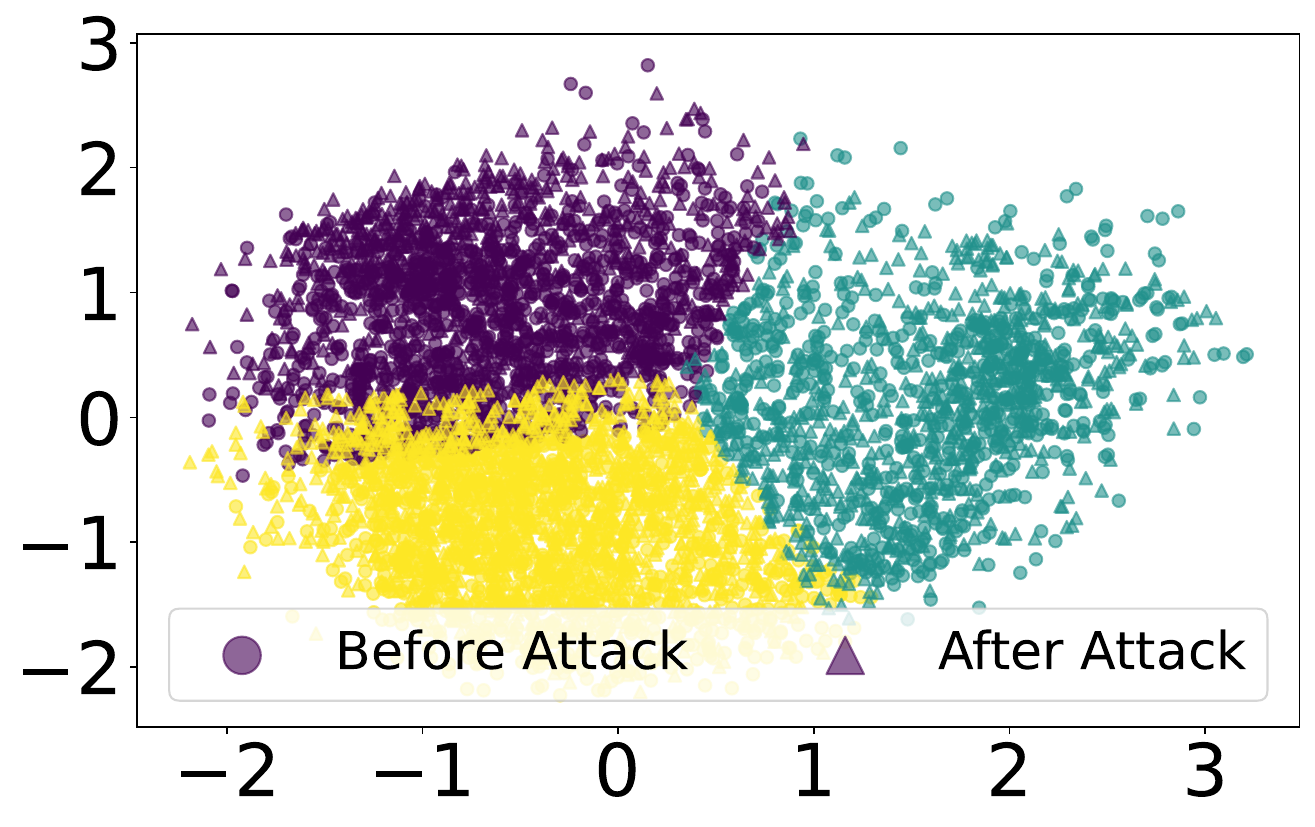}
	\end{tabular}
    \vspace{-1em}
	\caption{Anomaly node detection based on Girvan-Newman community discovery (left) and KMeans clustering (right).}
    \vspace{-0.8em}
	\label{fig8}
\end{figure}

\section{DEFENSES}\label{sec:def}
In this section, we evaluate the resilience of our attack approach against potential defenses. We sequentially consider two possible defenses: \textit{graph homophily analysis} (Sec.~\ref{sec7.1}) and \textit{anomaly node detection} (Sec.~\ref{sec7.2}). We also discuss other possible defenses in Sec.~\ref{sec7.3}.

\subsection{Graph Homophily Analysis}\label{sec7.1}
Existing studies~\cite{chenunderstanding} have shown that attacks~\cite{zou2021tdgia,zugner2018adversarial} against graphs can significantly disrupt the homophily distribution~\cite{mcpherson2001birds} of the original graph. To address this, defenses based on homophily analysis~\cite{chenunderstanding,zhang2020gnnguard,sun2022adversarial,zugner2020adversarial} have been proposed to detect such attacks. In this paper, we conduct graph homophily analysis on all datasets both before and after the attack. Specifically, we use two homophily metrics~\cite{chenunderstanding}: \textit{node-centered homophily}, which measures the similarity between a target node and its neighbors, and \textit{edge-centered homophily}, which compares the similarity between nodes connected by a target edge. See App.~\ref{app4.2} for more details on these metrics.

\textbf{Results.} We perform homophily analyses using cosine similarity, and Fig.~\ref{fig7} compares the homophily distributions of different datasets before and after the attack. As noted in~\cite{chenunderstanding}, homophily-based defenses can detect malicious nodes or edges with abnormally low homophily. However, as shown in the figure, our attack method does not cause significant deviations. This is because our carefully designed node features and links perturb the original graph on a much smaller scale than typical adversarial attacks~\cite{zou2021tdgia}. Additionally, Fig.~\ref{fig7} shows that node-centered homophily demonstrates higher similarity than edge-centered homophily across all datasets. This observation aligns with previous studies~\cite{zou2021tdgia,mcpherson2001birds,meng2023devil}. These results demonstrate that our attack method is highly resistant to homophily-based defenses.

\subsection{Anomaly Node Detection}\label{sec7.2}
We further evaluate the resistance of our attacks to existing anomaly detection methods~\cite{ma2021comprehensive}. Specifically, we employ two widely used techniques: \ding{172} \textit{Girvan-Newman Community Discovery}. The Girvan-Newman (GN) algorithm~\cite{girvan2002community} recursively divides communities by removing edges with the highest betweenness centrality. In a typical scenario, normal nodes maintain strong connectivity with others in their community, while malicious attacks can lead to misclassification or isolation of certain nodes, which appear as anomalies in the graph structure. \ding{173} \textit{Node Features Based on KMeans Clustering}. KMeans clustering~\cite{likas2003global} is an unsupervised learning method that partitions nodes into clusters. Anomalous nodes might either form outliers or be misassigned to incorrect clusters, making them detectable via feature-based anomaly detection.

\textbf{Results.} As shown in the left panel of Fig.~\ref{fig8}, our attack method does not introduce noticeable anomalies in community division, indicating that it maintains strong stealth under GN detection. Meanwhile, as shown in the right panel of Fig.~\ref{fig8}, the perturbation introduced by our attack in the feature space is minimal and does not significantly alter the node distribution, making it difficult to detect under KMeans evaluation. These anomaly node detection experiments further validate the stealthiness of our attack method. 

\vspace{-0.5em}
\subsection{Other Defenses}\label{sec7.3}
We also consider other potential defenses. First, this paper aims to compromise locally private graph learning protocols by injecting fake nodes and edges, and authenticated data feed systems may be able to help reestablish trust in the data source. However, current solutions are limited to well-known entities~\cite{zhang2016town}, and authenticating data from unknown sources in distributed privacy scenarios remains an open problem. Furthermore, while adversarial defense methods~\cite{chenunderstanding,zhang2020gnnguard,sun2022adversarial,zugner2020adversarial,zhu2019robust} may provide some mitigation, our experiments demonstrate that due to the careful design of node features and links, the perturbation scale of our attack is significantly smaller than that of adversarial attacks~\cite{zou2021tdgia,zugner2018adversarial}. Consequently, our attack exhibits strong resistance against adversarial defenders. Finally, other possible defenses include hardware-assisted trusted computation~\cite{cirne2024hardware,lee2020keystone,li2022design,li2024sok}, but this approach may introduce additional costs for hardware-software co-design and security auditing to prevent extensive side-channel attacks~\cite{murdock2020plundervolt,van2018foreshadow,schwarz2017malware}.

\section{Related work}\label{sec:rek}
\subsubsection{Data Poisoning Attacks on GNNs}
Data poisoning attacks aim to disrupt the learning process of machine learning models by injecting malicious data into the training set. These attacks~\cite{zou2021tdgia,zugner2018adversarial,sun2022adversarial,zugner2020adversarial} are especially concerning in the context of GNNs due to the complex dependencies between nodes, edges, and graph structures that GNNs rely on to learn node representations. As discussed in Sec.~\ref{sec1}, the data poisoning attack introduced in this paper, specific to locally private graph learning protocols, differs fundamentally from previous work on traditional attacks against GNNs. This distinction primarily arises from the strict data constraints, limited background knowledge, and the complexity of the protocols involved.

\subsubsection{Locally Private Graph Learning Protocols}. Recently, locally private graph learning protocols have attracted significant attention in the security research community~\cite{sajadmanesh2021locally,lin2022towards,pei2023privacy,li2024privacy,qi2024linkguard,zhu2023blink,hidano2022degree,zhang2024locally}. These approaches leverage the benefits of LDP~\cite{yang2024local,dwork2006calibrating} in safeguarding data privacy and ensuring strict privacy guarantees for user data. Moreover, during server-side privacy graph learning, GNN's multi-layer message passing mechanism
~\cite{feng2022powerful,abu2019mixhop} effectively reduces the estimation error in node embeddings after perturbation and aids in the calibration of noisy data, thus maintaining high utility for the local privacy graph learning task. However, despite these advantages, such protocols may be vulnerable to data poisoning attacks. More details on this section are in App~\ref{ddd}.

\section{CONCLUSION}\label{sec:con}
In this paper, we present the first study on data poisoning attacks targeting locally private graph learning protocols, with the aim of compromising the utility of private graph learning within these protocols. Through theoretical analysis and empirical evaluation, we demonstrate that our attacks significantly degrade the utility of private graph learning. Additionally, we examine several potential defense strategies to mitigate these attacks, but our findings underscore the urgent need for more robust countermeasures. We believe this work is essential for advancing the development of secure and resilient privacy-preserving graph learning frameworks.


\bibliographystyle{ACM-Reference-Format}
\bibliography{sample-base}

\appendix

\section{LDP mechanisms}\label{app2}
In Sec.~\ref{2.3.1}, we introduce three different LDP mechanisms for node feature perturbation. To enhance clarity and provide a structured understanding, we present a generalized algorithmic description using the piecewise mechanism~\cite{wang2019collecting,pei2023privacy} as a representative example, as outlined in Alg.~\ref{alg:1}. In the one-dimensional piecewise mechanism, the input domain is $[\alpha, \beta]$, and the perturbed data range is $[-s,s]$, where $s=\frac{e^{\epsilon/2}+1}{e^{\epsilon/2}-1}$. Given an original value $x$, the perturbed value $x^\prime$ is sampled from the following probability density function:
\begin{equation}\label{app1eq}
	\mathrm{Pr}[x^\prime=c|x] = 
	\begin{cases}
		p,   &\text{if } c\in[l(x), r(x)]  \\
		p/e^\epsilon, &\text{if } c\in[-s, l(x))\cup (r(x), s] 
	\end{cases},
\end{equation}
where $p=\frac{e^\epsilon -e^{\epsilon/2 }}{2e^{\epsilon/2 }+2}$, $l(x)=\frac{s+1}{2} \cdot x-\frac{s-1}{2}$, and $r(x)=l(x)+s-1$.
\begin{algorithm}[h]
	\small
	\SetKwInOut{Input}{Input}\SetKwInOut{Output}{Output}
	\LinesNumbered
	\KwIn{node feature vector $\mathbf{x}\in[\alpha,\beta]^d$ and privacy budget $\epsilon$.
	}
	\KwOut{perturbed node feature vector $\mathbf{x}^\prime\in[-s\cdot d,s\cdot d]^d$.}
	\DontPrintSemicolon
	\caption{Piecewise Mechanism For Node Feature}
	\label{alg:1}
		Let $\mathbf{x}^\prime=\langle   0,0,\cdots,0\rangle$;\;
        Let $m=\max  \{ 1,\min \{  d, \lfloor \epsilon /2.5 \rfloor  \}   \} $;\;
        Sample $m$ values uniformly without replacement from $\{1,2,\hspace{-0.1em}\cdots\hspace{-0.1em},\hspace{-0.1em}d\};$\;
		\For{each sampled dimension $i$} {
			Feed $\mathbf{x}_{i}$ and $\frac{\epsilon}{m}$ as input to Eq.~(\ref{app1eq}), and obtain
a noisy value $t_{i}$;\;
			$\mathbf{x}^\prime_{i}=\frac{d}{m}\cdot t_{i};$\;
		}
	\Return $\mathbf{x}^\prime$\;
\end{algorithm}
\section{Theoretical Proof}
\subsection{Proof of Proposition~\ref{prop1}}\label{app2.1}
\begin{proof}
To ensure that the degree remains constant before and after the attack, we have the following equation for the average degree of the original graph:
\begin{equation}\label{b11}
    \langle d_{\mathrm{orig.}}\rangle=2\left ( \left | \mathcal{E} \right | +  \left | \mathcal{V}_t \right |+\binom{\left |\mathcal{V}_{ \mathrm{atk} } \right | }{2}\cdot q  \right )/\left ( \left | \mathcal{V} \right |+ \left |\mathcal{V}_{ \mathrm{atk} } \right | \right ), 
\end{equation}
where $\langle d_{\mathrm{orig.}}\rangle $ represent the average degree of the original graph $\mathcal{G}$, $|\mathcal{V}|$ and $|\mathcal{E}|$ denote the number of nodes and edges in the graph $\mathcal{G}$, respectively. $\left |\mathcal{V}_t \right |$ denotes the number of target nodes. $\left |\mathcal{V}_{ \mathrm{atk} } \right |$ denotes the number of fake nodes. Additionally, we have the following relationship for the number of edges in the original graph:
\begin{equation}\label{b22}
\left | \mathcal{E} \right |=\left ( \langle d_{\mathrm{orig.}}\rangle\cdot \left | \mathcal{V} \right | \right )/2.   
\end{equation}
Substituting this into the first equation, we have
\begin{equation}
       \langle d_{\mathrm{orig.}}\rangle=\frac{2\cdot\left ( \frac{ \langle d_{\mathrm{orig.}}\rangle\cdot \left | \mathcal{V} \right |}{2}   +  \left | \mathcal{V}_t \right |+\binom{\left |\mathcal{V}_{ \mathrm{atk} } \right | }{2}\cdot q  \right )}{\left | \mathcal{V} \right |+ \left |\mathcal{V}_{ \mathrm{atk} } \right | }. 
\end{equation}
Solving the above for $q$, we get:
\begin{equation}
q=\frac{\langle d_{\mathrm{orig.}}\rangle-2\cdot|\mathcal{V}_t|/|\mathcal{V}_{\text{atk}}|}{|\mathcal{V}_{\text{atk}}|-1} 
\end{equation}
where \( 1<|\mathcal{V}_{\text{atk}}| \le|\mathcal{V}_t| \). Considering the reasonable range of $q$,  the above equation should satisfy the following range restriction:
\begin{equation}
     0<\frac{\langle d_{\mathrm{orig.}}\rangle-2\cdot|\mathcal{V}_t|/|\mathcal{V}_{\text{atk}}|}{|\mathcal{V}_{\text{atk}}|-1}\le1. 
\end{equation}
To satisfy $q \le 1$, we have
\begin{equation}
    \langle d_{\mathrm{orig.}}\rangle-2\cdot|\mathcal{V}_t|/|\mathcal{V}_{\text{atk}}|\le|\mathcal{V}_{\text{atk}}|-1.
\end{equation}
Rearranging this inequality:
\begin{equation}
    \left ( |\mathcal{V}_{\text{atk}}| \right )^2-\left ( 1+\langle d_{\mathrm{orig.}}\rangle \right )\cdot |\mathcal{V}_{\text{atk}}|+2\cdot|\mathcal{V}_t|\ge 0.  
\end{equation}
Providing that $\left ( 1+\langle d_{\mathrm{orig.}}\rangle \right )^2-8|\mathcal{V}_t|\le 0$, we have the following bound for $|\mathcal{V}_t |$,
\begin{equation}\label{b2222}
    |\mathcal{V}_t |\ge (\langle d_{\mathrm{orig.}}\rangle+1)^2/8.
\end{equation}
Furthermore, to satisfy $q > 0$, we have
\begin{equation}\label{b4444}
|\mathcal{V}_{\text{atk}}|>2|\mathcal{V}_t|/\langle d_{\mathrm{orig.}}\rangle.
\end{equation}
Consider $|\mathcal{V}_{\text{atk}}| \le|\mathcal{V}_t|$, we have
\begin{equation}\label{b3333}
   \langle d_{\mathrm{orig.}}\rangle\ge2.  
\end{equation}
It should be noted that Eq.~(\ref{b2222}), Eq.~(\ref{b4444}) and Eq.~(\ref{b3333}) are easily satisfied in realistic graph learning scenarios, as can be verified in the experimental setup.
\end{proof}

\subsection{Proof of Proposition~\ref{prop2}}\label{app2.2}
First, we provide a brief description of Definition~\ref{def33}. Consider the first layer of aggregation operations as follows:
\begin{equation}
\widehat{\mathbf{h}}_{v}=\textsc{Aggregate}(\{\mathbf{x}_u^\prime, \forall u \in \mathcal{N}(v)\}),
\end{equation}
so, we have:
\begin{equation}	\mathbb{E}\left[\widehat{\mathbf{h}}_v\right] 
 = \mathbb{E}\left[\textsc{Aggregate}\left(\{\mathbf{x}_u^\prime, \forall u \in \mathcal{N}(v)\}\right)\right].
\end{equation}
Since $\textsc{Aggregate}(\cdot)$ is linear, we have:
\begin{equation}
\mathbb{E}\left[\widehat{\mathbf{h}}_v\right] 
= \textsc{Aggregate}\left(\{\mathbb{E}\left[\mathbf{x}_u^\prime\right], \forall u \in \mathcal{N}(v)\}\right).
\end{equation}
Considering $\mathbb{E}[\mathbf{x}^\prime_u]=\mathbf{x} _u$ and hence have:
\begin{equation}	\mathbb{E}\left[\widehat{\mathbf{h}}_v\right] 
= \textsc{Aggregate}\left(\{\mathbf{x}_u, \forall u \in \mathcal{N}(v)\}\right)=\mathbf{h}_v.
\end{equation}
Therefore, provided that the perturbed node features remain unbiased, the subsequent aggregator also provides an unbiased estimate. Since our feature crafting method follows a uniform distribution, it does not destroy the unbiased property.

Then, we analyze the variance increment caused before and after the attack under the consideration of the two-layer aggregation operation. For the \textit{before attack}, consider the mean aggregation function, the first layer embedding of the target node $v_t$ is:
\begin{equation}
\widehat{\mathbf{h}}_{v_t}^{(1)} = \frac{1}{|\mathcal{N}({v_t})| + 1} \left( \mathbf{x}^\prime_{v_t} +\textstyle \sum_{i \in \mathcal{N}({v_t})} \mathbf{x}^\prime_i \right),
\end{equation}
where $\mathbf{x}^\prime_{v_t}$ denotes the perturbed node feature vector of node $v_t$, $\mathcal{N}({v_t})$ denotes the set of genuine neighbor nodes of $v_t$. Further, consider $\text{Cov}(\mathbf{x}^\prime_{v_t}, \mathbf{x}^\prime_i)=0$, then the variance of $\widehat{\mathbf{h}}_{v_t}^{(1)}$ is
\begin{equation}
    \text{Var}(\widehat{\mathbf{h}}_{v_t}^{(1)}) = \frac{1}{(|\mathcal{N}({v_t})| + 1)^2} \left( \text{Var}(\mathbf{x}^\prime_{v_t}) + \textstyle\sum_{i \in \mathcal{N}(v_t)} \text{Var}(\mathbf{x}^\prime_i) \right)
\end{equation}
Assuming that the variance of the node features perturbed by the LDP mechanism is $\sigma^2$, then
\begin{equation}
    \text{Var}(\widehat{\mathbf{h}}_{v_t}^{(1)}) = \frac{\sigma^2+ |\mathcal{N}({v_t})|\cdot \sigma^2}{(|\mathcal{N}({v_t})| + 1)^2}= \frac{\sigma^2}{|\mathcal{N}(v)| + 1}.
\end{equation}
Further, considering the second layer of aggregation, we have
\begin{equation}
    \widehat{\mathbf{h}}_{v_t}^{(2)} = \frac{1}{|\mathcal{N}(v_t)| + 1} \left( \widehat{\mathbf{h}}_{v_t}^{(1)} + \textstyle\sum_{i \in \mathcal{N}(v_t)} \widehat{\mathbf{h}}_i^{(1)} \right).
\end{equation}
Then, its variance is
\begin{equation}
    \text{Var}(\widehat{\mathbf{h}}_{v_t}^{(2)}) = \frac{ \text{Var}(\widehat{\mathbf{h}}_{v_t}^{(1)})}{|\mathcal{N}(v_t)| + 1}= \frac{ \sigma^2}{(|\mathcal{N}(v_t)| + 1)^2}.
\end{equation}
For \textit{after attack}, the first layer embedding of the node $v_t$ is:
\begin{equation}
\overline{\mathbf{h}}_{v_t}^{(1)}=\frac{1}{\left | \mathcal{N}(v_t) \right | +2 } \left ( \mathbf{x}_{v_t}^\prime+ \mathbf{x}^\prime _{v_{\mathrm{atk} }}+\textstyle \sum_{i\in \mathcal{N}(v) } \mathbf{x}^\prime _i   \right )   
\end{equation}
Assuming that the variance of the malicious node features is $\sigma_{\text{atk}}^2$, then
\begin{equation}
  \text{Var}(\overline{\mathbf{h}}_{v_t}^{(1)}) = \frac{\left ( |\mathcal{N}(v)|+1 \right )  \cdot \sigma^2 + \sigma_{\text{atk}}^2}{(|\mathcal{N}(v)| + 2)^2}. 
\end{equation}
In addition, the first layer embedding of the node $v_{\text{atk}}$ is:
\begin{equation}
\overline{\mathbf{h}}_{v_{\mathrm{atk} }}^{(1)}=\frac{1}{q\cdot\left ( |\mathcal{V}_{\mathrm{atk} } |-1 \right ) +2} \left ( \mathbf{x}^\prime _{v_t}+ \mathbf{x}_{v_{\text{atk}}}^\prime+\textstyle \sum_{i=1}^{q\cdot\left ( |\mathcal{V}_{\mathrm{atk} } |-1 \right )} \mathbf{x}^\prime _i   \right ).
\end{equation}
Let $N_{\text{atk}}=q\cdot\left ( |\mathcal{V}_{\mathrm{atk} } |-1 \right )$, then, the variance of $\overline{\mathbf{h}}_{v_{\mathrm{atk} }}^{(1)}$ is
\begin{equation}
    \mathrm{Var} \left ( \overline{\mathbf{h}}_{v_{\mathrm{atk} }}^{(1)} \right ) =\frac{\sigma^2+\left (N_{\text{atk}} +1 \right )\cdot\sigma_{\mathrm{atk} }^2+2\sum_{i=1}^{N_{\text{atk}} } \text{Cov}(\mathbf{x}_{v_t}^\prime ,\mathbf{x}^\prime _i  )}{\left ( N_{\text{atk}} +2 \right )^2 }.
\end{equation}
Further, considering the second layer of aggregation, we have
\begin{equation}
\overline{\mathbf{h}}_{v_t}^{(2)}=\frac{1}{\left | \mathcal{N}(v_t) \right | +2 } \left ( \overline{\mathbf{h}}^{(1)}  _{v_t}+\overline{\mathbf{h}}^{(1)}  _{v_{\mathrm{atk} }}+\textstyle \sum_{i\in\left | \mathcal{N}(v_t) \right |}\overline{ \mathbf{h}} ^{(1)} _i   \right )   
\end{equation}
Then, its variance is
\begin{equation}
\text{Var}(\overline{\mathbf{h}}_{v_t}^{(2)}) = \frac{(|\mathcal{N}(v_t)|+1) \cdot\text{Var}(\overline{\mathbf{h}}_{v_t}^{(1)}) + \overline{\mathbf{h}}_{v_{\mathrm{atk} }}^{(1)}}{(|\mathcal{N}(v_t)| + 2)^2}.
\end{equation}
To summarize, we have
\begin{align}
    \mathbb{ E}&[\Delta_{\text{Var}} ] =\mathbb{ E}\left [ \text{Var}(\overline{\mathbf{h}}_{v_t}^{(2)}) -\text{Var}(\widehat{\mathbf{h}}_{v_t}^{(2)})\right ] \\&=\frac{(|\mathcal{N}(v_t)|+1)^2 \cdot\sigma^2}{(|\mathcal{N}(v_t)| + 2)^4}+\frac{\sigma^2+\left (N_{\text{atk}} +1 \right )\cdot\sigma_{\mathrm{atk} }^2}{(|\mathcal{N}(v_t)| + 2)^2\cdot \left ( N_{\text{atk}} +2 \right )^2 }
    \\ &\quad-\frac{ \sigma^2}{(|\mathcal{N}(v_t)| + 1)^2}>0.
\end{align}
The above process quantifies the error estimation under the consideration of identical feature crafting approach, which is conducive to maximize the damage to the data utility of the target node.
\begin{table}[t]
	\centering
	\caption{Comparison of accuracy before and after the attack across different GNN models. ↓ represents accuracy drop.}
    \vspace{-0.5em}
	\label{tab:ldpp}
	\resizebox{\linewidth}{!}{\begin{tabular}{c|l|llllll}
		\toprule
		GNN  & Meth. & Cora     & CiteSeer & Pubmed & LastFM  & Twitch & Facebook         \\
		\midrule
		 \multirow{2}*{GCN}   
		& \textsc{pm}    & 77.3  & 64.5 & 79.1& 82.5& 53.4&88.4       \\
		& \textsc{pm}$\ssymbol{2}$     & 62.1 \roundbox{↓}  & 60.4 \roundbox{↓} & 74.3 \roundbox{↓} & 77.2 \roundbox{↓} & 51.6 \roundbox{↓} & 83.2 \roundbox{↓}    \\		
		\midrule
        \multirow{2}*{SAGE}   
		& \textsc{pm}       & 75.3  &64.9 & 76.7& 80.5& 54.7 & 83.7       \\
		& \textsc{pm}$\ssymbol{2}$     & 65.6 \roundbox{↓}  & 59.1 \roundbox{↓} & 70.6 \roundbox{↓} & 71.4 \roundbox{↓} & 52.8 \roundbox{↓} & 76.9 \roundbox{↓}    \\
		\midrule
		 \multirow{2}*{GAT}   
		& \textsc{pm}       & 78.9  & 65.3 & 77.5& 81.0& 56.2 & 83.2       \\
		& \textsc{pm}$\ssymbol{2}$     & 64.4 \roundbox{↓}  & 60.6 \roundbox{↓} & 75.1 \roundbox{↓} & 62.9 \roundbox{↓} & 54.5 \roundbox{↓} & 81.0 \roundbox{↓}    \\
		\bottomrule
	\end{tabular}}
    \begin{tablenotes}
     \item[1] $\ssymbol{2}$ represents the accuracy after untargeted attack.
   \end{tablenotes}
   \vspace{-1.5em}
 \label{tab3}
\end{table}
\subsection{Global Error Increment}\label{b.3}

In this section, we derive the global error increment \( \Delta\Psi \) that quantifies the distortion of node representations caused by the attack. The global error energy function before the attack is defined as:
\[
\Psi(\mathcal{G}) = \sum_{v \in \mathcal{V}} \| \mathbf{h}_v^{(K)} - \mathbf{x}_v \|^2 + \lambda \sum_{(u, v) \in \mathcal{E}} \| \mathbf{h}_u^{(K)} - \mathbf{h}_v^{(K)} \|^2,
\]
where \( \mathbf{h}_v^{(K)} \) represents the feature of node \( v \) after \( K \)-layer aggregation, and \( \mathbf{x}_v \) is the original feature of node \( v \). The parameter \( \lambda \) controls the balance between node similarity and edge similarity. After the attack, we define the graph as \( \mathcal{G}^\prime \) and the corresponding error energy function as \( \Psi(\mathcal{G}^\prime) \). The global error increment is
\[
\Delta \Psi = \Psi(\mathcal{G}^\prime) - \Psi(\mathcal{G}).
\]
Given the perturbations, the expected deviation of the global error increment is derived as:
\[
\mathbb{E}[\Delta \Psi] = \left( 1 + \frac{\lambda q}{1 - \lambda q} \right) \cdot |\mathcal{V}_{\mathrm{atk}}| \cdot B^2 \cdot K.
\]
This expression shows that the attack’s impact on the graph is proportional to the number of fake nodes \( |\mathcal{V}_{\mathrm{atk}}| \), the strength of the perturbation \( B^2 \), and the number of aggregation layers \( K \). The term \( \left( 1 + \frac{\lambda q}{1 - \lambda q} \right) \) reflects the influence of aggregation, where \( \lambda \) controls the trade-off between node and edge feature similarities, and \( q \) is the probability of connectivity between fake nodes.

\section{Additional Details of Experiments}

\subsection{Hyperparameters}\label{c.1}
For the hyperparameters, we conduct a grid search to identify the optimal choice. The learning rate is selected from the set $\{10^{-1}, 10^{-2},$ $10^{-3}\}$, the weight decay is chosen from \(\{10^{-3}, 10^{-4}, 10^{-5}, 0\}\), the dropout rate is sampled from \(\{10^{-1}, 10^{-2}, 10^{-3}, 0\}\). All models are trained for up to 300 epochs using the Adam optimizer~\cite{kingma2014adam}, and the best model was selected for testing based on validation loss.

\subsection{More Details About Validation}\label{c.2}
In our experiments, we comprehensively evaluate the effectiveness of our attack methodology by considering two fundamental tasks: node classification and link prediction. Furthermore, we assess our attack under both \textit{targeted} and \textit{untargeted} attack setups. 
\begin{itemize}[leftmargin=*, itemindent=0em]
    \item \textit{Targeted Attack}. This setup aims to evaluate the attacker's impact on a specific target node \(v_t\in\mathcal{V}_t\). The goal is to assess how the attack alters the performance related to this particular node, such as its classification accuracy or predictive capabilities.
     \item \textit{Untargeted Attack}. In contrast, the untargeted attack measures the overall effectiveness of the attack across the entire graph, without focusing on any specific node. This setup aims to explore the broader effects of the attack on the graph as a whole.
\end{itemize}

\subsection{More Experimental Results and Analysis}
\subsubsection{Attack effectiveness under different GNN models}\label{e.2}
In Sec.~\ref{sec6.3}, we analyze the attack performance under different GNN models. Tables~\ref{tab2} and \ref{tab3} present the experimental results under the targeted and untargeted attack settings, respectively. The results from the untargeted attack setting provide further support to the analysis in Section~\ref{sec6.3}, demonstrating the consistency and effectiveness of our attack across different GNN models.

\subsubsection{On the analysis of $\eta_1$ and $\eta_2$}\label{e.4}
\begin{figure}[t]
	\centering
    \vspace{-1em}
	\begin{tabular}{cc}
		\multicolumn{2}{c}{\textbf{}} \\
  \includegraphics[width=0.45\linewidth]{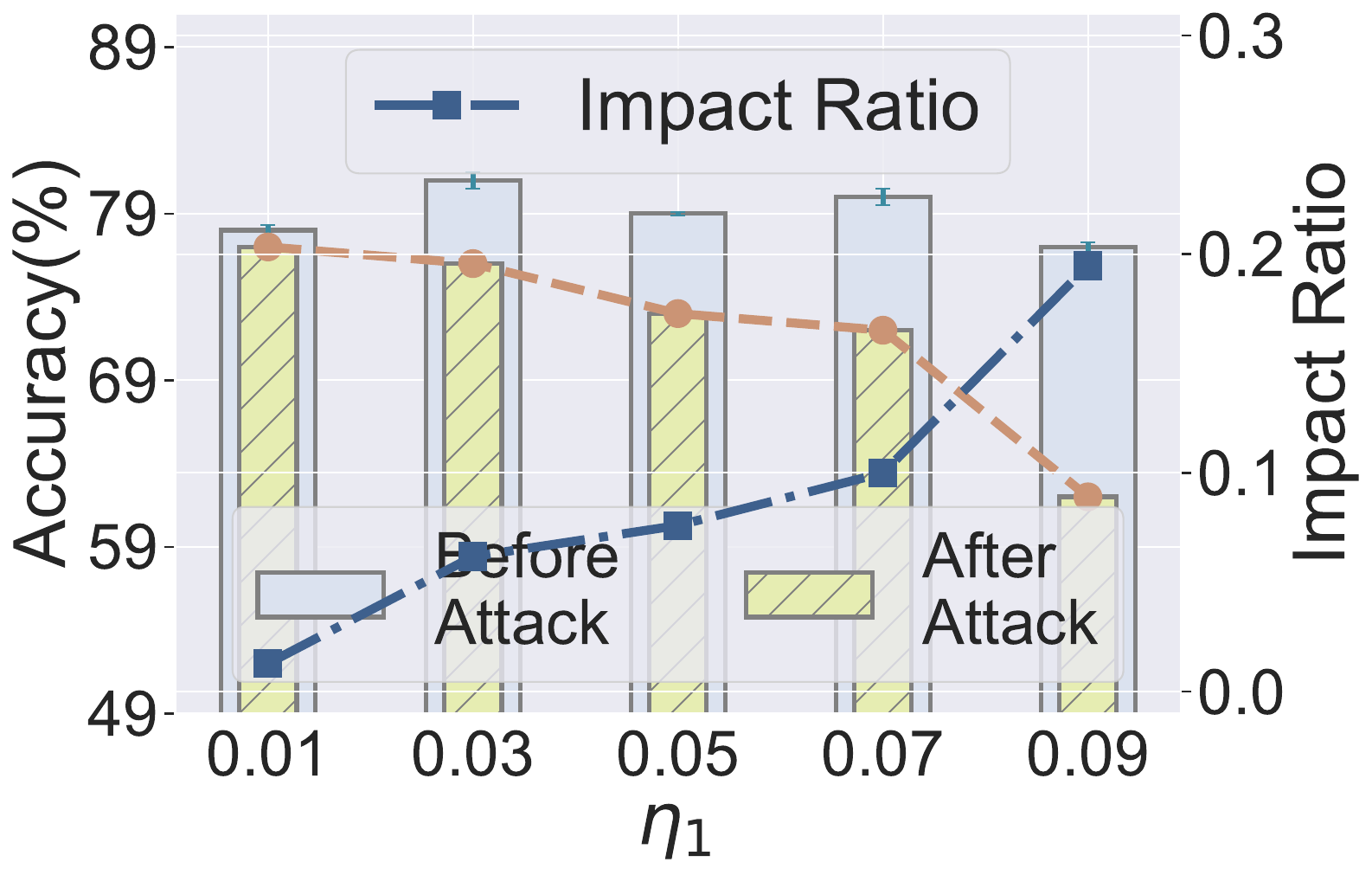} & \includegraphics[width=0.46\linewidth]{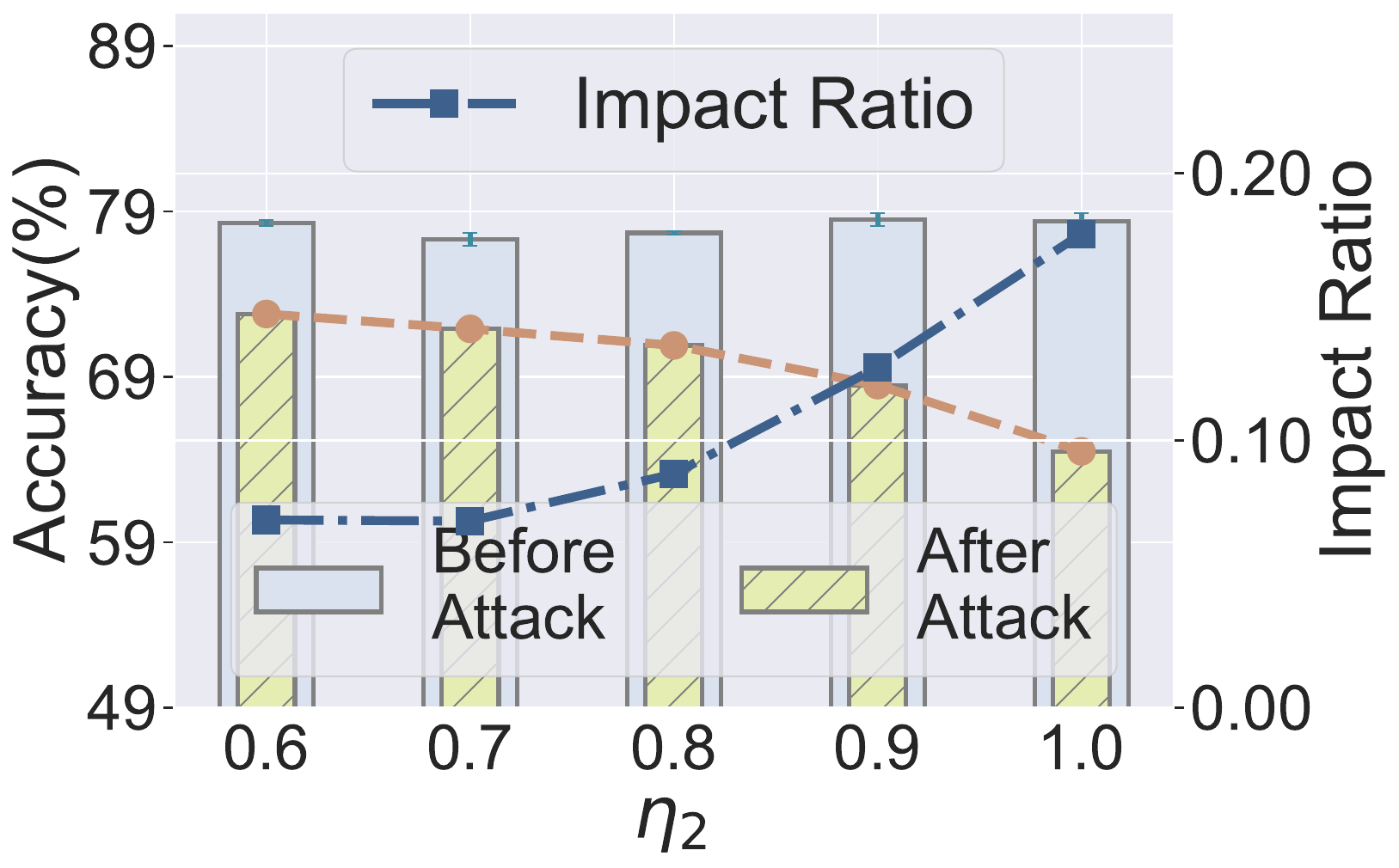}\\
	\end{tabular}
    \vspace{-1em}
	\caption{Variation in untargeted attack accuracy at different $\eta_1$ and $\eta_2$ values. The impact ratio represents the proportion of accuracy reduction compared to the normal accuracy.}
	\label{fig9}
\end{figure}

In Sections~\ref{6.4.1} and~\ref{6.4.2}, we analyze the effects of the parameters \(\eta_1\) and \(\eta_2\) on the performance of our attack. Fig.~\ref{fig4} presents the experimental results under the targeted attack setting, while Fig.~\ref{fig9} shows the experimental results for the untargeted attack setting. The experiments conducted under the untargeted attack setting further corroborate the analysis of the results presented in Sec~\ref{6.4.1} and~\ref{6.4.2}, providing additional evidence for the effectiveness of our attack under different configurations.
\subsubsection{On the analysis of the parameter $K$}\label{e.5}
In Sec.~\ref{6.4.3}, we validate the change in accuracy before and after the attack with different values of \(K\in\{0,2,4,8,16\}\). The results are shown in Fig.~\ref{fig6}. The parameter \(K\) represents the number of steps in the second phase (calibration) of the locally private graph learning protocol \(\Pi\). As seen in Fig.~\ref{fig6}, the accuracy before and after the attack exhibits an \textit{increasing and then decreasing} trend with the increasing \(K\).

The initial increase in accuracy can be attributed to the calibration of noise by neighbor aggregation. As defined in Def.~\ref{def3}~\cite{sajadmanesh2021locally}, the error deviation is measured by the embedding disparity of \(v\) before and after the perturbation, which is influenced by the neighborhood size \(|\mathcal{N}(v)|\) as given in Eq.~(\ref{eqdef3}). The estimation error decreases at a rate proportional to the square root of the node degree. Therefore, increasing \(K\) expands the neighborhood that can be aggregated, facilitating better noise calibration, which improves the accuracy.

\begin{definition}[Error Deviation]\label{def3}
Given the aggregator function for the first layer and $\delta>0$, with probability at least $1 - \delta$ for any node $v$, we have the following bound on the aggregation result:
\begin{equation}
\max_{i\in\left \{ 1,2,\cdots,d \right \} } \left |(\widehat{ \mathbf{h}}^{(1)}_v  )_i - (\mathbf{h}^{(1)}_v)_i \right |=\mathcal{O}\left ( \frac{\sqrt{d \,\mathrm{log}(d/\delta )} }{\epsilon \sqrt{|\mathcal{N}(v)| } }  \right ), \label{eqdef3} 
\end{equation}
where $\widehat{ \mathbf{h}}^{(1)}_v$ is the perturbed aggregated embedding of node $v$ at the first layer, and $\mathbf{h}^{(1)}_v$ is the true aggregation without noise. \( \mathcal{N}(v) \) is the set of neighbors of node \( v \).
\end{definition}

However, the later decrease in accuracy when \(K\) becomes large is due to over-smoothing~\cite{li2018deeper}. Over-smoothing occurs as node features tend to converge towards a common value with an increasing number of aggregation layers, resulting in diminished utility for graph learning tasks. This is a well-known issue in GNNs~\cite{keriven2022not}. Importantly, our attack method significantly reduces the accuracy of the original method across all values of \(K\). This validates the generalization and effectiveness of our attack approach, showing that it is robust to different values of \(K\) and can significantly degrade the performance of the locally private graph learning protocol.

\subsubsection{Metrics for graph homophily analysis}\label{app4.2}
In the defense experiments in Sec.~\ref{sec7.1}, we consider two homology metrics~\cite{chenunderstanding}: \textit{node-centered homology} and \textit{edge-centered homology}. The former evaluates the similarity between a target node and its neighboring nodes, while the latter assesses the similarity between nodes connected by a target edge. They are defined as follows:
\begin{definition}[Node-Centric Homophily]
The homophily of a node $v$ is quantified by measuring the similarity between its feature vector and the aggregated embedding of its neighboring nodes:
\begin{equation}
h_v=sim\left ( r_v, \mathbf{x}_v \right ),\quad r_v= \sum_{j\in \mathcal{N} (v)}\frac{1}{\sqrt{d_j} \sqrt{d_v}} \mathbf{x}_j, 
\end{equation}
where $\mathbf{x}_v$ denotes the feature vector of node $v$, $d_v$ represents the degree of $v$, and $sim(\cdot)$ is a similarity metric, \textit{e.g.}, cosine similarity.
\end{definition}

\begin{definition}[Edge-Centric Homophily]
The homophily for an edge $(v,u)$ can be defined as follows:
\begin{equation}
h_e=sim\left ( \mathbf{x}_v, \mathbf{x}_u \right ), 
\end{equation}
where $\mathbf{x}_v$ denotes the feature vector of node $v$ and $sim(\cdot)$ represents a distance metric, \textit{e.g.}, cosine similarity.
\end{definition}

\section{More Details on Related Work}\label{ddd}
Recently, locally private graph learning protocols have attracted significant attention in the security research community~\cite{sajadmanesh2021locally,lin2022towards,pei2023privacy,li2024privacy,qi2024linkguard,zhu2023blink,hidano2022degree,zhang2024locally}. \citeauthor{sajadmanesh2021locally}~\cite{sajadmanesh2021locally} extended the one-bit mechanism~\cite{ding2017collecting} to multidimensional space and introduced the multi-bit mechanism (MB) for protecting node features, which serves as a foundation for private graph learning. \citeauthor{lin2022towards}~\cite{lin2022towards} further refined the locally private graph learning protocol based on the MB mechanism~\cite{sajadmanesh2021locally}. Additionally, \citeauthor{pei2021decentralized}~\cite{pei2021decentralized} investigated privacy-preserving learning for decentralized graphs using the piecewise mechanism (PM)~\cite{wang2019collecting}, while \citeauthor{li2024privacy}~\cite{li2024privacy} extended the original square wave mechanism (SW)~\cite{li2020estimating} to enhance the utility of locally private graph learning. Some studies~\cite{qi2024linkguard,zhu2023blink,hidano2022degree,zhang2024locally,lei2025achieving} have also explored link protection. These approaches leverage the benefits of LDP~\cite{yang2024local,dwork2006calibrating} in safeguarding data privacy and ensuring strict privacy guarantees for user data. Moreover, during server-side privacy graph learning, GNN's multi-layer message passing mechanism
~\cite{feng2022powerful,abu2019mixhop} effectively reduces the estimation error in node embeddings after perturbation and aids in the calibration of noisy data, thus maintaining high utility for the local privacy graph learning task. However, despite these advantages, such protocols may be vulnerable to data poisoning attacks, a threat that has not been considered in previous research. Identifying and addressing these threats is crucial for ensuring the robustness and security of private graph learning frameworks. This work introduces the first data poisoning attack targeting locally private graph learning protocols.
\end{document}